\documentclass[runningheads]{llncs}

\usepackage{eccv}

\usepackage{eccvabbrv}

\usepackage{mathrsfs}
\usepackage{siunitx}
\usepackage{adjustbox}  %

\newcommand{\methodname}[0]{EquiFusion}

\usepackage{overpic}
\usepackage[pdftex,outline]{contour}

\usepackage{tabstackengine}
\TABstackMath
\setstackgap{S}{2pt}

\usepackage{array}
\newcolumntype{H}{>{\setbox0=\hbox\bgroup}c<{\egroup}@{}}
\newcolumntype{L}[1]{>{\raggedright\let\newline\\\arraybackslash\hspace{0pt}}m{#1}}
\newcolumntype{C}[1]{>{\centering\let\newline\\\arraybackslash\hspace{0pt}}m{#1}}
\newcolumntype{R}[1]{>{\raggedleft\let\newline\\\arraybackslash\hspace{0pt}}m{#1}}

\usepackage{amssymb}%
\usepackage{pifont}%
\newcommand{\cmark}{\ding{51}}%
\newcommand{\xmark}{\ding{55}}%

\newcommand{\gmark}{\textcolor{green!60!black}{\ding{51}}}  %
\newcommand{\rmark}{\textcolor{red!70!black}{\ding{55}}}   %
\newcommand{\ymark}{\textcolor{black!70!black}{--}}       %

\definecolor{pastgreen}{rgb}{0.0, 0.59019, 0.362745}
\definecolor{futurewhite}{rgb}{0.9431372,0.9431372, 0.9431372}

\definecolor{bluepred}{rgb}{0.0, 0.0, 0.960784}

\definecolor{futurewhite2}{rgb}{0.49,0.49, 0.49}
\definecolor{bluepred2}{rgb}{0.184,0.255, 0.533}

\def\Kkin#1{\mathcal{K}^{#1}} %
\def\Weights{\mathbf{W}}
\def\futureframes{T_F}
\def\pastframes{T_P}
\def\adj{\mathbf{A}}

\def\latentvar#1{\boldsymbol{z}_{#1}}

\def\motion{\textbf{M}}

\def\joints{\textbf{M}}
\def\dir#1#2#3{^{#1}\textbf{L}_{#2}^{#3}}
\def\pastmotion#1#2{^{#1}\textbf{X}_{#2}}
\def\futuremotion#1#2{^{#1}\textbf{Y}_{#2}}
\def\predmotion#1#2{\tilde{\futuremotion{#1}{#2}}}
\def\perm{\mathbf{P}}
\def\netfunct{f}
\def\numnodes{N}
\def\numbones{B}
\def\featvar{\textbf{Z}}
\def\feat{\featvar^{}}

\def\identity{\mathbb{I}}
\def\alphaT#1#2{#1{\alpha}_{#2}}

\def\pastvar{\textbf{X}}
\def\futurevar{\textbf{Y}}

\def\futurewindow{F}

\def\numjoints{J}
\def\encnet{\mathrm{{e}}} %

\definecolor{pastgreen}{rgb}{0.0, 0.59019, 0.362745}
\definecolor{futurewhite}{rgb}{0.9431372,0.9431372, 0.9431372}

\definecolor{bluepred}{rgb}{0.0, 0.0, 0.960784}

\definecolor{futurewhite2}{rgb}{0.49,0.49, 0.49}
\definecolor{bluepred2}{rgb}{0.184,0.255, 0.533}
\definecolor{amass}{RGB}{142 99 124}
\definecolor{h36m}{RGB}{88 113 86}
\definecolor{nymeria}{RGB}{159 153 83}
\definecolor{occluded}{RGB}{148 78 64}
\definecolor{limb_gen}{RGB}{55 139 165}

\usepackage[accsupp]{axessibility}  %
\usepackage{graphicx}
\usepackage{booktabs}
\usepackage[pagebackref,breaklinks,colorlinks,citecolor=eccvblue]{hyperref}

\usepackage{orcidlink}

\def\app {App.}
\usepackage{multirow}

\addtocontents{toc}{\protect\setcounter{tocdepth}{-1}}

\title{\methodname: Kinematics-Agnostic Human Motion Prediction via Equivariant Latent Diffusion} 
\titlerunning{\methodname: Kinematics-Agnostic HMP}

\author{
Cecilia Curreli\inst{1,2}  \hspace{0.7cm}  Florian Hofherr\inst{1,2}  \hspace{0.7cm}  Dominik Muhle\inst{1,2} \\
Abhishek Saroha\inst{1,2}  \hspace{0.7cm}  Riccardo Marin\inst{1,2} \hspace{0.7cm}   Daniel Cremers\inst{1,2} \\
}
\authorrunning{C.~Curreli et al.}
\institute{Technical University of Munich, Munich, Germany \and
Munich Center for Machine Learning, Munich, Germany}

\definecolor{teaserpink}{rgb}{1.0, 0.411764, 0.7058823}

\newcommand{\crefcrossfile}[1]{\cref{#1}}

\begin{document}
\setcounter{tocdepth}{2}
\maketitle

\begin{center}
\captionsetup{type=figure}
    \includegraphics[width=0.99\linewidth]{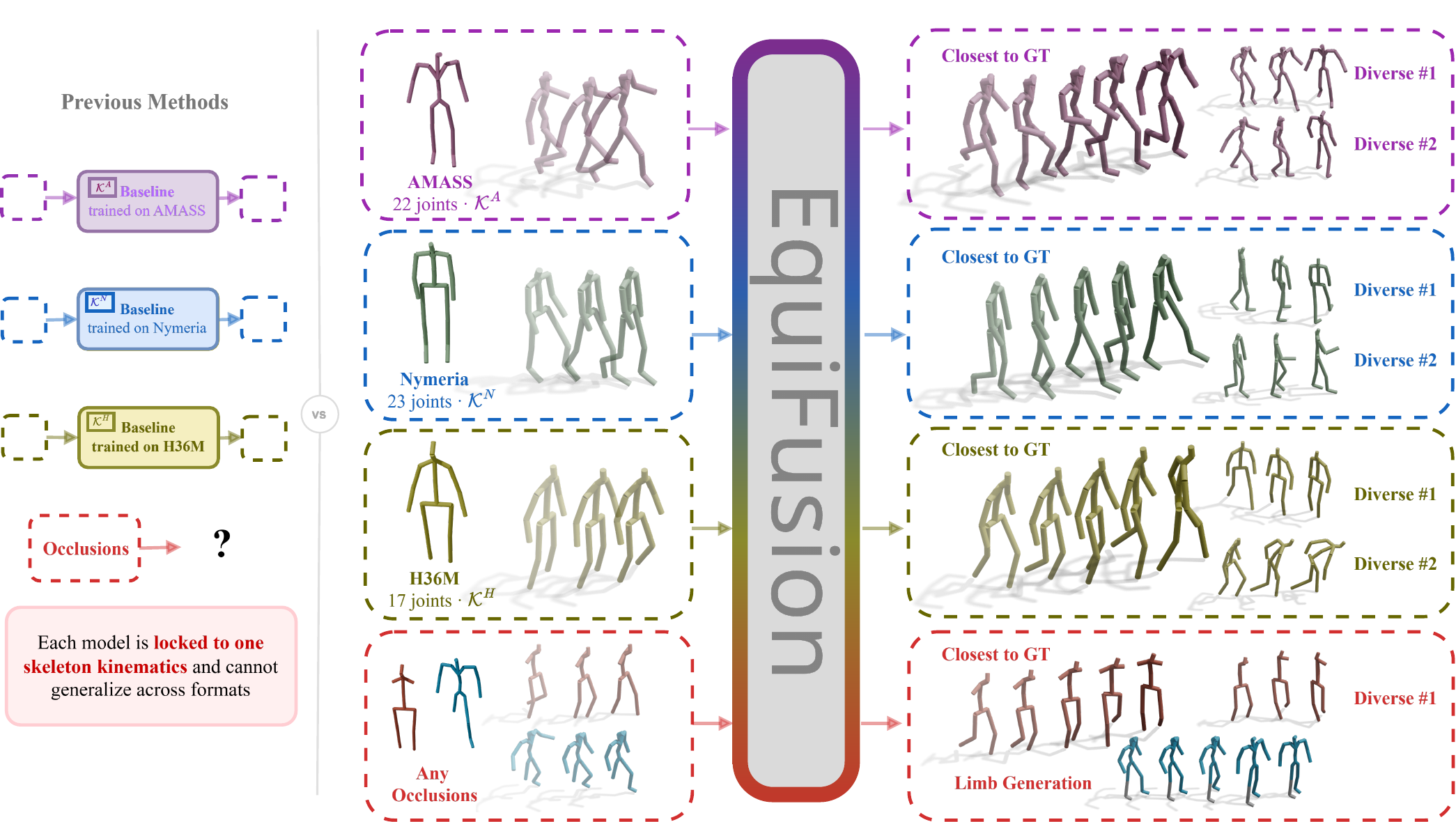}
 
  \caption{\textbf{\methodname}. We introduce the first model for stochastic human motion prediction that generalizes to unseen skeleton parameterization, i.e. \textit{kinematics}. While previous methods require a trained instance for each dataset or better kinematics, with a single model we unlock training on multiple datasets and inference on motion parametrized with different kinematics. \methodname~ is the first SHMP model to handle \underline{\textbf{zero-shot}} novel kinematics and occluded limbs without explicit training,  while achieving state-of-the-art results on established benchmarks.
  }
\label{fig:teaser}
\end{center}

\begin{abstract}
    Existing Stochastic 3D Human Motion Prediction models are fundamentally constrained by hard-coding the skeleton kinematics, severely limiting generalization, preventing cross-dataset training, and requiring complex data retargeting. We introduce {\em\methodname}, the first kinematics-agnostic model to solve this bottleneck, implementing a latent diffusion model with a permutation equivariant architecture. \methodname~treats the kinematics' connectivity as an explicit input parameter, ensuring its internal computations are inherently agnostic to joint ordering and graph structure. This novel design enables truly cross-dataset generalization to unseen kinematics and unlocks novel zero-shot directions, such as motion prediction from partial or occluded observations and targeted limb generation. \methodname~achieves state-of-the-art results on major benchmarks, being up to 75\% more compact than previous kinematics-specific methods, while achieving faster training and inference. \methodname~thus establishes a new, flexible standard for robust human motion prediction. 
    Model and training code available on our \href{https://ceveloper.github.io/publications/equifusion/}{project page}.
\end{abstract}

\section{Introduction}
\label{sec:intro}
Predicting future human motion from past observations is a core component of human intelligence and a prerequisite for mature spatial AI. 
Due to the inherent ambiguity of human intent, the field has shifted from deterministic toward Stochastic Human Motion Prediction (SHMP), which models a distribution of  diverse, physically plausible futures from an observation, with applications including human-robot collaboration, autonomous navigation, and augmented telepresence. 
While recent advances in generative models~\cite{ho2020denoising, rombach2022highresolution, videoworldsimulators2024} have further popularized SHMP probabilistic formulations~\cite{suncomusion, curreli2025nonisotropic, barquero2023belfusion, chen2023humanmac}, a fundamental bottleneck remains: kinematics rigidity. 
Diverse datasets (e.g., Human3.6M~\cite{Ionescu2014}, AMASS \cite{mahmood2019amass}, and more recently Nymeria \cite{ma2024nymeria}) often employ different motion-capturing technologies, resulting in diverse skeleton parametrizations \ie \emph{kinematics}. 
Motion retargeting~\cite{alibeigi2016fast, lee2023same, holden2016deep}, \ie, converting a motion to a different kinematics parametrization, is often not possible without introducing errors and drifts~\cite{villegas2018neural, lim2019pmnet, villegas2021contact}. The downstream impact on SHMP methods is tremendous. Previous works~\cite{curreli2025nonisotropic, barquero2023belfusion, chen2023humanmac, yuan2020dlow, suncomusion, dang2022diverse, mao2021generating} have hard-coded kinematics as a network structural prior, resulting in a plethora of skeleton-specific networks. Training a network for every different kinematics is impractical, inefficient, and does not generalize to new skeletal configurations.

We break these dataset-specific boundaries by addressing settings involving heterogeneous kinematic structures. We propose two \textit{zero-shot kinematics} inference tasks for SHMP: predicting motion for (i) full-body kinematics unseen at training time, and (ii) kinematics with occlusions or missing parts (\ie, partial observations). 
Both these tasks address real-world use cases and important milestones toward mature spatial AI and foundation human motion models. To date, this has been unaddressed in the SHMP literature, while deterministic HMP solutions (\eg, hand-crafted training to address specific occlusions or skeletal conversions ~\cite{cui2021towards, xu2023auxiliary}) are impractical and require manual engineering. Can we provide such flexibility for SHMP?

In this paper, we answer this question by proposing the first kinematics-agnostic method for SHMP. To handle arbitrary kinematics chains, a model’s learned weights must be independent of the cardinality of the joint set. Most existing works violate this principle by employing joint-dependent weights or temporal-only transformers that treat joints as feature channels. We identify permutation equivariance as the key mathematical property to achieve this independence.
We thus present \methodname~, the first kinematics-agnostic approach to SHMP, implemented as an equivariant latent diffusion model that explicitly incorporates the kinematics connectivity as a model input.
By design, we achieve robustness to partial and unseen kinematics without requiring explicit exposure to such corruptions during training, and we demonstrate this through extensive experiments.
Our design allows us to break the single-dataset barrier, enabling a single model to be trained simultaneously on heterogeneous datasets such as AMASS and Nymeria for the first time, unlocking the future potential of ``foundation models'' in human motion.
By leveraging the inductive bias of equivariance with formal guarantee \cite{lyle2020benefits, bietti2021sample}, \methodname~ achieves state-of-the-art performance on standard benchmarks even when trained on single-dataset distributions while being \emph{75\% smaller} than the closest competitor~\cite{curreli2025nonisotropic}, and significantly faster in training and inference. These improvements are particularly remarkable in zero-shot scenarios, where our model demonstrates a performance boost of at least 25\% across all metrics and up to 70\% in terms of fidelity. Our contributions are:
\begin{itemize}
\item We formalize and analyze for the first time the challenge of zero-shot kinematics SHMP, introducing two novel tasks of high relevance for the real-world setting, while opening new applications and use cases: training on datasets having multiple kinematics, zero-shot handling of occlusion, and zero-shot limb generation without ad-hoc training.
\item We identify joint-ordering permutation equivariance as a key property to enable kinematics-agnostic SHMP and propose a principled architecture that inherently satisfies this requirement.
\item We present \methodname, which not only achieves zero-shot kinematics by design but  
outperforms existing SHMP methods in extensive experiments. Our method achieves state-of-the-art results on standard benchmarks (AMASS, Human3.6M) with  75\% fewer parameters than the closest competitor. Additionally, our model’s complexity remains invariant to the number of joints, kinematics, or datasets, offering superior scalability.
\end{itemize}
\vspace{-0.27cm}

\section{Related Works}

\label{sec:related_works}
\subsection{Human Kinematics Representations}\label{sec:rel_works:skeletons}

In SHMP, human motions are represented as the temporal evolution of body joints; we refer to them as skeleton kinematics $\Kkin{}$, which are generally inherited from the Motion Capture (MoCap) system used to register the human movement. Hence, depending on the MoCap system, the kinematic system can differ in the number and positions of the joints, resulting in a plethora of different formats. Examples are: the AMASS \cite{mahmood2019amass} dataset uses the SMPL \cite{SMPL2015} format, and is parametrized by 22 joints; H36M\cite{Ionescu2014} kinematics comprises 17 joints and does not include the foot joints; Nymeria\cite{ma2024nymeria} has been captured by  META's ARIA devices \cite{engel2023project} and parametrized with XSens \cite{mvnlink} by 23 joints, exhibiting different spine and hip modeling from the aforementioned formats (see \cref{fig:teaser} for visualizations). With the increasing availability and affordability of wearable sensors \cite{wang2023wearable, engel2023project, petrov2025echo, guzov2024interaction}, new kinematic configurations are expected to emerge \cite{yang2025strengthsense, fritsche2025ultra}, making it crucial to address multiple kinematics, since differences in kinematic structure hinder data combination during training\cite{engel2023project}.

\subsection{Stochastic Human Motion Prediction}
Human Motion Prediction (HMP) aims to predict future motion from past  observations. While deterministic HMP  \cite{cui2020learning, li2020dynamic, li2019actional} forecasts a single future,  Stochastic HMP (SHMP) predicts multiple plausible ones. 
SHMP is an ill-posed problem with multiple solutions, which has been modeled by different generative approaches involving GANs \cite{barsoum2018hp, kundu2019bihmp, liu2021aggregated}, VAEs \cite{walker2017pose, yan2018mt, cai2021unified, mao2021generating, gu2024learning}, diverse sampling \cite{dang2022diverse, yuan2020dlow, xu2022diverse}, and lately denoising diffusion models \cite{suncomusion, chen2023humanmac, barquero2023belfusion, saadatnejad2023generic, wei2023human, curreli2025nonisotropic}. 
All such models are trained and evaluated independently for each kinematics $\Kkin{i}$, which means that each requires ad hoc training, computational waste, hyperparameter engineering, and lacks generalization at inference time to novel kinematics $\Kkin{j}\,\text{with}\, j\neq i$, or occluded data (\ie, partial kinematics). Digesting motions in different kinematics formats has been tackled in other computer vision tasks, such as human pose estimation from video \cite{sarandi2025neural},  2D-to-3D keypoints lifting \cite{dabhi20243d}, unconditional animal motion generation \cite{gat2025anytop}, motion classification \cite{lee2023same}, or character animation via text \cite{huang2025animaxanimatinginanimate3d, liu2025text}.  Occluded motions and training on multiple kinematics have already been investigated by deterministic HMP,  but to the best of our knowledge, only explicitly, by directly training for occlusions \cite{cui2021towards, xu2023auxiliary} or for multiple skeletons without supporting zero-shot inference on new kinematics \cite{sarandi2023learning}. Yet, no method has been proposed for SHMP, which is the core of our work and a mandatory stage towards foundation models for SHMP. We provide an extended discussion on existing methods in the \cref{app:rel_works:shmp,app:rel_works:det_hmp,app:rel_works:others}. We will build our framework on top of a newly designed permutation equivariant diffusion model. Diffusion methods equivariant to SE(3) or permutation groups have provided promising results for the task of drug molecule generation \cite {schneuing2024structure, 11357162, laabid2024equivariant}, but they haven't found broad application in computer vision, especially not in HMP or SHMP to date. A reasonable alternative to enable flexibility of existing approaches would be solving for skeleton retargeting, and we discuss such techniques in the next section.

\vspace{-0.2cm}

\subsection{Kinematics Conversion and Motion Retargeting} 
\label{sec:rel_works:retargeting}
As current SHMP approaches do not support inference on kinematics  $\Kkin{j}$ different from the training kinematics $\Kkin{i}$, the only way to allow inference on $\Kkin{j}$ is to first convert the input motion from kinematics $\Kkin{j}$ to the same motion with $\Kkin{i}$. 
 This conversion challenge is commonly addressed as \textit{skeleton retargeting}. The  large literature on this topic   ranges from robotics \cite{figuera2024redefining, yoon2024spatio, delhaisse2017transfer} to computer graphics \cite{li2023ace,  alibeigi2016fast, hu2023pose, lee2023same} and specializes in isomorphic \cite{gleicher1998retargetting, lee1999hierarchical, choi2000online, tak2005physically, feng2012automating, jang2018variational, delhaisse2017transfer, villegas2018neural, lim2019pmnet, aberman2019learning, villegas2021contact, musoni2021reposing, musoni2021functional},
homeomorphic\cite{aberman2020skeleton, hu2023pose, zhang2024semantics}, and 
non-homeomorphic graphs~\cite{jang2024geometry,mourot2023humot, cao2025g, kim2025moreflow, li2024walkthedog,holden2016deep, wang2023zero, liao2022skeleton, lee2023same}. 
For the SHMP use case, only \textit{non-homeomorphic} techniques can be considered, since kinematics can have different end-effectors and a varying number of joints. As not all approaches are applicable to SHMP's data structure, in this paper, we consider the retargeting approach from Holden \etal\cite{holden2016deep}, which is already established in SHMP by HumanMAC\cite{chen2023humanmac}. More recent approaches for non-homeomorphic retargeting on robotics and character animation are not applicable to our case. These require joint rotations of end-effectors and reference poses~\cite{lee2023same},  skinning ~\cite{liao2022skeleton},  meshes \cite{wang2023zero, saito2026soma}, or both ~\cite{jang2024geometry} -- even without the skeleton itself \cite{liao2022skeleton, wang2023zero} -- which are not always available in SHMP. Or, they present a too high reconversion error (ca. 100mm\cite{mourot2023humot} or 90mm \cite{cao2025g}, in both cases, code is not available). We discuss retargeting approaches and each graph category in detail in \cref{app:rel_works:retargeting}, and here we maintain a high-level contextualization, taking as an example the AMASS and H36M skeletons depicted in \cref{fig:teaser}.  
 These kinematics are {non-homeomorphic} to each other, so the conversion is not bijective and causes: \textbf{(a) Error accumulation}, as converting H36M (17 joints) to AMASS (22 joints) requires adding non-existent feet (with errors of \SI{22.18}{\milli\metre} or \SI{2.27}{\milli\metre} depending on the conversion direction), \textbf{(b) Distribution shift}: additionally, this kinematics conversion shifts the input to a distribution that differs from the one seen by the SHMP models at train time, resulting in additional noise for the models. 
 As SHMP approaches achieve precision up to \SI{70}{\milli\metre}, the additional retargeting error is not negligible. 

\vspace{-0.2cm}

\subsection{Human Pose Representations} 
\label{sec:related_works:motion_param}
\vspace{-0.06cm}
Another relevant aspect of SHMP is the motion parametrization. In SHMP, motion sequences are conventionally represented as trajectories of 3D joint coordinates in Euclidean space\cite{yuan2020dlow, dang2022diverse, barquero2023belfusion, mao2021generating, wei2023human, walker2017pose}. While such representation is flexible and a natural output of upstream pipelines \cite{lohit2021recovering, simon2017hand, cao2017realtime, wei2016cpm}, \eg human tracking in videos \cite{zhou2023human, park2020hmpo, roudsarabi2008solving, phu2025predicting}, it is under-constrained and allows non-realistic predictions. This often results in limb stretching or jitter, requiring subsequent SHMP approaches to directly measure \cite{curreli2025nonisotropic}  and address  \cite{barquero2023belfusion, curreli2025nonisotropic, dang2022diverse} this issue. Instead of making the model learn from data what we already know \ie  human bones have fixed length, we propose to include this prior directly in the parametrization. Intuitively, consistent limb lengths are guaranteed by design in rotation-based representation, where the joint positions are expressed as angles from a canonical pose, and the bone lengths are fixed along a sequence. Among the many possible representations for rotations \cite{geist2024learning}, some common choices are quaternions \cite{salzmann2022motron} or axis-angles \cite{SMPL2015, tevet2022motionclip, li2025unimotion, tang2025stochastic} as in SMPL\cite{SMPL2015}. However, rotations also tie the pose representation to a specific skeleton kinematic chain and rest pose, which can cause instabilities \cite{barquero2023belfusion}.
In this work, we adopt a spatial representation, where every joint is encoded by the relative direction from its parent, \ie its \textit{bone direction}. In this way, predictions are guaranteed to have perfectly consistent bones by design, and networks do not have to learn this additional prior. This representation is general and can be applied to both any SHMP kinematics and any SHMP method.

\section{Methodology}
\label{method}
\begin{figure*}[t]
  \centering
  \includegraphics[clip,width=0.99\textwidth]{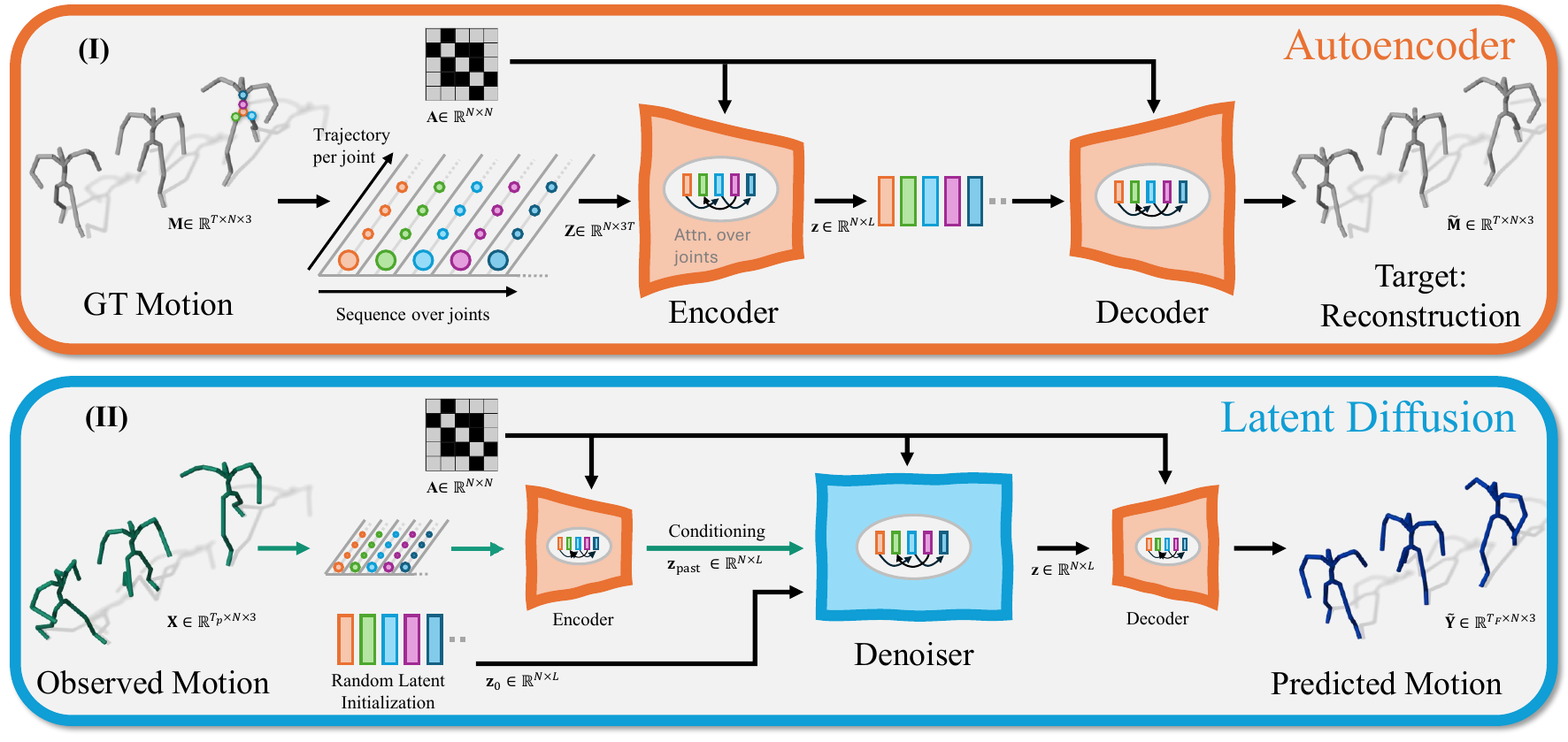}
\caption{
    \textbf{\methodname~is the first skeleton-agnostic model for SHMP, implemented as a novel equivariant latent diffusion model that adds the adjacency matrix $\adj$ of the motion kinematics as input}. 
     \textbf{(I)} A transformer autoencoder learns a latent space $\latentvar{}$ equivariant to joint order permutation. \textbf{(II)} A denoiser predicts future motion in this latent domain conditioned on the observed past $\pastvar{}$.
  \vspace{-2em}
}
  \label{fig:method}
\end{figure*}

\subsection{Problem Definition}
\label{sec:method:problem_def}
In stochastic human motion prediction (SHMP), a motion sequence is represented as a trajectory $\motion \in \mathbb{R}^{T \times \numjoints \times 3}$, describing the temporal evolution of $\numjoints$ body joints over $T$ frames. Given an observed motion  $\pastmotion{}{} = \motion_{0:\pastframes} \in \mathbb{R}^{\pastframes\times\numjoints\times3}$ of length $\pastframes$, the objective is to generate multiple plausible future trajectories $\predmotion{}{}\in\mathbb{R}^{\futureframes\times\numjoints\times3}$  over a prediction horizon of $\futureframes$ frames.
Each motion sequence is described with respect to a kinematic structure $\Kkin{}$, as already discussed in  \cref{sec:rel_works:skeletons}. Formally,  we define a  kinematics $\Kkin{}:= (V_{\boldsymbol{\tau}}, E)$ as an undirected graph with semantically labeled vertices $V_{\boldsymbol{\tau}}$ and edges $E$. We refer to  nodes as joints and edges as limbs or bones, and we define the cardinality $|\Kkin{}|$ equal to the number of joints.  Each vertex $v_{\boldsymbol{\tau},i}$ represents a joint with a connected semantic convention label that distinguishes among joint types and kinematics conventions. 
The kinematic configuration $\Kkin{}$ is fixed within each dataset across all motion sequences, so we usually refer to a kinematics with the name of the dataset: \eg $\Kkin{A}$ for AMASS \cite{mahmood2019amass}, $\Kkin{H}$ for H36M~\cite{Ionescu2014}, and $\Kkin{N}$ for Nymeria~\cite{ma2024nymeria}, where different individuals in the same dataset have the same kinematics $\Kkin{}$ but different bone lengths. 
Typically, models are evaluated on the test set of the dataset they were trained on, \textit{in-domain}, and within the same kinematics. The popular mesh SMPL model, which underlies the AMASS kinematics, has led to smaller data collections~\cite{tripathi2023ipman, vonMarcard2018} that share the kinematics $\Kkin{A}$. These datasets are occasionally used for evaluation on out-of-distribution motions, i.e., zero-shot motion for SHMP. For the first time in SHMP, we investigate \textbf{zero-shot kinematics}, the case where inference kinematics differ from those seen at training. While both evaluations are theoretically independent, the nature of the datasets leads to zero-shot kinematics, which often implies zero-shot motion. Zero-shot kinematics also covers \textit{partial} kinematics, representing limbs that are occluded or missing due to physical impairments. 
While partiality is highly relevant for real-world applications, data collection is not straightforward, and there are no specific SHMP datasets to date. Partial motions are thus usually investigated by masking limbs of motions parametrized with existing full-body kinematics.

\subsection{Kinematics-Agnostic SHMP}
While kinematics-agnostic models have been investigated in other domains, such as character animation from text \cite{liu2025text, gat2025anytop}, motion classification \cite{lee2023same}, and robotics \cite{alattar2022kinematic}, the problem of addressing multiple skeleton kinematics has not been addressed or formalized in SHMP so far. Existing SHMP approaches assume a single, fixed skeleton kinematics inherited from the training dataset~\cite{yuan2020dlow, dang2022diverse, hu2023pose, gil2023human, barquero2023belfusion, chen2023humanmac, suncomusion, curreli2025nonisotropic}. Instead, we are interested in a model that natively supports 1) training on multiple kinematics, and 2) inference on novel kinematics, partial or full-body, not seen at training time, i.e., in a zero-shot kinematics setting. 
We call such a model \textit{kinematics-agnostic}.

Formally, let $\mathbb{K}_{\text{train}} = \{ \Kkin{1}, \Kkin{2}, \ldots, \Kkin{S}\}$ be the set of kinematic settings seen a training time by a method $f$, $S$ the number of train skeletons, and $\mathbb{K}_{\text{test}}$ the set of kinematics used for evaluation. When we perform evaluation on a motion under the kinematic setting $\Kkin{'} \in \mathbb{K}_{\text{test}}$ with $\mathbb{K}_{\text{train}} \cap \mathbb{K}_{\text{test}} = \emptyset$, we regard this as a zero-shot scenario.
Previous SHMP approaches have been limited to $N = 1$ and $\mathbb{K}_{\text{train}} = \mathbb{K}_{\text{test}}$ due to their kinematics-specific design decisions, requiring a trained instance for each kinematics.
While zero-shot kinematics may be achieved at least in some cases through overly convoluted engineering, we advocate for a method that achieves zero-shot kinematics by design. 

Intuitively, the core requirement is that the model must accommodate kinematic chains with an arbitrary number of joints. We formalize this as a constraint on the learnable parameters $\Theta$.
\newtheorem{observation}[theorem]{Lemma}
\begin{observation}\label{lem:joint_independenec}
To handle arbitrarily sized kinematic chains $\Kkin{}$, the learned parameters $\Theta$ of a model cannot be dependent on the number of joints \ie the cardinality of the input $\Kkin{}$:
\begin{equation}
   \frac{d|\Theta|}{d\numjoints} =0 \qquad \ie \qquad |\Theta| \in O(1)
\end{equation}
\end{observation}
Indeed, in the trivial case $\Weights\in \mathbb{R}^{\numjoints \times \numjoints}$, the weights do not generalize to any new $|\Kkin{'}|>\numjoints$. 
And generally, allocating independent parameters $\Weights_j$ for each joint $j \in 1\, \dots \, \numjoints$ lets the parametrization $\Theta$ grow with $\numjoints$ and change whenever the kinematic chain changes, violating the Lemma.
Yet previous works in both deterministic \cite{zhong2022spatio, li2020dynamic, mao2019learning} and stochastic HMP\cite{suncomusion, salzmann2022motron, curreli2025nonisotropic} intentionally learn joint-dependent weights to extract strong dataset-~\cite{zhong2022spatio} and kinematics-specific~\cite{salzmann2022motron, curreli2025nonisotropic} priors, and thus cannot cannot satisfy Lemma 1 (we prove this by counterexample for each architecture class in \cref{app:pe:observation_counterexamples}).

We recognize that this requirement is naturally fulfilled by models $\netfunct$ that are permutation equivariant   $\netfunct(\perm \pastmotion{}{}) = \perm \netfunct(\pastmotion{}{})$ with respect to arbitrary joint reordering $\perm \in \mathbb{R}^{\numjoints \times \numjoints}$.
\begin{theorem}
\label{thm:peq_sufficient}
Let $o(\mathbf{X}) = \mathbf{W}\mathbf{X}\mathbf{G}$ be a general network operation for feature extraction on an input $\mathbf{X} \in \mathbb{R}^{J \times F}$. If $o(\mathbf{X})$ is permutation equivariant under joint reordering, i.e.
$\mathbf{P}\,o(\mathbf{X}) = o(\mathbf{P}\mathbf{X})$ for any permutation
$\mathbf{P} \in \mathbb{R}^{J \times J}$, then the number of learned parameters $|\Theta|$ is constant in $J$,
and Lemma~\ref{lem:joint_independenec} is satisfied.
\end{theorem}
In other words, weights must be shared across all joints, and the model must behave consistently under any reordering of these instances. Permutation equivariance is therefore sufficient, though not necessary, for kinematics-agnosticism. We report the full proof in \cref{app:pe:proof_observation}. 
Since equivariance is preserved under composition, a network $f$ built entirely from equivariant operations $o$ is end-to-end permutation equivariant. 
With this motivation, we decide to implement our kinematics-agnostic solution as a permutation equivariant model. 
While we already mentioned that previous work do not fulfill the Lemma, we also prove numerically and mathematically in \cref{app:pe:shmp_works} that their architectural designs are not equivariant. We present an intuitive high level explanation on why this is the case in \cref{app:pe:reasoning} and in the next section. 

\subsection{\methodname}
\paragraph{Overview.} 
Based on the previously presented findings on permutation equivariance fulfilling the premise of a kinematics-agnostic model, we implement \methodname~as an equivariant latent diffusion model (\cref{fig:method}). We design a novel end-to-end equivariant framework consisting of (i) an autoencoder mapping motion sequences $\motion$ to and from the latent space $\latentvar{} \in \mathbb{R}^{ \numjoints \times L}$, and (ii) a denoiser  that predicts future motions in the latent domain $\latentvar{\boldsymbol{\theta}}$ conditioned on the embedding $\latentvar{past} = \encnet(\pastmotion{}{})$ of the input past $\pastmotion{}{}$.

\paragraph{Equivariant Latent Diffusion}
The generative process of diffusion models~\cite{ho2020denoising} is known to be computationally expensive in input space~\cite{dhariwal2021diffusion,patterson2021carbon}. To gain in efficiency we operate in a lower-dimensional latent space \cite{curreli2025nonisotropic, barquero2023belfusion} and opt for latent diffusion models (LDM) \cite{rombach2022highresolution}. 
While previous approaches learn a mapping $\netfunct_{\Kkin{}}(\pastmotion{}{}) = \predmotion{}{}$ for a fixed skeletal kinematics $\Kkin{}$, we support operations on different kinematics out-of-the-box, by making the connectivity $\adj$ of the kinematics $\Kkin{}$ an explicit input to the model.
We thus design a novel framework architecture  that is end-to-end formally permutation equivariant w.r.t. its inputs: 
\begin{equation}
\netfunct(\pastmotion{}{}, \adj) = \predmotion{}{}\,,\quad \text{with}  \quad \netfunct(\perm \pastmotion{}{}, \perm \adj \perm^\top) = \perm \netfunct(\pastmotion{}{}, \adj).
\end{equation}
Equivariance in LDMs requires an equivariant denoiser and a latent space that preserves input permutations~\cite{lin2025equivariant, thiede2020general, wad2022equivariance}.Although this space can be learned non-deterministically (e.g., via VAEs \cite{rombach2022highresolution}), we adopt a deterministic approach to improve training stability \cite{yao2025reconstruction}.
At inference, new latent variables are sampled from a univariate Gaussian distribution. Since this sampling is i.i.d., sample-wise equivariance is guaranteed only in a deterministic setting where the noise is fixed and permuted accordingly. Indeed, in generative models, permutation equivariance holds at a distribution level. We provide a detailed discussion in \cref{app:pe:diffusion}. Differently from previous LDM in SHMP\cite{curreli2025nonisotropic, barquero2023belfusion}, we implement the autoencoder as a transformer rather than a recurrent network, gaining in inference speed.
 We follow the training paradigm of LDM \cite{rombach2022highresolution} adapted to SHMP \cite{barquero2023belfusion, curreli2025nonisotropic}. Further details and equations in \cref{app:method_details}.

\paragraph{\methodname's Architecture.}
\label{sec:method:arch}
In looking for a permutation equivariant (PEQ) architecture, the reader may already be thinking that Graph Convolution Networks (GCN)\cite{kipf2016semi} naturally fulfill the requirement\cite{keriven2019universal}. However, while many SHMP models are based on GCN~\cite{barquero2023belfusion, curreli2025nonisotropic, li2020dynamic, suncomusion, salzmann2022motron}, no one of them enjoys PEQ. The reason is that independently extracted features are not aggregated according to the connectivity of the graph\cite{kipf2016semi}, but according to learned weights $\Weights \in \mathbb{R}^{\numjoints\times \numjoints}$ that explicitly depend on the number and position of joints \cite{curreli2025nonisotropic, li2020dynamic, ramesh2022hierarchical}. Such practice in deterministic \cite{zhong2022spatio} and  SHMP\cite{suncomusion, salzmann2022motron, curreli2025nonisotropic} showed advantages in leveraging dataset-specific priors (\eg action \cite{zhong2022spatio} or joint-specific \cite{salzmann2022motron, curreli2025nonisotropic}). However, such implicit bias is detrimental in our case. We advocate for a kinematics-agnostic  formulation aligning with \cite{kipf2016semi}. Specifically, for an initial input $\feat \in \mathbb{R}^{\numjoints \times 3T}$
, we implement a convolution $c$ as: 
\begin{equation}
\label{eq:graph_conv}
c( \feat, \adj)
=
 \Weights\feat\mathbf{G_1}
+ \feat\mathbf{G_0}
+ \boldsymbol{b}\,,\quad \text{with}  \quad \Weights = \mathbf{D}^{-1} \adj\,
\end{equation}
where the weights $\mathbf{G_0},\mathbf{G_1}\!\in\!\mathbb{R}^{F_\text{i}\times F_\text{o}}$ learn to  extract features for a joint itself or its neighbours respectively, $\boldsymbol{b}\!\in\!\mathbb{R}^{F_\text{o}}$ is learned, $\mathbf{D}$ the normalizing degree matrix\cite{kipf2016semi}  of $\adj$, $F_\text{i}$ and $F_\text{o}$ are the input and output feature dimensions. This operation fulfills equivariance not only with respect to a vector $\pastmotion{}{}$, but also with respect to the input matrix $\adj$~\cite{thiede2020general}, which excludes otherwise compelling layers~\cite{ying2021transformers}.
We want to consider the skeleton graph on a global scale in addition to the local scale of \cref{eq:graph_conv}, and do so via  Graph Attention (GAT) \cite{velivckovic2017graph} or multi-head self-attention \cite{vaswani2017attention} with joints as token dimension. 
\begin{equation}
\label{eq:attention}
    \text{Att}(\feat, \adj)_h = \text{softmax}\left(c_Q(\feat, \adj)c_K(\feat, \adj)^\top/\sqrt{F_o}\right)c_V(\feat, \adj)
\end{equation}
Particularly, we employ the operation in \cref{eq:graph_conv} to compute $\mathbf{Q}$, $\mathbf{K}$, and $\mathbf{V}$ for attention on each of the $h$ heads. Building on these operations, we design a novel end-to-end equivariant architecture.  Noticeably, by definition, attention layers~\cite{vaswani2017attention} without positional encodings are PEQ along their token dimension. But in transformer-based SHMP approaches~\cite{suncomusion, chen2023humanmac, dang2022diverse, yuan2020dlow, barquero2023belfusion, mao2021generating, walker2017pose}, or approaches that treat joints as features~\cite{chen2023humanmac, li2021skeleton, saadatnejad2023generic}, tokens are reserved for the time dimension instead of the joints (against Lemma \ref{lem:joint_independenec}).
We provide mathematical proof that our architecture is end-to-end equivariant in \cref{app:pe_proof_equifusion}, together with experimental results on equivariance for the baselines. To the best of our knowledge, an end-to-end PEQ architecture over joint orderings has not been demonstrated before for SHMP.

\begin{figure*}[t]
  \begin{subfigure}{0.49\linewidth}
      \includegraphics[trim=0cm 0.3cm 0cm 0cm, clip,width=\linewidth]{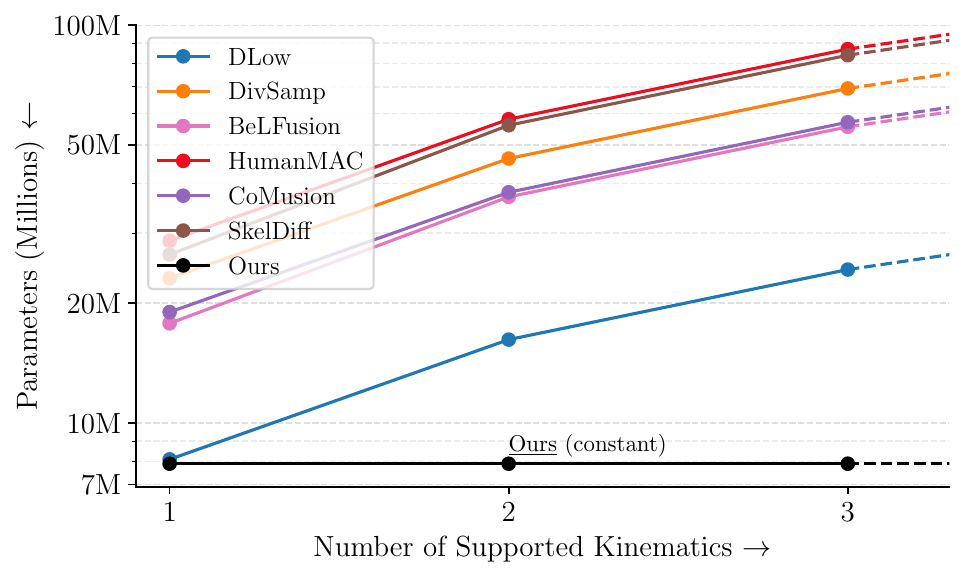} 
  \end{subfigure}
  \begin{subfigure}{0.49\linewidth}
      \includegraphics[trim=0cm 0.3cm 0.0cm 0.0cm, clip,width=\linewidth]{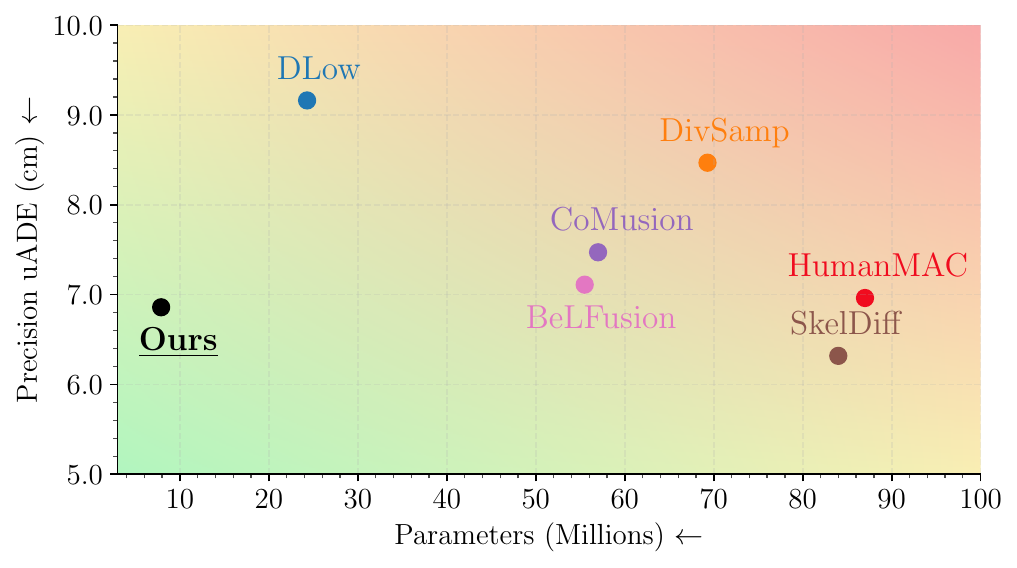}    
  \end{subfigure}
  \caption{
   \textbf{State-of-the-art with 75\% fewer parameters}: (left) baselines scale with the number of supported datasets i.e. kinematics, we do not; (right) we achieve SOTA precision, averaged over three datasets\cite{Ionescu2014, mahmood2019amass, ma2024nymeria}, with a single model instance.
  }
  \label{fig:plots_params}
  \vspace{-0.4cm}
\end{figure*}

\paragraph{Additional Advantages.} While  equivariance may be learned via extensive augmentation (as for molecule generation\cite{abramson2024accurate, wang2024understanding}), we chose to fulfill it mathematically by design, reducing model and learning complexity  \cite{lyle2020benefits, bietti2021sample}: on just a \underline{single} dataset\cite{Ionescu2014}, our kinematics-agnostic approach achieves state-of-the-art results with 75\% fewer parameters than the latest baseline SkelDiff\cite{curreli2025nonisotropic}, requiring around half of the training and inference time (see \cref{fig:plots_params} and discussion in \cref{sec:exp:analysis}). 
Furthermore, our equivariance property allows us to train natively on multiple datasets $\mathbb{K}_{\text{train}} = \{\mathcal{K}^{A}, \mathcal{K}^{N}\}$ leveraging for the first time the two largest SHMP datasets at once, AMASS and Nymeria.

\subsection{Directions as Motion Parametrization}

Motion representation is a critical choice in SHMP, especially for our aims of ensuring compatibility across disparate datasets and motion formats. Despite its importance, this remains underinvestigated in SHMP literature, where 3D joint absolute coordinates~(\cref{sec:related_works:motion_param}) remain the de facto standard.
While flexible, this representation suffers from limb stretching and physically inconsistencies~\cite{curreli2025nonisotropic, dang2022diverse}. Conversely, joints' rotation angles relative to the kinematic chain parent(as in SMPL\cite{SMPL2015}) preserve the body structure by design, but require a canonical pose, which is not always available. We propose a robust middle ground by representing the joints as the relative direction vector w.r.t. the parent joint, \ie $\dir{}{t}{} \in \mathbb{R}^{\numbones \times 3}$, where each limb $i$ is defined by the vector between a joint $i$ and its parent.
\begin{equation}
\dir{}{t}{i} = \joints_t^{i} - \joints_t^{\mathrm{parent}(i)}, \quad i = 1, \dots, \numbones\, .
\end{equation}
We report the details in \cref{app:parametrization}. This parametrization has several advantages: 1) it is easily derived from both positions and angles pose representations, facilitating cross-dataset training; 2) it eliminates limb stretching by rescaling relative distances between joints based on past observations at inference time; 3) it remains numerically stable during training \cite{bie2022hit} without singularities typical of rotation spaces~\cite{ salzmann2022motron, geist2024learning}. We use the average limb length of the training data as a heuristic for generating missing limbs.
Empirical results in \cref{tab:directions} demonstrate that this representation improves realism and diversity metrics also for other SHMP methods, without requiring any change to  existing SHMP architectures.

\section{Experiments}

\subsection{Experimental Settings}
\label{subsec:experimental_settings}
\paragraph{Datasets.}
We follow the evaluation settings of \cite{barquero2023belfusion, suncomusion, curreli2025nonisotropic, chen2023humanmac}, and include an additional dataset\cite{ma2024nymeria}: in addition to the protocols involving the highly diverse AMASS (A) dataset \cite{mahmood2019amass}, and the widely employed, but involving only 7 subjects, H36M (H)\cite{Ionescu2014}, we also include Nymeria (N)\cite{ma2024nymeria}, after adapting it to SHMP. Details of this process are reported in \cref{app:exp_setting:nymeria}, but overall the final size is comparable to AMASS and the quality of the motions is less dynamic. While the notation $\Kkin{}$ denotes exclusively kinematics, the letters A,H,N denote training data, automatically implying the corresponding kinematics. See \cref{sec:method:problem_def} and \cref{sec:rel_works:skeletons} for details on kinematics. We test models trained on A for the zero-shot motion scenario of MoYoga ($\Kkin{A}$)\cite{tripathi2023ipman}. 
\paragraph{Metrics.}
\label{metrics}
Conventionally employed metrics \cite{yuan2020dlow, barquero2023belfusion, curreli2025nonisotropic} can be categorized into precision, diversity, realism, and body realism (see \cref{app:exp_setting:metrics} for extensive definitions). %
Current metrics ADE, FDE, APD cannot be used to compare methods across datasets because they are dependent on the number of joints and cannot be measured in meters i.e. they include a cofactor $\sim\sqrt{\numjoints}$. Hence we introduce corresponding revised, unified versions  \textit{u}ADE, \textit{u}FDE, \textit{u}APD  measured in centimeters (\textit{u}ADE, \textit{u}FDE with their multimodal counterpart) or meters (\textit{u}APD). Note that \textit{u}ADE is mathematically equivalent to the mean per-joint projection error (MPJPE), widely employed in other vision tasks. Since they differ by just a cofactor, unified metrics rank identical to conventional metrics, which are provided for every table in the appendix.

\vspace{-0.1cm}
\paragraph{Baselines.} SOTA baselines \cite{curreli2025nonisotropic, chen2023humanmac, barquero2023belfusion, suncomusion, yuan2020dlow, walker2017pose, dang2022diverse, mao2021generating} do not support novel kinematics out-of-the-box, while we do so by design. To enable evaluation in a zero-shot kinematics setting for baselines, as described in \cref{sec:rel_works:retargeting}, we perform kinematics conversions following  Holden \etal established by \cite{chen2023humanmac}. We include the ZeroVelocity baseline (ZeroVel), a competitive algorithmic baseline that repeats the last frame of the past observation for all future frames\cite{yuan2020dlow}.

\begin{table}[t]	
\centering
\scriptsize 
\setlength{\tabcolsep}{0.9pt}
\caption{
\textbf{Evaluation of zero-shot kinematics on H36M ($\Kkin{H}$)\cite{Ionescu2014}}. We support inference on novel kinematics out-of-the-box, while previous approaches require additional kinematics conversion \cite{chen2023humanmac, holden2016deep}.  Baselines are trained on AMASS~\cite{mahmood2019amass} (A, $\Kkin{A}$). We are the first to natively support multiple kinematics and thus present a model trained additionally on Nymeria~\cite{ma2024nymeria} (N, $\Kkin{N}$).  The best results are highlighted in \textbf{bold}, second-best are \underline{underlined}. Conventional SHMP metrics rank identically, see \cref{tab:crossdataset_trainAMASSevalH36M_old_metrics}.  
}
\begin{tabular}{H l HcH   rrr rrH r rr  rrHH}
\toprule 
 & &\multicolumn{3}{c}{}  &  \multicolumn{3}{c}{Precision $\downarrow$} &  \multicolumn{3}{c}{MM GT $\downarrow$} & \multicolumn{1}{c}{Div $\uparrow$} & \multicolumn{2}{c}{Realism $\downarrow$} & \multicolumn{4}{c}{Body Real $\downarrow$}\\
 \cmidrule(lr){6-8} \cmidrule(lr){9-11} \cmidrule(lr){12-12}  \cmidrule(lr){13-14}
 \cmidrule(lr){15-18}
& Units &  & \multirow{2}{*}{$\Kkin{}$ } & – & \SI{}{\centi\metre} & \SI{}{\centi\metre} & deg\textdegree & \SI{}{\centi\metre} & \SI{}{\centi\metre} & – & \SI{}{\metre} & – & – & \SI{}{\percent} & \SI{}{\percent} & \SI{}{\percent} & \SI{}{\percent} \\
& Method & \multicolumn{1}{c}{} & new & $\mathrm{RT}$ & \textit{u}ADE & \textit{u}FDE & MAE & \textit{u}MMA & \textit{u}MMF & APDE & \textit{u}APD & CMD & FID & str & jit & str & jit\\ %
\midrule
-&ZeroVel&-&\cmark&$-$ & 11.77 & 17.88 & 6.753 & 13.74 & 18.56 & 8.085 & 0.000 & 22.822 & - &  0.00 & 0.00 & 0.00 & 0.00 \\
\multirow{4}{*}{VAE}&{TPK~\cite{walker2017pose}+\cite{holden2016deep}}&{A}&{\xmark}&$\mathrm{RT}_\mathrm{I/O}$ & 13.81 & 16.13 & 22.276 & 14.60 & 16.22 & \bfseries{1.968} & 1.469 & 10.051 & 3.773 & 19.55 & 0.46 & 22.21 & 0.73 \\
-&{DLow~\cite{yuan2020dlow}+\cite{holden2016deep}}&{A}&{\xmark}&$\mathrm{RT}_\mathrm{I/O}$ & 12.71 & 14.80 & 21.887 & 13.60 & 14.97 & \underline{2.060} & 2.060 & 9.204 & 2.875 & 20.26 & 0.53 & 23.47 & 0.82 \\
-&{GSPS~\cite{mao2021generating}+\cite{holden2016deep}}&{A}&{\xmark}&$\mathrm{RT}_\mathrm{I/O}$ & 9.29 & 11.91 & 8.107 & 10.85 & 12.35 & 2.373 & 2.069 & 7.409 & 1.735 & 11.51 & 0.38 & 14.00 & 0.49 \\
-&{DivSamp~\cite{dang2022diverse}+\cite{holden2016deep}}&{A}&{\xmark}&$\mathrm{RT}_\mathrm{I/O}$ & 9.27 & 12.61 & 8.374 & 11.42 & 13.27 & 10.510 & \bfseries{4.210} & 47.783 & 5.629 & 18.47 & 1.01 & 24.51 & 1.39 \\
\multirow{3}{*}{DM}&{BeLFusion~\cite{barquero2023belfusion}+\cite{holden2016deep}}&{A}&{\xmark}&$\mathrm{RT}_\mathrm{I/O}$ & 9.24 & 11.62 & 8.200 & 10.93 & 12.16 & 2.284 & 1.305 & 8.031 & 1.195 & 9.81 & 0.34 & 12.16 & 0.46 \\
-&{CoMusion~\cite{suncomusion}+\cite{holden2016deep}}&{A}&{\xmark}&$\mathrm{RT}_\mathrm{I/O}$ & 10.07 & 12.15 & 21.066 & 12.49 & 12.96 & 2.370 & \underline{2.070} & 8.587 & 1.426 & 15.98 & 0.51 & 17.57 & 0.68 \\
-&SkelDiff~\cite{curreli2025nonisotropic}+\cite{holden2016deep}&A&\xmark&$\mathrm{RT}_\mathrm{I/O}$ & 10.81 & 14.99 & 14.947 & 12.75 & 15.42 & 2.995 & 0.992 & 7.616 & 5.252 & 11.25 & 0.28 & 12.86 & 0.39 \\
\midrule
\midrule
\multirow{2}{*}{DM}&\methodname(A)&A&\cmark&- & \underline{7.86} & \underline{10.47} & \underline{5.861} & \underline{10.66} & \underline{11.61} & 2.248 & 1.973 & \bfseries{7.061} & \underline{0.691} & \bfseries{0.00} & \bfseries{0.00} & \bfseries{0.00} & \bfseries{0.00} \\
-&\underline{\methodname(A+N)}&A+N&\cmark&- & \bfseries{7.71} & \bfseries{10.21} & \bfseries{5.683} & \bfseries{10.58} & \bfseries{11.36} & 2.597 & 1.797 & \underline{7.349} & \bfseries{0.504} & \bfseries{0.00} & \bfseries{0.00} & \bfseries{0.00} & \bfseries{0.00} \\
\bottomrule
\end{tabular}

\label{tab:crossdataset_trainAMASSevalH36M}
\vspace{-0.3cm}
\end{table}

\vspace{-0.2cm}
\subsection{Comparison}

\begin{figure*}[t]
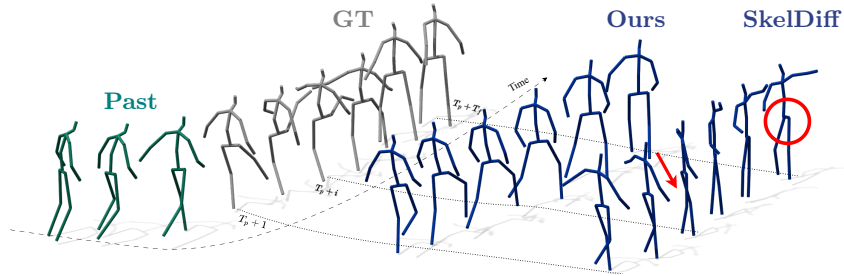

  \centering

	\begin{overpic}[trim=0cm 9cm 0cm 12cm, clip,width=0.99\textwidth, tics=5]{fig/images/qualitative_retargeting.pdf}
   \put(15,18){\textbf{\color{teal}Past}}
   \put(40,27){\textbf{\color{gray}GT}}
   \put(70,27){\color{bluepred2}\textbf{Ours}}
   \put(85,27){\color{bluepred2}\textbf{SkelDiff}} %
\end{overpic}
  
  \vspace{-0.1cm}
\caption{\textbf{Qualitatives for zero-shot kinematics on H36M($\Kkin{H}$)~\cite{Ionescu2014}}. We report the prediction closest to the GT for our method and SkelDiff\cite{curreli2025nonisotropic}, which does not support novel skeletons natively and is thus combined with a  retargeting procedure \cite{holden2016deep}. While this challenging dynamic kick is not reproduced by either works, our prediction is realistic and coherent with the observation. SkelDiff is unable to generate a semantically close motion and cannot recover from the degradation introduced by retargeting.
          }
  \label{fig:qualitative}
  \vspace{-0.4cm}
\end{figure*}

\paragraph{Zero-Shot full-body Kinematics}. In the main body of this paper, we concentrate on the realistic case that the SHMP model is trained on the kinematic $\Kkin{A}$ corresponding to the dataset with the largest data amount available, AMASS\cite{mahmood2019amass}, and inference is performed on a skeleton kinematics $\Kkin{H}$ with fewer joints and  a smaller dataset, H36M\cite{Ionescu2014}. This is the most advantageous setting for a SHMP model, as it is trained with a stronger prior (we discuss the more challenging, reverse scenario in \cref{app:exp:zero-shot-amass}). We report quantitative results in \cref{tab:crossdataset_trainAMASSevalH36M}.  Current SOTA approaches do not support this zero-shot kinematics scenario on $\Kkin{H}$ out-of-the-box, and the input motion must first be converted via retargeting \cite{holden2016deep} to the kinematics supported by the model (see discussion in \cref{sec:rel_works:retargeting}). Isolating the retargeting error from the baseline error completely is not possible, but via triangle inequality we estimate an upper bound. We see that baselines can recover from input retargeting: SkelDiff's total error  10.81~cm is around half of the upper bound 22~cm (further analysis in \cref{app:exp:retargeting_scenarios_explained}).
Our model instead, \cref{tab:crossdataset_trainAMASSevalH36M}, can operate natively on any kinematics and achieves the best results, with an improvement up to 27\% for precision metrics, and 72\% for realism.  
As already discussed by previous works\cite{barquero2023belfusion, curreli2025nonisotropic}, evaluating the  multidimensional HMP problem is complex, and the numerous metrics are often complementary: a high diversity scores (APD) can be originated by unrealistic and ill-posed motions, as shown by the poor realism and body realism results of the VAE-based method DivSamp\cite{dang2022diverse}. When considering the latest diffusion method, SkelDiff\cite{curreli2025nonisotropic}, our generated motions are almost twice as diverse. Additionally, our motion representation as bone direction  delivers perfect body realism by definition, and we discuss it further in \cref{sec:exp:analysis}. A qualitative result can be seen in \cref{fig:qualitative}.
\vspace{-0.1cm}
\paragraph{Multi-dataset Training.} Our key ability to train on multiple kinematics translates into the ability to train on any dataset simultaneously. We leverage for the first time the two largest SHMP datasets, AMASS (A) and Nymeria (N), and report results for our method trained on the combination of both (A+N). As expected, training on an additional kinematic type  (i.e. $\Kkin{N}$) and more data unlocks better performance. Our gains in this cross-dataset setting do not come from increased data diversity alone: we see that only 2\% of our 29\%ADE improvement and 27\% of our 90\% FID  come from the multi-dataset training.  The only difference between \methodname(A), trained only on AMASS, and   \methodname(A+N) is the training data, while the architecture and hyperparameters remain the same. We find the fact remarkable, highlighting the need for methods that natively reason over any dataset and kinematics and opening future discussion on the impact of different motion distributions on training.

\paragraph{Scalability to Any Kinematics.} \cref{fig:plots_params} highlights the impact on scaling and efficiency of our method. The kinematics-agnostic property  grants us two advantages: 1) the number of training parameters does not increase with the number of joints $\numjoints$ of the training kinematics $\Kkin{}$ (\eg SkelDiff requires more parameters for AMASS compared to H36M as AMASS has more joints); 2) we scale constantly with the number of supported datasets or kinematics because a single instance of our method can handle all kinematics, full-body or partial, that currently exists or will be released in the future.  Instead, as shown on the left plot of \cref{fig:plots_params}, baselines require a new instance for each kinematics. Therefore, when comparing instances on a single dataset (AMASS), we are 75\% more compact (70\% on H36M) than the latest baseline SkelDiff\cite{curreli2025nonisotropic}, while when considering three datasets (right plot of \cref{fig:plots_params}), we are 90\% more compact. Our inference and training time are halved, despite training on multiple datasets (\cref{app:exp:efficiency}).

\vspace{-0.1cm}
\paragraph{Out-of-distribution Motions on MoYoga ($\Kkin{A}$).} In \cref{tab:moyo} we report quantitative evaluation on the MoYoga dataset~ \cite{tripathi2023ipman}, containing out-of-distribution motions recorded via MoCap. Since this dataset shares the same kinematics as the training dataset AMASS ($\Kkin{A}$), baselines allow testing without retargeting. Our architecture outperforms the most recent and competitive baseline, SkelDiff\cite{curreli2025nonisotropic}, showcasing the stronger generalization capability of our model and again the advantages of using a richer prior. This is best highlighted in the qualitative results provided on our webpage.
\vspace{-0.25cm}

\begin{table}[t]
  \hfill
  \begin{minipage}{0.48\linewidth}
    \centering
    \caption{Out-of-distribution MoYoga~ \cite{tripathi2023ipman} for models trained on AMASS~\cite{mahmood2019amass}. 
    }    
    \begin{adjustbox}{width=\textwidth}
    \begin{tabular}{l c   r r r r r r r r}
        \toprule 
        &&\multicolumn{7}{c}{ MoYoga ($\Kkin{A}$)} \\
        \cmidrule(lr){3-9}
        & \multirow{2}{*}{$\Kkin{}$ } & \multicolumn{3}{c}{Precision $\downarrow$ }   & \multicolumn{1}{c}{Div$\uparrow$} & \multicolumn{1}{c}{Real$\downarrow$} & \multicolumn{2}{c}{B. Real$\downarrow$}\\
          \cmidrule(lr){3-5} \cmidrule(lr){6-6} \cmidrule(lr){7-7}  \cmidrule(lr){8-9}
          
        Method &  train %
        &  
         ADE  & FDE  & MAE   & APD  & CMD  &  str  & jit \\ %
        \midrule
         ZeroVel &  -  & 0.709 & 1.187 & 7.954 & 0.000 & 20.333 & 0.00 & {0.00}  \\

        SkelDiff~\cite{curreli2025nonisotropic}& A  & {0.567} & {0.892} & {8.048}   & \bfseries{13.304} & {15.710} & {7.25} & {0.29} \\
        \midrule
        \midrule
        \methodname&A & \underline{0.501} & \bfseries{0.780} & \underline{6.982} & \underline{13.153} & \underline{11.961} &  \bfseries{0.00} & \textbf{0.00}  \\
         \methodname&A+N & \bfseries{0.492} & \underline{0.786} & \bfseries{6.596} & {12.467} & \bfseries{7.095}  & \underline{0.00} & \underline{0.00}  \\   
        \bottomrule
    \end{tabular}
    \end{adjustbox}

    \label{tab:moyo}

  \end{minipage}
  \hfill
  \begin{minipage}{0.48\linewidth}
    \centering
    \caption{Occlusion of a random limb (leg or arm) at inference on AMASS test set. 
    } 
    \vspace{-0.12cm}
    \scriptsize 
    \begin{adjustbox}{width=\textwidth}

    \begin{tabular}{l cH rrr r rr}
    \toprule 
    &&\multicolumn{7}{c}{ AMASS ($\Kkin{A}$)} \\
        \cmidrule(lr){3-9}
    \multicolumn{1}{c}{} & \multirow{2}{*}{$\Kkin{}$ } & & \multicolumn{3}{c}{Precision $\downarrow$}  & \multicolumn{1}{c}{Div $\uparrow$} & \multicolumn{2}{c}{B. Real $\downarrow$} \\
     \cmidrule(lr){4-6} \cmidrule(lr){7-7} \cmidrule(lr){8-9} 
    \multirow{1}{*}{Method} &\multicolumn{1}{c}{train} & \multirow{1}{*}{} &  ADE  & FDE    & MAE  & APD  &  str  & jit   \\ %
    \midrule
    SkelDiff~\cite{curreli2025nonisotropic} & A  & \xmark & - & - & - & - & - & - \\
    SkelDiff~\cite{curreli2025nonisotropic}+rp & A  & \xmark & 0.574 & 0.727 & \bfseries{6.996} & 8.890 & 8.15 & 0.27 \\
    SkelDiff~\cite{curreli2025nonisotropic}+sl & A  & \xmark & 0.567 & 0.683 & \underline{7.162} & \underline{9.274} & 5.64 & 0.23 \\
    \midrule
    \midrule
    \methodname&A& \cmark & \underline{0.553} & \underline{0.618} & 10.190 & \bfseries{10.152} & \underline{0.00} & \underline{0.00} \\
    \methodname&A+N& \cmark & \bfseries{0.499} & \bfseries{0.553} & 8.777 & 9.099 & \bfseries{0.00} & \bfseries{0.00} \\

    \bottomrule
    \end{tabular}
    \end{adjustbox}

    \label{tab:drop_limb}
  \end{minipage}
\vspace{-0.47cm}
\end{table}

\paragraph{Zero-Shot Partial Kinematics.} Performing SHMP with partial input skeletons in a zero-shot setting, without training for it specifically, constitutes an exciting opportunity opened by our method. We also find it particularly relevant for applications, as it allows, depending on the upstream acquisition (MoCap or video), to represent uncertainty or occlusions. \methodname~natively supports flexible kinematics, and so also this scenario. For quantitative evaluation,  we randomly remove full limbs (i.e., legs or arms consisting of three joints) from every input sequence in the AMASS test set and pass them to the networks (\cref{tab:drop_limb}). 
To let the most recent and competitive baselines operate in this case, we adopt two heuristics. In the first, we complete the missing information with the one from the rest pose (+rp), simulating a ``mask'' effect. In the second step, we replicate the motion observed from the symmetric counterpart (+sl), resulting in more realistic motions.
Despite the symmetric completion enhancing the diversity of predictions and improving body realism, we observe that precision is still lacking. Our model instead predicts a reasonable future regardless of the missing part. Also in this case, relying on a wider motion prior (A+N) helps overcoming the missing information. Qualitative results are presented on our website. Apart from its applicative relevance to face occlusions, we also foresee our flexibility handy for tackling marginalized categories, such as individuals with diverse body types and those with missing limbs.

\vspace{-0.15cm}
\paragraph{Conventional Single-kinematics SHMP.} 
\label{sec:exp:analysis} 
Beyond cross-kinematics, our approach achieves state-of-the-art competitive performance on the in-domain, same kinematics evaluation of conventional SHMP benchmarks, \ie. on the designed test set of each dataset, for the kinematics belonging to that dataset. Besides AMASS\cite{mahmood2019amass} in \cref{tab:amass_intra} and H36M\cite{Ionescu2014} in \cref{tab:h36m_intra}, we additionally train and evaluate the latest SHMP diffusion models on Nymeria in \cref{tab:nymeria_intra}. The precision \textit{u}ADE averaged over all three datasets is reported in \cref{fig:plots_params}. Remarkably, while different methods require manual tuning of hyperparameters for each dataset~\cite{chen2023humanmac, curreli2025nonisotropic, suncomusion, barquero2023belfusion}, our method employs only a single configuration.

\begin{table}[t]
  \centering
  \begin{minipage}{0.48\linewidth}
    \centering
      \caption{Ablations for methods trained on multiple datasets, AMASS($\Kkin{A}$) and Nymeria($\Kkin{N}$), tested on AMASS ($\Kkin{A}$) as usual for single-kinematics and on H36M ($\Kkin{H}$) for zero-shot kinematics. 
      }
    \scriptsize 
    \begin{adjustbox}{width=\textwidth}
    
    \begin{tabular}{lccc ccc}
    \toprule
           &\multicolumn{3}{c}{ AMASS ($\Kkin{A}$)}  &\multicolumn{3}{c}{ H36M ($\Kkin{H}$)} \\

 \cmidrule(lr){2-4}  \cmidrule(lr){5-7}
         Method & ADE$\downarrow$ & MAE $\downarrow$& CMD$\downarrow$ &  ADE$\downarrow$ & MAE $\downarrow$& CMD$\downarrow$\\

    \midrule
    Modified~\cite{curreli2025nonisotropic}&\bfseries{0.500} & 7.007 & \bfseries{19.233} & 0.670  & 12.065 &  31.461  \\
    Ours w/o Eq& 0.539 & {6.974} & 22.294 & 0.467 & 7.245 & 9.559  \\
    Ours & {0.512} & \underline{6.525} & {19.699} & \textbf{0.380}& \underline{5.685}  & \bfseries{8.086}   \\
    Ours on $\perm+\perm\epsilon$ & 0.512 & \underline{6.525} & 19.699  & \textbf{0.380} & \underline{5.685} & \bfseries{8.086} \\
    Ours on $\perm$ & 0.513 & 6.529 & 19.686 & \bfseries{0.380} & \bfseries{5.680} & \underline{8.088} \\
    Ours+$S$ & \underline{0.508} & \bfseries{6.456} & \underline{19.492} & 0.381 & 5.708 & 8.526 \\
    \bottomrule
    \end{tabular}
\label{tab:ablations}

    \end{adjustbox}

  \end{minipage}
  \hfill
  \begin{minipage}{0.48\linewidth}
    \centering

    \caption{Models trained on AMASS\cite{mahmood2019amass} with two different motion parametrization: the proposed bone directions $\dir{}{}{}$ or the conventional 3D positions $\motion$.  
    } 
    \label{tab:directions}

    \begin{adjustbox}{width=\textwidth}

    \begin{tabular}{l c H rrr r r rr}
    \toprule 
        &&\multicolumn{8}{c}{ AMASS ($\Kkin{A}$)} \\
    
    \multicolumn{1}{c}{} & \multicolumn{1}{c}{}& \multicolumn{1}{c}{}  & \multicolumn{3}{c}{Precision $\downarrow$}  & \multicolumn{1}{c}{Div $\uparrow$}& \multicolumn{1}{c}{Real $\downarrow$}  & \multicolumn{2}{c}{B. Real $\downarrow$} \\
     \cmidrule(lr){3-3} \cmidrule(lr){4-6} \cmidrule(lr){7-7} \cmidrule(lr){8-8} \cmidrule(lr){9-10}
    \multirow{1}{*}{}  & \multirow{1}{*}{Mot.} &\multicolumn{1}{c}{} &  ADE  & FDE    & MAE  & APD & CMD &  str  & jit   \\ %
    \midrule
    
    SkelDiff~\cite{curreli2025nonisotropic} & $\motion$ & A& \textbf{0.480} & \textbf{0.545} & \textbf{6.124} & 9.456 & {11.417} &   {3.15} &   {0.20}  \\
    SkelDiff~\cite{curreli2025nonisotropic} & $\dir{}{}{}$ & A & 0.496 & 0.546 & 6.193 & \textbf{9.960} & \textbf{9.143} & \textbf{0.00} & \textbf{0.00} \\
     \midrule
    Ours & $\motion$ & A & 0.501 & 0.561 & 6.551 & 8.348 & 13.963 & 3.58 & 0.27 \\
    Ours & $\dir{}{}{}$ & A & \bfseries{0.498} & \bfseries{0.559} & \bfseries{6.173} & \bfseries{8.413} & \bfseries{12.530} & \bfseries{0.00} & \bfseries{0.00} \\
    \bottomrule
    \end{tabular}
    \end{adjustbox}

  \end{minipage}
  \vspace{-0.37cm}
\end{table}

\vspace{-0.15cm}

\paragraph{Ablations.} %
We first validate our approach in \cref{tab:ablations}. We show that a model whose weights are learned in dependence of joints, despite being trained with data augmentation on multiple kinematics on the same amount of data (A+N), does achieve competitive performance on the training kinematics (A), but fails at zero-shot kinematics on H36M. This shows in current settings, data alone does not guarantee kinematics-agnostic models. See \cref{app:exp_settings:baselines} for how we adapt the most competitive baseline~\cite{curreli2025nonisotropic} to multidataset training. 
Breaking our equivariance property by inserting positional encoding (\textit{Ours w/o Eq}) leads to failure at zero-shot kinematics: equivariance guarantees kinematics-agnostic capabilities. We verify that our model remains equivariant under permutations of joints, adjacency, and noise (\textit{Ours on $\perm+\perm\epsilon$}). We empirically validate end-to-end equivariance under stochastic sampling (i.e. we permute joints and adjacency on the same noise realization and compare outputs) as (\textit{Ours on $\perm$}). 
Furthermore, providing explicit joint semantics (e.g., limb side or type) is non-trivial (permutation equivariance must hold) and yields no gains (\textit{Ours+$S$}); the motion’s temporal information alone suffices to distinguish limbs, as human degrees of freedom are invariant.
In \cref{tab:directions} we validate the choice of using bone directions as motion representation, showing its contribution to both our method and the baseline SkelDiff. In both cases, such a representation has a limited impact on precision and diversity, while leading to improved realism by 10\% and achieving perfectly consistent limb length by design.
Further metrics and more detailed ablations are reported in the appendix.

\section{Conclusion}

\paragraph{Limitations and Future Work}
We introduced a novel motion parameterization, bone directions, whose formulation guarantees perfect adherence to bone length constraints. However, this representation handles only missing joints which are leaves of the kinematic chain; it does not support, e.g., a missing elbow when the hand joint is present. While such occlusions can be represented naturally in the input adjacency matrix, extending bone directions to handle them remains open. 
Beyond this, several broader directions warrant exploration. Training on multiple datasets with differing motion distributions raises open questions about how data quality, particularly the prevalence of dynamic movements, affects performance at inference time in both in-domain and cross-dataset scenarios.  Finally, future work may integrate skeletons beyond human kinematics, investigating bipedal and quadrupedal animal motion.

\paragraph{Conclusions} We presented \methodname, a novel and fundamentally more generalizable approach to Stochastic 3D Human Motion Prediction (SHMP). By formulating the skeleton kinematics as an explicit input and designing an end-to-end permutation-equivariant architecture for a latent diffusion model, we successfully sever SHMP models' reliance on graph structures hard-coded at training time. This paradigm enables a single model to generalize zero-shot to unseen datasets and eliminates the need for expensive and inaccurate data retargeting.
Beyond resolving the generalization bottleneck, \methodname~achieves state-of-the-art performance with remarkably enhanced efficiency and supports novel tasks such as partial motion prediction and targeted limb generation. This represents a fundamental shift from kinematics-specific models toward a truly kinematics-agnostic, generalizable SHMP framework.

\paragraph{Acknowledgments}
This work was supported by the European Research Council (ERC) Advanced Grant SIMULACRON. Thanks to Maolin Gao and Felix Wimbauer for proofreading, Thomas Dagès for the detailed and constructive suggestions, Stefania Zunino and the CVG team for their unwavering support.

{
    \small
    \bibliographystyle{splncs04}
    \bibliography{main}
}

\clearpage
\addtocontents{toc}{\protect\setcounter{tocdepth}{2}}
\setcounter{tocdepth}{2}
\setcounter{page}{1}
\setcounter{linenumber}{911}

\addtocontents{toc}{\protect\setcounter{tocdepth}{2}} %
\makeatletter
\renewcommand*\l@author[2]{} %
\renewcommand*\l@title[2]{}  %
\makeatother
\tableofcontents

\appendix

\section{Extended Discussion on Related Works and Positioning}
\label{app:rel_works}
\subsection{Motion Retargeting}
\label{app:rel_works:retargeting}

In the following, we discuss retargeting approaches in detail and highlight why methods addressing isomorphic and homeomorphic kinematics do not apply to our SHMP case.

\paragraph{Isomorphic Kinematics.}
  If two kinematics differ only in the length of their bones and are otherwise identical, we are dealing with isomorphic graphs. This case represents for us different human subjects of the same HMP dataset, and in our notation, both skeletons belong to the same kinematics $\Kkin{}$ (\cref{sec:method:problem_def}. 
 In the case of isomorphic graphs, motion retargeting between humanoid characters for digital animations has been vastly investigated, moving first from expensive optimization pipelines \cite{gleicher1998retargetting, lee1999hierarchical, choi2000online, tak2005physically, feng2012automating} to learned approaches with \cite{jang2018variational, delhaisse2017transfer} or without \cite{villegas2018neural, lim2019pmnet, aberman2019learning, villegas2021contact} paired GT data, involving not only kinematics but also RGB \cite{aberman2019learning}, skinning information \cite{lim2019pmnet, musoni2021reposing, musoni2021functional}, or both \cite{villegas2021contact}. While this line of approaches achieves high precision, it is not relevant for us, because we are interested in transferring motion \underline{across} datasets, not within.

 \paragraph{Homeomorphic Kinematics.}
Other approaches \cite{aberman2020skeleton, hu2023pose, zhang2024semantics} deal with homeomorphic graphs, i.e. kinematics that share the same end-effectors and can be translated into each other by subdivision or merging of edges. Such case is not present among existing SHMP kinematics, as it suffices to day that end-effectors varies (H36M does have feet, FreeMan \cite{Wang2023freeman} has ears, etc.), but it would theoretically correspond to translating any kinematics $\Kkin{}$ to a primal coarse skeleton with a very low number of keypoints (probably 5) and consequently vast information loss.
Considering existing homeomorphic approaches, they often rely on information not available in SHMP, such as end-effector rotations \cite{hu2023pose} or rigged skeleton meshes such as SMPL. 

 \paragraph{Non-homeomorphic Kinematics.}
Instead, our work considers kinematics conversion between non-homeomorphic skeletons. Here we follow the retargeting approach from Holden \etal\cite{holden2016deep}, already established in SHMP by HumanMAC\cite{chen2023humanmac}. While more recent approaches for non-homeomorphic retargeting exist from the domain of robotics and character animation, they are not applicable to our case. These require joint rotations of end-effectors and reference T-poses \cite{lee2023same},  skinning \cite{liao2022skeleton},  meshes \cite{wang2023zero}, or both skinning and meshes ~\cite{jang2024geometry} - even without the skeleton itself \cite{liao2022skeleton, wang2023zero} - which is information not always available in SHMP. 
 Recent robotics approaches for non-homeomorphic humanoids achieve \cite{mourot2023humot} around 100mm or ca 90mm \cite{cao2025g} reconstruction error for a whole sequence, but their code is not publicly available. 
Other approaches focus on semantic motion transfer through common latent spaces between humanoid and four-legged animals, where, due to lack of GT, the error can only be measured in terms of fidelity \cite{kim2025moreflow, li2024walkthedog}. Overall, we see our line of work as orthogonal: we do not seek to benchmark retargeting approaches and find the most suitable one for SHMP, we aim to solve SHMP end-to-end.

\paragraph{Learned Human Priors.}
Since the very successful human parametrization SMPL, a wide line of works has developed, estimating SMPL parameters from images, single or multiview videos. This line of works also gave birth  to methods that learn pose priors from data \cite{sarandi2025neural, vposer2019smplx, kolotouros2019learning, rempe2021humor} and allow reprojection of noisy motion to a more realistic space. 
While such prior does not solve skeleton retargeting, it could, in theory, further refine the output of retargeting.  
However, these spaces are only designed to suit the SMPL parametrization, and often require direct parametrization over the SMPL parameters\cite{SMPL2015}, which may amount to several minutes (!) for a single motion sequence. Additionally, not all priors consider the temporal dimension of a motion, increasing yes the realism of human poses for a single timestep, but possibly decreasing temporal coherence between frames of the same motion. Furthermore, SMPL is only compatible with the AMASS kinematics, and not with others. For these computational and applicability reasons we do not further consider pose or motions priors among our retargeting possibilities.

\subsection{Stochastic 3D Human Motion Prediction (SHMP)}
\label{app:rel_works:shmp}

\begin{table}[t]\footnotesize	
\centering
\caption{\textbf{\methodname~ supports any input kinematics, including missing joints.} Current SHMP Lock acquired approaches cannot perform inference on unseen or partial skeletons out-of-the-box, and need to be piped with retargeting or completion pipelines that increase complexity and lower performance. } 
\begin{tabular}{l   c c c c c}
\toprule 

 \multirow{2}{*}{Method} & joint & new  & occluded & limbs  & motion  \\ 
 & perm & $\Kkin{}$ & limbs & gen & repr\\
 \midrule
Motron & \rmark  & \rmark & \rmark & \rmark & Quat \\
DLow & \rmark  & \rmark & \rmark & \rmark & Eucl  \\
BeLFusion &  \rmark  & \rmark & \rmark & \rmark & Eucl  \\
CoMusion& \rmark  & \rmark & \rmark & \rmark & DFT  \\
SkelDiff & \rmark  & \rmark & \rmark & \rmark & Eucl  \\
+retargeting & \gmark  & \ymark & \rmark & \rmark & //\\
+completion & \rmark  & \gmark & \ymark & \rmark & //\\
\midrule
\midrule
\methodname & \gmark & \gmark & \gmark & \gmark & bone dirs \\
\bottomrule
\end{tabular}

\label{tab:advantages}
\end{table}

Human Motion Prediction (HMP) aims to predict a future motion given a past observation motion. 
HMP methods have used various deep learning architectures, such as recurrent networks \cite{fragkiadaki2015recurrent, jain2016structural, martinez2017human, gui2018adversarial, pavllo2018quaternet, liu2019towards}, temporal convolutions \cite{li2018convolutional, medjaouri2022hr}, and more recently transformers \cite{aksan2021spatio, cai2020learning, martinez2021pose} and graph neural networks (GCN) \cite{li2020dynamic, mao2019learning, dang2021msr, li2021skeleton,adeli2021tripod, mao2020history}. While some previous works tackled the problem in a deterministic manner  \cite{cui2020learning, li2020dynamic, li2019actional} forecasting a single future, probabilistic or stochastic HMP (SHMP) aims to predict multiple futures. The reason behind this distinction is also that the different tasks tend to consider different prediction time horizons: deterministic HMP deals with rather short-term forecasting, while stochastic SHMP makes predictions up to $\SI{2}{\second}$ by observing $\SI{0.5}{\second}$. The longer the prediction timespan, the higher the possibilities of different semantic actions to take place and thus the necessity to model the problem as probabilistic. Our paper aims at SHMP, and the deterministic case is treated in detail in the next subsection \cref{app:rel_works:det_hmp}.

\paragraph{Limitations.}
So far, SHMP methods are trained and evaluated on each kinematics $\Kkin{i}$ independently. This yields multiple trained instances per method where training hyperparameters are carefully adapted to each dataset manually, which is a laborious procedure for method applications \cite{chen2023humanmac}. On top of these limitations, existing approaches cannot be tested out-of-the-box on novel kinematics $\Kkin{j}\,\text{with}\, j\neq i$ and need to be retrained from scratch. This case also includes motions where limbs are occluded or not present in the subject i.e. partial kinematics. In this case, the occluded limbs need to be completed first via heuristics before being fed to current approaches(\cref{tab:advantages}). Our work fills this gap. While approaches that naturally support multiple skeleton graphs have already been proposed for other computer vision tasks  \cite{sarandi2025neural, dabhi20243d, gat2025anytop, lee2023same, huang2025animaxanimatinginanimate3d, liu2025text} and are discussed in section \cref{app:rel_works:others},  no such approach exists for SHMP yet.  The reason is that all previous SHMP architectures learn kinematics-specific weights to extract stronger prior on the training data.  We elaborate on this mathematical aspect in \cref{app:pe:shmp_works}.

\paragraph{Closest Competitor Baselines.}
We will now specifically comment in detail on the most recent state-of-the-art SHMP approaches, which base the generative modeling on denoising diffusion models: BeLFusion\cite{barquero2023belfusion}, CoMusion \cite{suncomusion}, and SkelDiff \cite{curreli2025nonisotropic}. CoMusion employs a diffusion transformer in input space to generate coarse predictions and refines them with a GCN\cite{kipf2016semi}. CoMusion is an exponent of the line of HMP works that employ the Discrete Cosine Transform (DCT) on the temporal dimension of the input and treat the body joints as features. This line of work draws on many exponents from the field of deterministic HMP. Instead, BeLFusion and SkelDiff are latent diffusion models that downsample the inputs in a lower-dimensional latent space via recurrent GCN-based architectures. While BeLFusion requires three training stages and an additional network to embed the observation, SkelDiff uses the same encoder, based on Typed-Graph Convolutions \cite{salzmann2022motron}, trained to handle flexible input lengths for both the GT and the observation, and thus requiring only two training stages. We follow the same insight that temporal compression should happen similarly for both future and past,  and reduce the number of required networks by employing the same encoder to embed both conditioning past and training GT in latent space. We further design our autoencoder as a masked transformer autoencoder where tokens are the joint dimensions, instead as a recurrent GRU \cite{barquero2023belfusion, suncomusion}.  It is interesting that although transformers and GCN are meant to handle input with a flexible structure, none of the methods above support different skeleton kinematics.

\paragraph{Our Kinematics-Agnostic Solution.}
With this motivation, we propose to broaden the horizon of SHMP and investigate realistic scenarios of zero-shot inference on novel kinematics in SHMP, a capacity particularly relevant when dealing with motions extracted from videos \cite{park2020hmpo, roudsarabi2008solving, phu2025predicting, lohit2021recovering}. 
We present \methodname, the first SHMP method that is  kinematics-agnostic by design. \methodname~ can also handle occluded or missing joints out-of-the-box without preprocessing (see overview in \cref{tab:advantages}). This property is achieved by treating the motion kinematics as input and defining the model's weights independently of the body joints. We discover that such condition holds naturally if a model is permutation equivariant (PEQ) with respect to the joint order. Despite including permutation equivariant components such as GCN layers and transformer layers, none of the existing SHMP baselines is end-to-end permutation equivariant w.r.t joint ordering and we prove it in \cref{app:pe:shmp_works}. In other applications \cite{wimmer2023scale, zhou2023permutation}, PEQ has already been shown to outperform other networks when data is scarce \cite{sosnovik2021disco, zhu2022scaling}.  We thus fill the gap by implementing \methodname~ as a permutation equivariant latent diffusion model.

\subsection{Deterministic Approaches to 3D Human Motion Prediction}
\label{app:rel_works:det_hmp}
Deterministic HMP addresses a time horizon significantly shorter than ours (usually 0.4s instead of 2s) and does so without modeling a distribution over possible futures but by regressing a single deterministic future.
Occluded motions and training on multiple kinematics have already been investigated by deterministic HMP,  but to the best of our knowledge, only \underline{explicitly}. Meaning that training directly targets occlusions and completion via auxiliary tasks, losses, and networks \cite{cui2021towards, xu2023auxiliary} and does not involve considerations about equivariance, while in our case we deal with occlusion without having seen any occluded motion at training time. Other works \cite{sarandi2023learning}, do yes train on multiple skeletons, but do not support skeletons that were not seen during training at inference time. Instead, we do support novel kinematics at inference even when relying on just a single one during training.

\subsection{Beyond our Task: Human Pose Estimation, Text-to-Motion, Unconditional Generation, and more}
\label{app:rel_works:others}

A vast number of tasks in computer vision deal with the human body, but from quite different viewpoints. In this section, we contextualize some of these tasks to our task of Human Motion Prediction (HMP) and comment on the techniques employed to address  multiple kinematics. 

\paragraph{Human Pose Estimation.} One of the oldest fields dealing with humans is what we can consider the upstream pipeline to HMP \cite{park2020hmpo, roudsarabi2008solving, phu2025predicting, lohit2021recovering}: human pose estimation from images. This task takes as input an image, and delivers 2D or 3D locations of human body joints or keypoints. Such a set of keypoints for a single timestep is referred to as \emph{pose}. When human pose estimation is applied to a series of consecutive images i.e. a video, the obtained 3D keypoint sequence is a \emph{motion} i.e. a sequence of poses. This motion sequence id the input of HMP models. Naturally, by definition, this task does not usually consider aspects that are at the core of HMP: 1) the time dimension, and 2) consequentiality and causality between past and future. However, as images naturally include occlusions of human body parts, this field has long been interested in dealing with occlusions or multiple keypoint configurations. We comment here on the most relevant approaches to us: GFPose~\cite{ci2023gfpose}, PUMPS~\cite{mo2025pumps}, and 3D-LFM~\cite{dabhi20243d}.

GFPose \cite{ci2023gfpose} solves human pose estimation for multiple applications - including pose completion, denoising, and lifting - but handles occlusions only through ad-hoc, explicit training (random masking) and does not handle multiple full-body skeleton kinematics.
3D-LFM~\cite{dabhi20243d} is a foundation model for 2D-to-3D lifting of keypoints, is agnostic to input categories but does not mention equivariance. 
Very recently, the Arxiv work PUMPS\cite{mo2025pumps}  investigates skeleton-agnosticism in the scope of human motion completion by preprocessing motions and transforming them as pointcloud. They do yes address motion denoising and 2d-to-3d motion lifting, but they finetune for it instead of doing it zero-shot.

\paragraph{Text-to-Motion Generation.} 
Another task involving human motion is text-to-motion generation. The domain strongly differs from ours, as the goal is to generate motion with high fidelity to the input text, regardless of the input motion observation. For example, MotionDiffuse \cite{zhang2024motiondiffuse}, which trains on SMPL and thus considers only the AMASS kinematics. Additionally, semantic action labels are part of the input in this task, where this is generally not the case for HMP.  Some works attempt to separate motion from the skeleton. For example,  the  approach of Liu \etal\cite{liu2025text} that generates motions from text for any skeleton with a token-based VQ-VAE, while instead we compress deterministically over the whole time dimension in a unique global latent code. They do not rely on equivariance and do not discuss partialities or occlusions. Another approach, AnyTop \cite {gat2025anytop}, generates realistic animal motion from an input skeleton relying on joint classes and text descriptors. Occlusions are also not discussed, but if removing end-effector is possible, it would at least necessitate ad-hoc modeling of the input.  While these approaches are related, they cannot be translated to HMP in a straightforward manner. We do not want to apply a motion to a new skeleton (where there is a clear separation), but rather observe the past of a previously unseen skeleton and predict its future in the same format. Our model must also extract semantic information from that past in a format that is both unseen and non-textual.

\paragraph{Motion Search and Tokenization}
In the attempt to categorize or semantically represent motions to allow grouping, motion transfer, or motion search, we see some works discussing topology-agnosticism. A very recent Arxiv work\cite{xu2026necromancer} deals with different animal or character skeletons by leveraging text, meshes, and delivering a unique representation token for the whole motion regardless of the skeleton format. Precisely, their representation is kinematics invariant, not equivariant, a consequence of the differences in problem statement. Another work \cite {lee2023same} addresses motion search and classification with a disentangled latent space for retargeting and character animation with a GCN-based architecture. However, it requires training pairs that showcase the same motion but with different kinematics, which would be a significant drawback in our case since such pairs are not directly available.

\section{About Lemma \ref{lem:joint_independenec}: Weights Dependent on the Kinematics Cardinality}
\label{app:sec:observation_proof}

As we observe in the main paper body with Lemma \ref{lem:joint_independenec}, the weights $\Weights$ learned by a model cannot depend on the number of joints seen at training, if the model wants to generalize to arbitrary kinematic chains $\Kkin{}$. Before we define the notation to prove this equation, we stress here a significant challenge of the SHMP that influenced the property of equivariance w.r.t. joint dimension in previous works: dealing with 4D data.

\subsection{Dealing with 4D Data.} \label{app:pe:reasoning} The input $\pastmotion{}{}  \in \mathbb{R}^{\pastframes\times\numjoints\times3}$  to SHMP is multidimensional by definition, as it includes temporal and spatial information (4D). In addition to these two dimensions, the joint dimension $\numjoints$ represents an additional degree. Together with the XYZ 3D dimensions, it delivers spatial information and is thus often merged together~\cite{suncomusion, chen2023humanmac, dang2022diverse, yuan2020dlow, walker2017pose}, but semantically it can represent an additional orthogonal dimension. Choosing how to deal with these dimensions is a challenge intrinsic to SHMP.  
Extracting meaningful features for SHMP means combining successfully temporal, spatial and joint information. In the process, the motion may be reshaped  and the joint dimension fused with others implicitly fixing the joint set. Any operation of this kind effectively treats the joints as features\cite{suncomusion, chen2023humanmac, dang2022diverse, yuan2020dlow, barquero2023belfusion, mao2021generating, walker2017pose}. In terms of choosing a suitable architecture for the problem, this translates to, for example, having multiple dimensions available as token dimension for a transformer architecture. If the time dimension is chosen as the token dimension,  joints are consequently treated as features and thus not in an equivariant manner. This is the case for all previous transformer approaches\cite{suncomusion, chen2023humanmac}, where joints are treated as tokens just alternately in subcomponents ~\cite{curreli2025nonisotropic, suncomusion} or at intermediate stages \cite{wei2023human, dang2022diverse, mao2021generating}.

\subsection{Counterexample for Lemma \ref{lem:joint_independenec}}
\label{app:pe:observation_counterexamples}
For the sake of discussion, let us reshape an input motion $\motion{}  \in \mathbb{R}^{T\times\numjoints\times3}$ to $\pastmotion{}{}  \in \mathbb{R}^{\numjoints \times F}$. Here $F = T \cdot 3$, but in the following, for simplicity, we omit the XYZ dimension, and continue our discussion with $F = T$ without loss of generalization. 
For such a two-dimensional input, network operations for feature extraction can be expressed in terms of a basic  matrix multiplication $o()$ as follows:
\begin{equation}
\label{eq:lem:counterexample}
    o(\pastmotion{}{}) = \Weights \pastmotion{}{} \mathbf{G},
\end{equation}
where $\Weights \in \mathbb{R}^{m_1 \times \numjoints}$, and $\mathbf{G}\in \mathbb{R}^{F \times m_2}$, with $m_1, m_2$ arbitrary dimensions. 
If $\Weights$ is learned from data, we are then training with a kinematics $\Kkin{} \in \mathbb{K}_{\text{train}}$ of cardinality $\numjoints = |\Kkin{}|$ or with multiple kinematics $\Kkin{1}, \dots \Kkin{B} \in \mathbb{K}_{\text{train}}$ where the kinematics with the largest number of joints has cardinality $\numjoints = \max_{i \in \{1,\dots, B\}}{|\Kkin{i}|}$. When performing inference on a motion $\pastmotion{}{}^{'}$ having kinematics $\Kkin{'}\notin \mathbb{K}_{\text{train}}$
\begin{equation}
    o(\pastmotion{}{}^{'}) = \Weights \pastmotion{}{}^{'} \mathbf{G} \quad \text{is undefined if} \quad |\Kkin{'}| \neq \numjoints.
\end{equation}
To be precise, in our mathematical setting, multiplication with learned matrices can only take place \textit{from the right} i.e. joint-independent feature extraction, and not from the left.  %
We can indeed easily see that, in our case, right multiplications $r(\pastmotion{}{})$  are permutation equivariant
\begin{equation}
    \perm r(\pastmotion{}{}) = \perm (\pastmotion{}{} \mathbf{G}) = (\perm \pastmotion{}{}) \mathbf{G} = r(\perm \pastmotion{}{})\,,
\end{equation}
due to the associative property of matrix multiplication. In contrast, left multiplications $l(\pastmotion{}{})$ are not permutation equivariant 
\begin{equation}
    \perm l(\pastmotion{}{}) = \perm (\Weights \pastmotion{}{}) \neq \Weights (\perm \pastmotion{}{}) = l(\perm \pastmotion{}{})\,,
\end{equation}
because matrix multiplication is in general a non-commutative operation. 
\subsection{Counterexamples for Previous SMP Approaches.} In the following, we lead back mathematical operations employed by previous approaches to the previous equation \cref{eq:lem:counterexample}, which proves Lemma \ref{lem:joint_independenec} by counterexample. We categorize these operations by the architecture they stem from. Overall, we can categorize these approaches in two groups: Approaches that treat joints as features (1,3,4), and approaches that rely on graph convolutions but learn the aggregation (2,5). 
\begin{enumerate}
    \item \textbf{MLP or Linear Layer.} Employed by \cite{barquero2023belfusion, dang2022diverse}. In our notation, this equals treating the joints as features i.e. multiplication from the left:
    \begin{equation}
    o_{MLP}(\pastmotion{}{}^{'}) = \Weights \pastmotion{}{}^{'}\quad \text{is undefined if} \quad |\Kkin{'}| \neq \numjoints.
\end{equation}
    \item \textbf{Graph Convolutions with Learned Aggregation Matrix}. Employed by \cite{suncomusion, xu2024learning, dang2022diverse, mao2021generating}, where $\Weights$ is learned. In our notation, this is coincident to our main example, with an additional sum of joint-independent features $\mathbf{
    \tilde{F}}\in \mathbb{R}^{F\times m_2 }$.
    \begin{equation}
    o_{GC_{learned}}(\pastmotion{}{}^{'}) = \Weights \pastmotion{}{}^{'} \mathbf{G} + \mathbf{
    \tilde{F}} \quad \text{is undefined if} \quad |\Kkin{'}| \neq \numjoints.
\end{equation}
    \item \textbf{Recurrent Units (GRU, LSTM)}. Employed to extract temporal features in \cite{walker2017pose, yuan2020dlow, barquero2023belfusion}. Any gate of a recurrent network (i.e. input gate, forget gate, output gate, etc.) is applied to a single pose $\pastmotion{}{t}^{} \in \mathbb{R}^{\numjoints \times 1 }$ i.e. to each timestep $t$ of sequence, and relies eventually on a hidden state $\mathbf{H}_{t} \in \mathbb{R}^{m_1 \times  1}$. In our notation, with the hidden state weight $\mathbf{G}_{RNN}\in \mathbb{R}^{m_1 \times  m_1}$it is expressed as 
    \begin{equation}
    o_{RNN}(\pastmotion{}{t}^{'}) = \Weights \pastmotion{}{t}^{'} + \mathbf{G}_{RNN}\mathbf{H}_{t-1} \quad \text{is undefined if} \quad |\Kkin{'}| \neq \numjoints.
\end{equation}
    \item \textbf{Transformers with Time as Token Dimension and Joints as Features}. Employed by \cite{chen2023humanmac, suncomusion}. Since the query, key, and value matrices are computed via linear layers i.e. multiplication from the left (see case 1) with learned weight matrices $\Weights_Q$, $\Weights_K$, and $\Weights_V \in \mathbb{R}^{m_1 \times \numjoints}$, this case also does not comply with the Lemma. 
    \begin{equation}
    \begin{split}
    o_{Tranf}(\pastmotion{}{}^{'}) = & \text{softmax}\left(\sigma(\Weights_Q\pastmotion{}{}^{'})\sigma(\Weights_K\pastmotion{}{}^{'})^\top/\sqrt{F_o}\right)\sigma(\Weights_V\pastmotion{}{}^{'}) \\
    & \quad \text{is undefined if} \quad |\Kkin{'}| \neq \numjoints.
    \end{split}
\end{equation}
Here $\sigma()$ represents the non-linear activation function of choice. 
    \item \textbf{Typed-Graph Convolutions}. A popular choice in SHMP is learning typed weights $\mathbf{G}_{\tau(j)} \in \mathbb{R}^{ F \times m_2}$, where each joint is assigned to a type  $\tau(j) \in \mathbb{T}$ (shoulder, hip, elbow, etc.)\cite{curreli2025nonisotropic, salzmann2022motron} regardless of the body side (left or right). Here, the motion of a single joint is expressed as $\pastmotion{}{j} \in \mathbb{R}^{ F}$, and $\mathbb{T}_{train}$ is the set of joint types seen at training.
        \begin{equation}
    o_{GC_{typed}}(\pastmotion{}{}^{'}) = \Weights \begin{bmatrix}
 \pastmotion{}{\tau(0)}^{'} \mathbf{G}_0 \\
 \pastmotion{}{\tau(1)}^{'} \mathbf{G}_1 \\
\vdots
\end{bmatrix} \quad \text{is undefined if} \quad \tau (j^{'}) \notin  \mathbb{T}_{train}.
\end{equation}
Regardless of whether the weights $\Weights$ are learned, this paradigm does naturally not generalize to joint whose types have not been sen during training (\eg as whne training on H36M, that has no feet, and testing on kinematics that have feet \cref{tab:crossdataset_trainH36MevalAMASS}).
\end{enumerate}
As an interesting side note for future works, the widely employed Discrete Cosine Transform (DCT)\cite{suncomusion, chen2023humanmac}, does not contradict Lemma \ref{lem:joint_independenec} per se, as it acts only on the temporal domain. However, up to date, it is usually paired only with approaches that treats the joints as features. 

\section{On Permutation Equivariance}

\subsection{Mathematical Proof of Equivariance Fulfilling Lemma 1}
\label{app:pe:proof_observation}
Given an input $\pastmotion{}{} \in \mathbb{R}^{\numjoints \times F}$ and a permutation matrix $\perm \in \mathbb{R}^{\numjoints \times \numjoints}$ that reorders the joints, an operation $o(\pastmotion{}{}) = \Weights \pastmotion{}{}\mathbf{G}$ as described in \cref{app:sec:observation_proof} is permutation equivariant if:
\begin{equation}
    \perm o(\pastmotion{}{}) = o(\perm \pastmotion{}{})
\end{equation}
Here we show that an operation $o()$ that is permutation equivariant w.r.t. $\perm$ naturally fulfils Lemma \ref{lem:joint_independenec}.

Let us substitute the operation $o()$ into the equivariance definition:
\begin{equation}
\begin{split}
    \perm(\Weights\pastmotion{}{}\mathbf{G}) = & \Weights(\perm \pastmotion{}{})\mathbf{G} \\
    \perm\Weights\pastmotion{}{}\mathbf{G} = & \Weights\perm \pastmotion{}{}\mathbf{G} \\
    \perm\Weights = & \Weights\perm  \\
    \end{split}
\end{equation}
Thus, equivariance is fulfilled if it holds $\perm\Weights =  \Weights\perm$.
Let us consider a general case where the permutation matrix $\perm$ swaps joints $i$ and $j$, but not joint $j$ i.e.

\begin{equation}
\perm_{kl} = 
\begin{cases} 
1 & \text{if } (k,l) \in \{(i,j), (j,i)\} \\
1 & \text{if } k=l \text{ and } k \notin \{i, j\} \\
0 & \text{otherwise}
\end{cases}
\end{equation}
We now consider different entries of the equation $\perm\Weights =  \Weights\perm$ in dependence of the indeces $i,j,k$.
\begin{enumerate}
    \item \textbf{Swapped Indeces ($i,j$).} We first consider the entry ($i,j$) where joints have been swapped.
    \begin{equation}
    \label{eq:pe_proof_obs:Indeces_ij}
    \begin{split}
    (\perm\Weights)_{ij} = & (\Weights\perm)_{ij}  \\
    \sum_m^\numjoints \perm_{im} \Weights_{mj} = & \sum_m^\numjoints \Weights_{im} \perm_{mj}\\
    \perm_{ij} \Weights_{jj}  + \sum_{m\neq j}^\numjoints \perm_{im} \Weights_{mj}= &  \Weights_{ii} \perm_{ij} + \sum_{m\neq i}^\numjoints \Weights_{im} \perm_{mj}\\
    1 \cdot \Weights_{jj} + 0  = & \Weights_{ii} \cdot 1 + 0 \\
    \Weights_{jj}  = & \Weights_{ii} \\
    \end{split}
    \end{equation}
\item
\textbf{Unswapped Indeces ($i,k$).} We now consider the entry ($i,k$) where joints have not been swapped.
    \begin{equation}
    \label{eq:pe_proof_obs:Indeces_ik}
    \begin{split}
    (\perm\Weights)_{ik} = & (\Weights\perm)_{ik}  \\
    \sum_m^\numjoints \perm_{im} \Weights_{mk} = & \sum_m^\numjoints \Weights_{im} \perm_{mk}\\
    \perm_{ij} \Weights_{jk}  + \sum_{m\neq j}^\numjoints \perm_{im} \Weights_{mk}= &  \Weights_{ik} \perm_{kk} + \sum_{m\neq k}^\numjoints \Weights_{im} \perm_{mk}\\
    \Weights_{jk}  = & \Weights_{ik} \\
    \end{split}
    \end{equation}
\item
\textbf{Unswapped Indeces ($j,k$).} We now consider the entry ($j,k$) where joints have not been swapped.
    \begin{equation}
    \label{eq:pe_proof_obs:Indeces_jk}
    \begin{split}
    (\perm\Weights)_{jk} = & (\Weights\perm)_{jk}  \\
    \sum_m^\numjoints \perm_{jm} \Weights_{mk} = & \sum_m^\numjoints \Weights_{jm} \perm_{mk}\\
    \perm_{ji} \Weights_{ik}  + \sum_{m\neq i}^\numjoints \perm_{jm} \Weights_{mk}= &  \Weights_{jk} \perm_{kk} + \sum_{m\neq k}^\numjoints \Weights_{jm} \perm_{mk}\\
    \Weights_{ik}  = & \Weights_{jk} \\
    \end{split}
    \end{equation}
\end{enumerate}
    From these equations we just obtained conditions for the weight matrix. According to the first equality \cref{eq:pe_proof_obs:Indeces_ij}, it follows that all diagonal entries must have the same value. For the off-diagonal elements, from the second equality \cref{eq:pe_proof_obs:Indeces_ik}  it follows that all elements of a column must be equal, and from the second equality \cref{eq:pe_proof_obs:Indeces_jk} it follows that all elements of a row must be equal. This means that all off-diagonal entries have the same value. 
    Therefore, every diagonal entry is represented by a scalar $\alpha$ and each off-diagonal entry by a scalar $\beta$:
    \begin{equation}
    \begin{split}
    \Weights_{ij} = & \alpha \mathbf{\delta}_{ij} + \beta (1-\mathbf{\delta}_{ij})\, , \quad \text{or} \\
    \Weights = & \alpha \identity + \beta \mathbf{1}\mathbf{1}^\top\, , \\
    \end{split}
    \end{equation}
where $\mathbf{\delta}$ is the Kronecker delta. 
With such learned weights, we formulate the permutation equivariant operation $o_{PEQ}()$ as
    \begin{equation}
    \label{eq:pe_proof_obs:o_pe}
    o_{PEQ}
    (\pastmotion{}{})_j = \alpha {\pastmotion{}{}}_j + \beta \cdot \sum_{m \neq j}^\numjoints {\pastmotion{}{}}_m
    \end{equation}
Let us thus show that such a permutation equivariant operation by design fulfills the Lemma \ref{lem:joint_independenec}: the learned weights are not dependent on the number of joints. First, the summation in \cref{eq:pe_proof_obs:o_pe} is invariant to the number of joints, which is a property of any sum operation. Second, the set of learnable parameters of this operation is 
    \begin{equation}
    \Theta = \{\alpha, \beta\}\, \qquad \text{and} \qquad |\Theta| = 2 
    \end{equation}
Here we see that $|\Theta| = 2 $ is not dependent of $\numjoints$. Indeed,
\begin{equation}
   \frac{d|\Theta|}{d\numjoints} = \frac{d(2)}{d\numjoints} =0  \implies |\Theta| \in O(1)
\end{equation}
Thus the number of parameters does not scale with $\numjoints$, but is constant:
\begin{equation}
   \forall \numjoints \in \mathbb{N}, \quad \text{dim}(\Theta) = \text{const.}
\end{equation}
The Lemma is thus naturally satisfied: equivariance is a sufficient condition, even if not a necessary one.
Instead, any learned matrix $\Weights$ that does not fulfill the equivariance constraint, has the set of learned parameters 
$\Theta_{non-PEQ} = \{w_{ij} \mid i, j \in \{1, \dots, \numjoints\}$, implying $|\Theta_{non-PEQ}| = J^2$ and that the number of learned parameters grows with the size of the kinematics $ |\Theta_{non-PEQ}| \in O(\numjoints^2)$, which contradicts the Lemma.

\subsection{Mathematical Proof of Equivariance for \methodname}
\label{app:pe_proof_equifusion}
A model is permutation equivariant with respect to a permutation matrix $\perm$ if all its operations are permutation equivariant. Here we prove mathematically that our graph convolution \cref{eq:graph_conv} and self-attention  \cref{eq:attention} operations are equivariant. 
\paragraph{Graph Convolution.}
For simplicity, we report again here \cref{eq:graph_conv}:
\begin{equation}
\label{eq:pe_conv:property}
c( \feat, \adj)
=
 \mathbf{D}^{-1} \adj\feat\mathbf{G_1}
+ \feat\mathbf{G_0}
+ \boldsymbol{b}\, \quad \text{with} \quad \mathbf{D}=\text{diag}(\adj\mathbf{1}) 
\end{equation}
We want to prove 
\begin{equation}
c( \perm \feat, \perm \adj \perm^\top)
= \perm c( \feat, \adj)
\end{equation}
and start from the left side:
\begin{equation}
\label{eq:pe_conv:proof}
\begin{split}
c( \perm \feat, \perm \adj \perm^\top)
= & \mathbf{D}_P ^{-1} \perm \adj \underbrace{\perm^\top\perm}_{\identity} \feat\mathbf{G_1}
+ \perm \feat\mathbf{G_0}
+ \boldsymbol{b} \\
= & \left(\perm \mathbf{D} \perm^\top\right)^{-1} \perm 
\adj \feat\mathbf{G_1}
+ \perm \feat\mathbf{G_0}
+ \boldsymbol{b}\\
 = & \left(\perm^\top\right)^{-1}  \mathbf{D}^{-1}\perm^{-1} \perm \adj \feat\mathbf{G_1}
+ \perm \feat\mathbf{G_0}
+ \boldsymbol{b}\\
 = &  \perm  \mathbf{D}^{-1}\underbrace{\perm^\top\perm}_{\identity}  \adj \feat\mathbf{G_1}
+ \perm \feat\mathbf{G_0}
+ \boldsymbol{b}\\
 = & 
 \perm \mathbf{D}^{-1} \adj \feat\mathbf{G_1}
+ \perm \feat\mathbf{G_0}
+ \boldsymbol{b}\\
 = & \perm \left( \mathbf{D}^{-1} \adj \feat\mathbf{G_1}
+ \feat\mathbf{G_0}
+ \boldsymbol{b}\right) \\
= & \perm c( \feat, \adj)
\end{split}
\end{equation}
where we used $\mathbf{D}_P =  \text{diag}(\perm \adj \perm^\top\mathbf{1}) = \perm \mathbf{D} \perm^\top$ in the second equality and $\perm \boldsymbol{b} = \boldsymbol{b}$ (since $\boldsymbol{b}\in\mathbb{R}^{1\times F_\text{o}}$ and $\perm \in \mathbb{R}^{\numjoints\times \numjoints}$) in the last, and $\perm^{-1} = \perm^ \top$ in general. 

\paragraph{Self-Attention}
Here we report the self-attention operation of the main paper body \cref{eq:attention}
\begin{equation}
\begin{split}
    \text{Att}(\feat, \adj) = & \text{softmax}\left(c_Q(\feat, \adj)c_K(\feat, \adj)^\top/\sqrt{F_o}\right)c_V(\feat, \adj) \\
    = & \text{softmax}\left(\mathbf{Q}\mathbf{K}^\top/\sqrt{F_o}\right)\mathbf{V},\\
\end{split}
\end{equation}
where the query, key and value matrices are defined as $\mathbf{Q}=c_Q(\feat, \adj)$, $\mathbf{K}=c_K(\feat, \adj)$, and $\mathbf{V}=c_V(\feat, \adj)$ respectively.   We want to prove:
\begin{equation}
\text{Att}(\perm \feat, \perm \adj \perm^\top) 
=
\perm \text{Att}(\feat, \adj)
\end{equation}
We already proved with \cref{eq:pe_conv:proof} that the convolution $c()$ is equivariant as defined in \cref{eq:pe_conv:property} and thus we can employ \cref{eq:pe_conv:property}. From the left
\begin{equation}
\begin{split}
\text{Att}&(\perm \feat, \perm \adj \perm^\top) \\
= & \text{softmax}\left(\frac{c_Q(\perm \feat, \perm \adj \perm^\top)c_K(\perm \feat, \perm \adj \perm^\top)^\top}{\sqrt{F_o}}\right)c_V(\perm \feat, \perm \adj \perm^\top) \\
= & \text{softmax}\left(\frac{\perm  \mathbf{Q} \left(\perm  \mathbf{K}\right)^\top }{\sqrt{F_o}}\right)\perm  \mathbf{V} \\
= & \text{softmax}\left(\frac{\perm  \mathbf{Q} \mathbf{K}^\top\perm^\top}{\sqrt{F_o}}\right)\perm  \mathbf{V} \\
 = & \perm\text{softmax}\left(\frac{\mathbf{Q} \mathbf{K}^\top }{\sqrt{F_o}}\right)\underbrace{\perm^\top\perm}_{\identity}  \mathbf{V} \\
  = & \perm\left(\text{softmax}\left(\frac{\mathbf{Q} \mathbf{K}^\top }{\sqrt{F_o}}\right)\mathbf{V}\right) \\
  = & \perm \text{Att}(\feat, \adj)
\end{split}
\end{equation}
where we considered that the softmax operation is equivariant to permutation, because its denominator (the sum of exponentials) is a commutative operation that remains constant regardless of the order of the elements. 

\paragraph{End-to-End Equivariance.} Since our network consists of these two equivariant operations and residual layers, as can be seen in \cref{fig:app:arch_layers} and is discussed in \cref{app:arch}, our architecture is end-to-end permutation equivariant. 

\subsection{Equivariance in a Generative Diffusion Model}
\label{app:pe:diffusion}
Let's discuss the implications of a generative formulation such as denoising diffusion models \cite{ho2020denoising} in the context of permutation equivariance. In a diffusion model, a  sample is generated from random noise $\varepsilon \sim \mathcal{N}(\mathbf{0}, \mathbf{I})$ sampled from a univariate Gaussian distribution. Since the architecture of our denoiser is end-to-end permutation equivariant with respect to the permutation matrix $\perm$, it naturally holds:  
\begin{equation}
\label{eq:pe_fixed_noise}
\netfunct(\perm \pastmotion{}{}, \perm \adj \perm^\top, \perm \varepsilon) = \perm \netfunct(\pastmotion{}{}, \adj, \varepsilon).
\end{equation}
We refer to this equation \cref{eq:pe_fixed_noise} as equivariance under "fixed noise", and prove this numerically as \textit{Ours on} $\perm + \perm\varepsilon$ in \cref{tab:ablations}. In this case, equivariance holds strictly, as fixing the noise results in a deterministic mapping. Instead, when the noise is not fixed, but randomly sampled as in every generative model, equivariance is said to hold on a distributional level and not on a sample level~\cite{lin2025equivariant, thiede2020general, wad2022equivariance}. We elaborate on this concept starting from a straighforward example. We sample two different noise variables $\varepsilon_1,\varepsilon_2  \sim \mathcal{N}(\mathbf{0}, \mathbf{I})$ and obtain two different samples, according to the definition of a generative model:  
\begin{equation}
\netfunct( \pastmotion{}{}, \adj ,\varepsilon_1) \neq \netfunct(\pastmotion{}{}, \adj, \varepsilon_2).
\end{equation}
Indeed, this core aspect is what allows us to generate novel latents from different noise samples. Of course, this holds also in the case of permutation
\begin{equation}
\netfunct(\perm \pastmotion{}{}, \perm \adj \perm^\top, \varepsilon_1) \neq \perm \netfunct(\pastmotion{}{}, \adj, \varepsilon_2).
\end{equation}
However, when considering not only a single sample   but the whole test set, the output distributions of $\netfunct(\perm \pastmotion{}{}, \perm \adj \perm^\top)$ and  $\netfunct(\pastmotion{}{}, \adj)$ have identical shape, up to the transformation $\perm$ that reorders the joint dimensions
\begin{equation}
p_{\netfunct(\perm \pastmotion{}{}, \perm \adj \perm^\top)} \simeq p_{\perm  \netfunct(\pastmotion{}{}, \adj)}.
\end{equation}
We refer to this property as \textit{distributional} equivariance or equivariance \textit{under stochastic sampling}~\cite{lin2025equivariant, thiede2020general, wad2022equivariance}   
\begin{equation}
\netfunct(\perm \pastmotion{}{}, \perm \adj \perm^\top) \overset{d}{=} \perm  \netfunct(\pastmotion{}{}, \adj).
\end{equation}
and prove it numerically as \textit{Ours on $\perm$} in \cref{tab:ablations}. 
\newline
Another interesting aspect here is the following.
Since a univariate Gaussian distribution is invariant to permutation, applying the permutation $\perm$ to a fresh random sample delivers another sample from the same distribution $\perm \varepsilon \sim \mathcal{N}(\mathbf{0}, \mathbf{I})$. 
In our setting, this means that computing the evaluation metrics for $\netfunct(\perm \pastmotion{}{}, \perm \adj \perm^\top)$ is coincident to computing the metrics of $\netfunct(\pastmotion{}{}, \adj)$ for a different initialization seed.  

\subsection{Evidence that Previous SHMP Approaches are not Equivariant}
\label{app:pe:shmp_works}
\label{app:exp:pe}

Previous SHMP approaches are not equivariant because they either learn the aggregation matrix for a specific kinematics, or because they treat the joints as features overfitting to the training kinematics. We provide experimental quantitative results in \cref{tab:perm_eq} and mathematical proof following the paradigm of \cref{app:sec:observation_proof}.

\paragraph{\textbf{Empirical Evidence via Quantitative Experiments (\cref{tab:perm_eq})}}.
In this section, we highlight the results in \cref{tab:perm_eq}, showing that current state-of-the-art HMP models are by far not permutation equivariant (PEQ). We consider models trained on a single Kinematics and its corresponding dataset, AMASS\cite{mahmood2019amass}, the largest and most diverse dataset in SHMP. When testing on the AMASS test split, which has the same kinematics as at training time, we apply a random permutation of the body joints in the input sequences.  We see that previous SHMP approaches fail dramatically, since the metrics strongly differ between conventional evaluation (as reported in \cref{tab:amass_intra}) and the evaluation under permutation (\textit{On $\perm$}). This behavior is to be expected, as previous approaches were never trained for permutations and for each kinematics the input joints are expected in a predefined order, matching the one seen at training time. Modifying any of these methods to an equivariant architecture is not straightforward (see \cref{app:sec:observation_proof}). Instead, our method has the only equivariant architecture and thus it is equivariant by design. The metrics for \methodname~ exhibit very low variation: as expected, our results are close but not identical under random permutation at inference time. The reason is that any generative model is permutation equivariant on a distribution level, as discussed in \cref{app:pe:diffusion}. The two evaluations can thus be interpreted as evaluations under different initial random seeds.

\paragraph{\textbf{Mathematical Evidence and Discussion}}.
In the following, we provide mathematical proofs of why current methods are not equivariant.  

\begin{table}[t]\footnotesize	
\vspace{0.3cm}
\centering
\setlength{\tabcolsep}{2.6pt}
\caption{Evaluation with random permutation at inference time. For each method, we report 1) conventional evaluation metrics on AMASS dataset\cite{mahmood2019amass} as in \cref{tab:amass_intra}, 2) Evaluation on the same data but with body joints  randomly permuted (\textit{On $\perm$}). Previous HMP approaches fail dramatically, as our method has the only permutation equivariant architecture. No model has seen permutations during training, and all models are trained on the same kinematics and dataset, AMASS. The most constant values under permutation are highlighted in \textbf{bold}, second-best are {underlined}.
}
\begin{tabular}{l H rrr rr rr}
\toprule 
& \multirow{2}{*}{} & \multicolumn{3}{c}{Precision $\downarrow$}  & \multicolumn{1}{c}{Div $\uparrow$} &  \multicolumn{1}{c}{Real $\downarrow$} & \multicolumn{2}{c}{B Real $\downarrow$} \\
 \cmidrule(lr){3-5}  \cmidrule(lr){6-6} \cmidrule(lr){7-7}\cmidrule(lr){8-9} 
Method &  & ADE  & FDE    & MAE  & APD & CMD  &  str  & jit   \\ %
\midrule
 \multirow{1}{*}{DLow~\cite{yuan2020dlow}} & w/o& 0.590 & 0.612 & 8.510 & {13.170}  & {15.185} & 8.41 & 0.40 \\
  \qquad on $\perm$ & w & 2.389 & 2.491 & 65.179 & 13.137 & 44.113 & 1.87 & 0.03 \\
 \cmidrule(lr){2-9}
 \multirow{1}{*}{DivSamp~\cite{dang2022diverse}} & w/o & 0.564 & 0.647 & 8.027 & \textbf{24.724}  & 50.239 & 11.17 & 0.82 \\ %
  \qquad on $\perm$ & w & 4.167 & 1.896 & 53.504 & {37.297} & 10171.783 & 2.73 & 0.96 \\
  \cmidrule(lr){2-9}
 \multirow{1}{*}{BeLFusion~\cite{barquero2023belfusion}}& w/o& {0.513} & {0.560} & 7.125 & 9.376 & 16.995 & 7.19 & 0.34  \\
 \qquad on $\perm$ & w & 1.530 & 1.719 & 54.826 & 11.902 & 40.985 & 1.85 & 0.04 \\
\cmidrule(lr){2-9}
\multirow{1}{*}{CoMusion~\cite{suncomusion}} & w/o
& {0.494} & {0.547} & 6.715 & {10.848}  & {9.636} & {4.04} & {0.25} \\
 \qquad on $\perm$ & w & 2.282 & 2.456 & 64.120 & {15.229} & 23.136 & 1.88 & 0.04 \\
\cmidrule(lr){2-9}
\multirow{1}{*}{SkelDiff~\cite{curreli2025nonisotropic}} & w/o& {0.480} & {0.545} & {6.124} & 9.456 & {11.417} &   {3.15} &   {0.20} \\ %

 \qquad on $\perm$ & w & 2.041 & 2.440 & 64.339 & 3.371 & 29.083 & 153.52 & 2.57 \\
\midrule
\midrule
 \multirow{1}{*}{\methodname~(A)}&w/o & \textbf{0.496 }& \textbf{0.560} & \textbf{6.214} & \textbf{8.241} & \textbf{13.097} & \textbf{0.00} & \textbf{0.00} \\
 \qquad on $\perm$ &w & \textbf{0.497} & \textbf{0.562} & \textbf{6.221} & \textbf{8.242} & \textbf{13.089} & \textbf{0.00} & \textbf{0.00} \\

\bottomrule
\end{tabular}

\label{tab:perm_eq}
\end{table}

\begin{enumerate}
\item \textbf{MLP or Linear Layer.} Since the weights $\Weights$ are learned for specific joint indices, swapping rows in the input does not swap the corresponding rows in the output.
\begin{equation}
o_{MLP}(P \pastmotion{}{}^{'}) = \Weights (\perm \pastmotion{}{}^{'}) \neq \perm (\Weights \pastmotion{}{}^{'}) \quad \text{because} \quad \Weights \perm \neq \perm \Weights \text{ (in general)}.
\end{equation}
\item \textbf{Graph Convolutions with Learned Aggregation Matrix.} Even with a spatial aggregation matrix $\mathbf{G}$, the left-multiplication by learned weights $\Weights$ ties features to specific indices.
\begin{equation}
    o_{GC_{learned}}(\perm \pastmotion{}{}^{'}) = \Weights (\perm \pastmotion{}{}^{'}) \mathbf{G} + \mathbf{\tilde{F}} \neq \perm (\Weights \pastmotion{}{}^{'} \mathbf{G} + \mathbf{\tilde{F}}).
\end{equation}

\item \textbf{Recurrent Units (GRU, LSTM)}. Because the hidden state $\mathbf{H}_{t-1}$ and the input transformation $\Weights$ are fixed to a specific joint ordering, permuting the input vector $\pastmotion{}{t}^{'}$ breaks the alignment with the learned parameters.
\begin{equation}
    o_{RNN}(\perm \pastmotion{}{t}^{'}) = \Weights (\perm \pastmotion{}{t}^{'}) + \mathbf{G}_{RNN}\mathbf{H}_{t-1} \neq \perm (\Weights \pastmotion{}{t}^{'} + \mathbf{G}_{RNN}\mathbf{H}_{t-1}).
\end{equation}

\item \textbf{Transformers with Time as Token Dimension and Joints as Features}. Since the projections to Query, Key, and Value spaces are linear layers acting on the joint dimension (multiplication from the left), the attention map becomes corrupted under permutation.
\begin{equation}
    o_{Tranf}(\perm \pastmotion{}{}^{'}) \propto \text{attn}(\Weights_Q \perm \pastmotion{}{}^{'}, \Weights_K \perm \pastmotion{}{}^{'}) (\Weights_V \perm \pastmotion{}{}^{'}) \neq \perm o_{Tranf}(\pastmotion{}{}^{'}).
\end{equation}
Additionally, current transformer-based approaches employ positional encoding, which is by definition not equivariant as it is designed to contain order information. When employing attention, the token dimension is typically associated with time~\cite{suncomusion, chen2023humanmac}, and just alternately in subcomponents ~\cite{curreli2025nonisotropic, suncomusion} or at intermediate stages \cite{wei2023human, dang2022diverse, mao2021generating} with joints.
\item \textbf{Typed-Graph Convolutions}. Because each joint index $j$ is mapped to a specific weight $\mathbf{G}_j$ based on its type $\tau(j)$, permuting the joints $\perm \pastmotion{}{}$ moves a joint of one type (e.g., "hip") into a slot evaluated by a weight for another type (e.g., "shoulder").
\begin{equation}
    o_{GC_{typed}}(\perm \pastmotion{}{}^{'}) = \Weights \begin{bmatrix} (\perm \pastmotion{}{}^{'})_{0} \mathbf{G}_0 \\ (\perm \pastmotion{}{}^{'})_{1} \mathbf{G}_1 \\ \vdots \end{bmatrix} \neq \perm o_{GC_{typed}}(\pastmotion{}{}^{'}).
\end{equation}
\end{enumerate}
These architectures cannot be straightforwardly rendered PEQ by removing or adapting components, and ultimately rely on a fixed joint number and ordering. 

\section{More Details on \methodname}
\label{app:method_details}
\subsection{Training Losses}
We follow the training paradigm of latent diffusion models \cite{rombach2022highresolution}: we first train an autoencoder to learn a temporally-compressed latent space and then a denoiser to denoise true latent variables in that latent space.
\paragraph{Autoencoder: Reconstruction Loss.}
Similar to other approaches \cite{barquero2023belfusion, curreli2025nonisotropic}, the autoencoder learns a latent space by reconstructing complete motion sequences from their latent representations.
Given a motion $\motion \in \mathbb{R}^{T \times N \times 3}$, the encoder compresses it into a latent vector $\latentvar{} \in \mathbb{R}^{ \numnodes \times L}$, and the decoder reconstructs the motion $\tilde{\motion}$ from this latent code.
The model is optimized following the reconstruction loss
\begin{equation}
  \mathcal{L}_{\mathrm{rec}} (\motion, \tilde{\motion}):= \| \motion -  \tilde{\motion}\|_1 .
  \label{eq:rec_loss}
\end{equation}

\paragraph{Denoiser: Diffusion Loss.} In the  diffusion training process $q(\latentvar{t} \mid \latentvar{t-1})$, a clean latent sample $\latentvar{0}$ is progressively perturbed over timesteps $t=1,\dots,T$ by adding noise of magnitude proportional to $t$ according to a cosine noise scheduler, yielding a Gaussian distribution at $t=T$. In the generative  denoiser learns an approximation of the reverse process $p_\theta(\latentvar{t-1} \mid \latentvar{t})$, which iteratively denoises $\latentvar{T}$ back towards the data distribution. For each diffusion timestep $t$, the denoiser regresses directly the denoised latent $\latentvar{\boldsymbol{\theta}}$ ~\cite{ramesh2022hierarchical, barquero2023belfusion, suncomusion, curreli2025nonisotropic}, rather than the added noise~\cite{ho2020denoising, rombach2022highresolution}.
To avoid penalizing samples that are different from the ground truth yet realistic, we relax the diffusion objective \cite{gupta2018social} by sampling $k=50$ times and backpropagating the loss to the closest sample \cite{curreli2025nonisotropic, barquero2023belfusion}:
\begin{equation}
\label{eq:diff_loss}
    \mathcal{L}_{\mathrm{diff}}(\latentvar{\boldsymbol{\theta}}^k, \latentvar{})= \mathbb{E}_{\futurevar, \pastvar, t}  \arg \min_k \left(\alphaT{\bar}{t} \|\latentvar{\boldsymbol{\theta}}^k  - \latentvar{}\| \right).
\end{equation}
As this relaxation achieves higher diversity in the predictions but extends training time, we thus present some of our ablations without relaxation $k=1$ \cite{curreli2025nonisotropic, barquero2023belfusion}. 

\subsection{Inference on Partial Skeletons}
Our method has never seen partial skeletons with missing joints during training. 
While partiality is highly relevant for real-world applications, data collection is not straightforward, and there are no specific SHMP datasets to date. Partial motions are thus usually investigated by masking limbs of motions parametrized with existing full-body kinematics. We follow this procedure and mask the input (both motion and adjacency) by randomly picking limb IDs. 

\methodname~ has never been trained for the generation of missing limbs either, but we observe this capability as a side effect. We generate missing limbs with a simple heuristic as a proof of concept. For an input observation $\pastmotion{}{}$ with missing limbs, we obtain its latent embedding as $\latentvar{past}$ with the corresponding partial adjacency matrix $\adj$. In the latent embedding, joints and limbs that were not present in the input are also not present. Since we know the adjacency matrix of the full-body skeleton, we can employ it in the diffusion process and in the decoder. During diffusion, we can effectively generate the missing parts by sampling noise also for the missing limbs.  But to do this, since the diffusion process uses the observation latent as conditioning, we need to find an initialization value for the missing limbs in the latent representation of the past motion.  
 To increase realism in the generated body part, we initialize the conditioning past latent through the values of the opposite limbs. In the qualitatives, we see that this does not result in symmetric motions for the generated output. We believe more ad-hoc or sophisticated inference heuristics, or rather training strategies, can be applied to target this issue, and leave it as an object of future work.
\begin{figure*}[t]
  \centering
  \includegraphics[trim=1.cm 0cm 1.cm 0cm, clip,width=01.\textwidth]{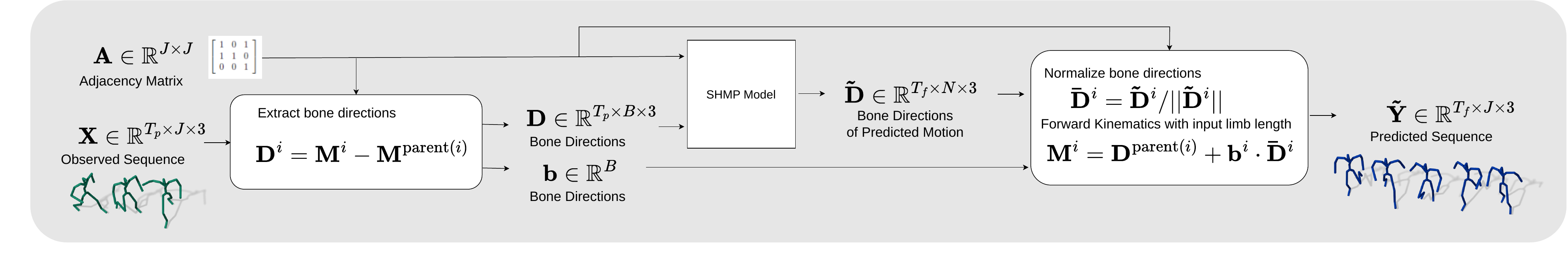}
  \caption{\textbf{Pipeline overview of  parametrizing motions as bone directions}. We also depict the masking procedure of \methodname. The parametrization can be applied to any SHMP model without additional modification. 3D keypoints are computed from bone directions via inverse kinematics.}
  \label{fig:app:parameterization}
\end{figure*}

\subsection{Parametrizing Motion as Bone Directions}
\label{app:parametrization}
Here, we provide more details about our motion parametrization as bone directions. An overview is provided in \cref{fig:app:parameterization}, displaying how this parameterization can be applied to any HMP model, not only \methodname.
We feed the  limb direction vectors to the encoder, allowing the network to access both orientation and implicit bone length information, since the directions are not normalized.
For decoding, the model predicts unnormalized limb directions, which are then normalized and rescaled by the bone lengths of the input skeleton.
This effectively corresponds to predicting pure limb directions while enforcing constant bone lengths, thereby removing length jitter trivially and ensuring geometric consistency in the reconstructed motions. During our preliminary studies we discovered that both feeding and computing the loss on the unnormalized bone directions instead of the normalized version improves training stability and delivers better performance.
This representation remains numerically stable during training \cite{bie2022hit} and does not suffer from singularities ~\cite{ salzmann2022motron, geist2024learning}.

\subsection{Architecture Details}
\label{app:arch}
\begin{figure*}[t]
  \centering
  \includegraphics[trim=0cm 0cm 0cm 0cm, clip,width=\textwidth]{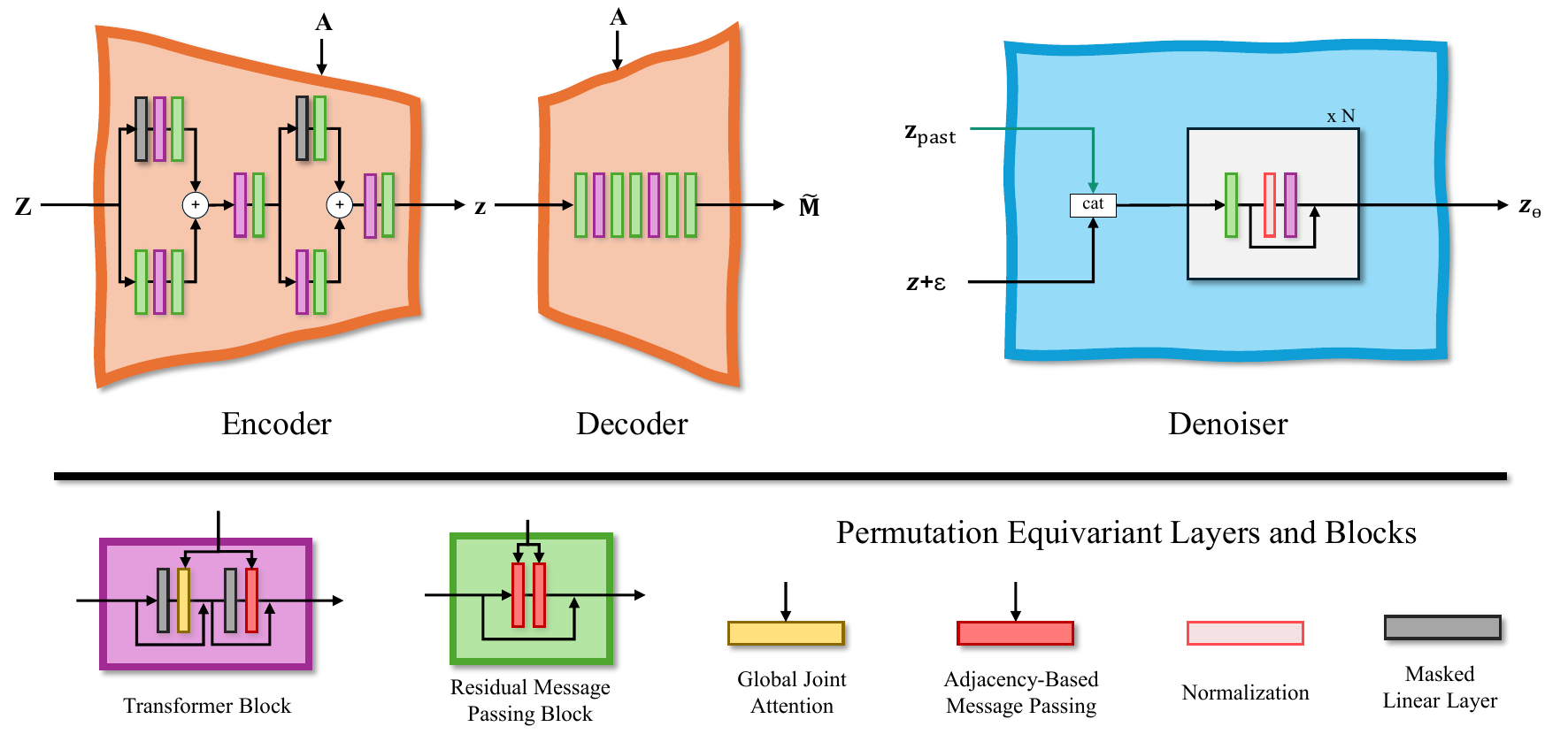}
  \caption{\textbf{Overview of the architecture layers of our model.} We provide a more detailed view on the permutation equivariant architecture of our model, including layers and connections described in \cref{app:arch}. Our code will be publicly available.}
  \label{fig:app:arch_layers}
\end{figure*}

\paragraph{Overview}
In this section, we provide a detailed description of the input-output flow of our model and the architecture of the autoencoder and denoiser. A visualization is given in \cref{fig:app:arch_layers}.
We remark here that we will make our code public.

\subsubsection{Single Networks}

\paragraph{Encoder}
The challenge of a fully PEQ architecture for HMP consists in extracting meaningful temporal and spatial information without breaking the PEQ property.
One could extract such information in parallel for time and joint dimensions (inspired by Google Inception architecture), which requires a higher amount of resources and slower forward passes. In preliminary experiments, we observed that this approach yielded no benefits.
This provided inspiration for parallel branches, which resulted in slightly improved reconstruction and faster convergence during autoencoder training.
While the first parallel branch halves the initial feature dimension of $3T$, the residual message-passing block before the second parallel branch reduces the feature dimension to $L$.
To allow for the processing of sequences of arbitrary length, we employ zero-padding along the time dimension of the encoder's input. This allows us to take the training stage II and at inference, use the past motion as input, following previous works \cite{curreli2025nonisotropic} and save resources for training a network solely for the past \cite{barquero2023belfusion}. The autoencoder is never trained with the gradient from the past observation.

\paragraph{Decoder}
For the decoder, we stack two transformer blocks, followed by two residual message passing blocks each. An initial residual message passing block is used to increase the feature dimension from $L$ to $3T$.

\paragraph{Denoiser}
In the denoiser, we concatenate the conditioning latent vector of the past motion $\latentvar{past}$ with the current latent vector $\latentvar{}$ before feeding it into the blocks. Each block consists of a message passing block and a transformer block with a skip connection, with Root Mean Square Layer normalization as commonly paired with transformers. %

\subsubsection{Layers}

\paragraph{Adjacency-Based Message Passing}
This layer performs the graph convolution operation described in \cref{eq:graph_conv}.

\paragraph{Transformer Block}
For the transformer Block, we use the attention mechanism described in 
\cref{eq:attention}
in the main paper, where the query, key, and value weights are extracted from the input features via graph convolutions. 
We add a residual connection around both blocks and nonlinearities. 

\paragraph{Residual Message Passing Block}
We group pairs of two of the adjacency-based message passing layers with a hyperbolic tangent  nonlinearity in between and a skip connection around both layers, following the successful fashion of  conventional residual blocks.

\paragraph{Masked Linear Layer}
To ensure that our architecture remains PEQ despite masking, we employ linear layers only on the feature dimension, never on the joint dimension, ensuring that masked joints always have zeroed features.

\section{Implementation Details}
\label{app:implementation_details}
We employ the same training hyperparameters for any dataset or dataset combination. 
The autoencoder is trained for 300 epochs, while the denoiser for 375 epochs with a learning rate of 0.005 and  $T=10$ diffusion steps, following a cosine noise scheduler \cite{nichol2021improved}. At inference we draw from a DDPM sampler \cite{ho2020denoising}. Both networks are trained with Adam on PyTorch. Our model is \textbf{always} trained for 7.9M parameters, at least twice as compact as the smallest diffusion competitor and even more compact than VAE baselines. Numbers reported with inference time in \cref{tab:computational_efficiency}. We train the autoencoder on an RTX5000 and the diffusion model on an NVIDIA A40. The longest training does not take more than 5 days, shorter than the closest competitor SkelDiff, which also needs to be trained anew for every new dataset. Following previous works \cite{curreli2025nonisotropic, salzmann2022motron}, we chose a latent dimension of $L=96$, achieving a 4x compression of the input space.
When training with kinematics that have a different number of joints (i.e. AMASS and Nymeria), we zero-pad the inputs to the highest joint number, which has no implication for our equations as the weight matrices are independent on the number of joints. 
As the HMP task is defined for a past of 0.5 seconds and a future of 2 seconds, it results in a different number of input and output frames depending on the FPS (Hz) of each dataset. We train our model on the maximum FPS (60 Hz, as in AMASS and Nymeria), and deal with lower FPS by frame interpolation (see \cref{app:exp_setting:fps}). When training the diffusion model, to avoid spurious correlations between the noise and the adjacency matrix (of which the model sees only one (AMASS) or two instances (AMASS + Nymeria), we include a random permutation for the joints of 0.5.

For our experiments without the bone direction parametrization, we train our model with 3D keypoints as input, employing the same rescaling approach of SkelDiff~\cite{curreli2025nonisotropic}.

\section{Metrics Definition}
\label{app:exp_setting:metrics}
\paragraph{Overview} We report the precision metrics of the Average Distance Error (ADE), the Final Distance Error (FDE), and the Mean Angle Error (MAE),in degrees, together with their multimodal equivalents (MMADE, MMFDE). For diversity, we report the Average Pairwise Distance (APD) between predictions. The Average Pairwise Distance Error (APDE), relates the APD with the multimodal GT.  Realism metrics are comprised of the Cumulative Motion Distribution (CMD), which penalizes deviations from the expected average displacement of the dataset and body realism in the form of limb stretching (str) and jittering (jit). Their respective mean and Root Mean Squared Error (RMSE) are reported in percentage. 
In the main paper body, for space reasons in most tables  we report a subset of all metrics, selecting the most relevant ones. The others behave analogously, as can be seen in the corresponding extended version of each table in the \app.

On AMASS, FID is conventionally not computed. The FID computation requires features from a classifier, and labels necessary to train this model for AMASS do not exist. On H36M, we do not compute FID for the ZeroVelocity baseline as it does not output a distribution.
For the detailed metrics equations, we refer to the appendix of Curreli \etal\cite{curreli2025nonisotropic} and the main body of Barquero \etal \cite{barquero2023belfusion}. 
For computing APDE with the different retargeting procedures, we do not perform retargeting on the reference values but keep them in the original space.

\subsection{Unified Metrics: Comparing among Datasets}
We realize that conventional SHMP metrics are unsuitable to compare model performance across datasets. 
Let's take as an example the Average Distance Error (ADE). For a set of k predictions $\tilde{\futurevar{}} \in \mathbb{R}^{\futurewindow \times \numjoints \times 3}$, the ADE is defined as 

 \begin{equation}
         \mathrm{ADE}(\tilde{\futurevar{}}, \futurevar{}) = \min_{k} \frac{1}{\futurewindow}\sum_{t=0}^{\futurewindow} \sqrt{\sum_{j=1}^{\numjoints*3} ( ^k\tilde{\futurevar{}_t}^j  - \futurevar{}_t^j)^2}.
 \end{equation}
 We note here that the joint dimension is included in the Euclidean Distance computation, thus not averaging over the number of joints. Summing instead of averaging over the number of joints does not allow to compare ADE scores on topologies or datasets that exhibit a different number of joints. 
 
 We simply employ a ADE version \emph{unified} among datasets, by averaging over the joint dimension: 
  \begin{equation}
         u\mathrm{ADE}(\tilde{\futurevar{}}, \futurevar{}) = \min_{k} \frac{1}{\futurewindow}\frac{1}{\numjoints}\sum_{t=0}^{\futurewindow} \sum_{j=1}^{\numjoints} \sqrt{\sum_{d=1}^{3} ( ^k\tilde{\futurevar{}_t}^{j,d}  - \futurevar{}_t^{j,d})^2}.
 \end{equation}
Analogously, we define \textit{u}FDE, uMMADE, and uMMFDE measured in decimeters; and \textit{u}APD
measured in meters.

\section{Experimental Settings}
\label{app:exp_settings}
Following the HMP task definition of previous works, in our experiments, we consider a past of 0.5 seconds and a future of 2 seconds. While the AMASS topology is designed for 22 joints, H36M has 17, and Nymeria 23 (including the hip joint).

\subsection{Baselines}
\label{app:exp_settings:baselines}
\paragraph{Details.} 
Since we are the first to adapt Nymeria for the task of HMP, we also need to train previous methods on this dataset.
When the code of the latest baseline was available, we trained them on Nymeria employing their configurations for AMASS. We chose this setting because the two datasets are comparable in size, and the number of keypoints differs only by one (while H36M is much smaller and has fewer joints).
For HumanMAC, the checkpoint on AMASS or its configurations is not available. Thus we increase the number of layers and the training time compared to the H36M configuration. To obtain the \textit{u}ADE, \textit{u}FDE, and \textit{u}APD metrics (\cref{fig:plots_params}) when the AMASS checkpoint is not available, we compute them through interpolation from their ADE, FDE, APD. 

\paragraph{Adapting a Baseline to Multidataset Training.}
In \cref{tab:ablations} we presented version of SkelDiff~\cite{curreli2025nonisotropic} adapted to multidataset training, showing that data alone does not lead to generalization in a zero-shot kinematics setting. To adapt the baseline to multidataset we first removed all components that were dependent on specific joint \textit{types}: 1) the weights of the typed-graph convolutions, which used to be learned  independently per joint type (\eg shoulders, legs, hands, etc.), were substituted by a single weight matrix learned for all joint types simultaneously, 2) the anisotropy was removed from the diffusion training, as it relied on fixed joint positions given by the adjacent matrix. Then we trained the model with joint permutation as data augmentation. This lead us to a baseline that employs similarly to us attention and graph convolutions, but in the graph convolution as described in \eqref{eq:graph_conv} learns the aggregation matrix $W$ from data instead of using the adjacency matrix. To support kinematics chains of different size, we set the size of this matrix equal to the largest cardinality of the kinematics in our training set. We note that this approach by definition cannot generalize at inference to kinematics chains of size larger than the ones seen at training. Beyond this specific scenario, the resulting baseline can potentially fulfill Lemma \ref{lem:joint_independenec} in dependence of the training data.

\paragraph{Retargeting.}
In the main paper, we discuss the retargeting from AMASS to H36M. AMASS has 22 joints, while H36M has 17, so joints that do not have any correspondence in H36M are simply discarded.  
Converting the output back from H36M to AMASS to evaluate in the input topology would imply creating limbs and joints that are not present in H36M. To this purpose, we completed the retargeting algorithm of Holden \etal \cite{holden2016deep} employed by HumanMAC\cite{chen2023humanmac} with suitable positioning of the collarbones on the shoulder line and feet perpendicular to the leg bones, making use of the foot lengths in the observation.

\subsection{Adapting Nymeria to HMP}
\label{app:exp_setting:nymeria}
The Nymeria dataset is a very large collection of motion data originated from egocentric motion in-the-wild through body tracking in connection with Project Aria. It is recorded for 23 joints, including the root hip joint. It contains 300 hours on a total of 1200 sequences with 264 participants, recorded in 50 indoor and outdoor locations. We find the quality of the recorded motion to be suboptimal for motion prediction, as it often occurs in real-life situations: a large percentage of recordings display moments of stillness. To derive a dynamic level more similar to AMASS, we prune the data, removing sequence parts that have a mean joint velocity between consecutive frames inferior to 6.913 cm, for a total of 51M removed frames. We thus retain 1094 of the original sequences and split 44 of them into suparts. This leaves us with a total of 10M frames, a size comparable to AMASS. The overall data remains relatively static, as evident from the evaluation scores of the ZeroVelocity baseline on AMASS and Nymeria: the ADE on Nymeria is significantly lower, despite having one additional joint.

\subsection{Inference on FPS Different From Training Time}
\label{app:exp_setting:fps}
As the HMP task is defined for a past of 0.5 seconds and a future of 2 seconds, it results in a different number of input and output frames depending on the FPS (Hz) with which each dataset was recorded. We train our model on the maximum FPS (60 Hz, as in AMASS and Nymeria), and deal with lower FPS by frame interpolation. For example, for cross-retargeting on H36M, collected at 50 FPS, we upsample the  25 input frames to 30 frames by duplicating some selected ones. The 120 output frames are donwsampled to 100 frames analogously, to match the FPS of the GT. So, when evaluating at a different FPS than the training data, we evaluate at the frequency of the evaluation dataset (the original FPS of the GT). We follow the same procedure to allow baselines to perform cross-topology (through retargeting) at different FPS than the one seen at train time (\cref{tab:crossdataset_trainAMASSevalH36M_old_metrics}, \cref{tab:crossdataset_trainAMASSevalH36M}, \cref{tab:crossdataset_trainH36MevalAMASS}, \cref{tab:crossdataset_trainH36MevalAMASS_old_metrics}).
To show that such procedure does not give us any advantage in comparison with other state-of-the-art baselines, we conduct two experiments. 
\begin{table}[th!] \footnotesize	
\centering
    \caption{\textbf{Quantitative results on AMASS for methods fed an input downsampled to 30 FPS and upsampled to 60 FPS again.} The dataset is recorded at 60 FPS, and baselines trained at the same resolution. Results and rankings are consistent with \cref{tab:amass_intra}, showing that methods are robust to such small temporal distortion in the input data. 
    }
\setlength{\tabcolsep}{2.3pt}
\begin{tabular}{l ccccc cccc}
    \toprule
          & \multicolumn{3}{c}{Precision $\downarrow$}  & \multicolumn{1}{c}{Div $\uparrow$} & \multicolumn{1}{c}{Real $\downarrow$} & \multicolumn{4}{c}{{Body Real $\downarrow$}}\\
  \cmidrule(lr){2-4}   \cmidrule(lr){5-5} \cmidrule(lr){6-6} \cmidrule(lr){7-10}
 \multirow{2}{*}{Method}  & &  & & & & \multicolumn{2}{c}{mean $\downarrow$} & \multicolumn{2}{c}{RMSE $\downarrow$} \\  
 & ADE  & FDE  & MAE   & APD  & CMD  &  str  & jit & str  & jit\\
    \midrule
DLow & 0.589 & 0.615 & 0.148 &  \underline{13.166} & \underline{15.849} & 8.40 & 0.39 & 11.05 & 0.55 \\
DivSamp & 0.564 & 0.649 & 0.140 &  \bfseries{24.719} & 50.230 & 11.18 & 0.82 & 16.73 & 1.07 \\
BeLFusion & \underline{0.507} & \underline{0.570} & \underline{0.124} &  7.461* & 19.627 & \underline{7.21} & \underline{0.22} & \underline{8.74} & \underline{0.29} \\
SkelDiff & \bfseries{0.480} & \bfseries{0.548} & \bfseries{0.107} &  9.453 & \bfseries{11.418} & \bfseries{3.15} & \bfseries{0.20} & \bfseries{4.44} & \bfseries{0.26} \\
    \bottomrule
\end{tabular}

\label{app:tab:amass_downsample30fps_interpolated}
\end{table}

\begin{table}[ht!] \footnotesize	
\centering
    \caption{\textbf{Quantitative results on AMASS for methods evaluated at 30FPS instead of 60 FPS.} Dataset is recorded at 60 FPS, and baselines trained at the same resolution. The GT and the method's output are downsampled to 30 FPS. Results are consistent with \cref{tab:amass_intra}, up to range changes introduced by averaging on a smaller number of frames (APD).}
\setlength{\tabcolsep}{2.3pt}
\begin{tabular}{l ccccc cccc}
    \toprule
          & \multicolumn{3}{c}{Precision $\downarrow$}  & \multicolumn{1}{c}{Div $\uparrow$} & \multicolumn{1}{c}{Real $\downarrow$} & \multicolumn{4}{c}{{Body Real $\downarrow$}}\\
  \cmidrule(lr){2-4}   \cmidrule(lr){5-5} \cmidrule(lr){6-6} \cmidrule(lr){7-10}
 \multirow{2}{*}{Method}  & &  & & & & \multicolumn{2}{c}{mean $\downarrow$} & \multicolumn{2}{c}{RMSE $\downarrow$} \\  
 & ADE  & FDE  & MAE   & APD  & CMD  &  str  & jit & str  & jit\\
    \midrule
    DLow & 0.586 & 0.611 & 0.148 &  \underline{9.289} & 15.849 & 8.37 & 0.78 & 11.03 & 1.10 \\
DivSamp & 0.561 & 0.645 & 0.139  & \bfseries{17.451} & 50.230 & 11.13 & 1.65 & 16.68 & 2.13 \\
BeLFusion & 0.505 & 0.565 & 0.124  & 5.261 & 19.627 & 7.18 & \underline{0.44} & 8.73 & \underline{0.59} \\
CoMusion & \underline{0.491} & \underline{0.546} & \underline{0.117}  & 7.644 & \bfseries{9.661} & \underline{4.03} & 0.65 & \underline{5.61} & 0.85 \\
SkelDiff & \bfseries{0.477} & \bfseries{0.544} & \bfseries{0.106} & 6.665 & \underline{11.418} & \bfseries{3.14} & \bfseries{0.39} & \bfseries{4.44} & \bfseries{0.51} \\
    \bottomrule
\end{tabular}

\label{app:tab:amass_downsample30fps}
\end{table}

\paragraph{Table \ref{app:tab:amass_downsample30fps_interpolated}}. First, we show that current methods are robust to FPS interpolation in \cref{app:tab:amass_downsample30fps_interpolated}. 
We downsample the input to 30FPS and upsample it again before feeding it to methods trained with 60 FPS on AMASS. By comparing the evaluation numbers with the standard evaluation scores of \cref{tab:amass_intra}, we see that methods are robust to such light temporal distortion in the input. The ranking remains unchanged and we observe just negligible variations in the score. These may be dependent on the fact that generative methods are sensitive to random generator states, which are initialized differently between different GPU architectures despite the same random seed. We conduct both experiments on a RTX6000.

\paragraph{Table \ref{app:tab:amass_downsample30fps}}. Second, we show that evaluating methods by changing the FPS of the GT does not change the ranking (\cref{app:tab:amass_downsample30fps}). We conduct this experiment on AMASS by downsampling both output and GT to 30 FPS.  While the ranking is unchanged, we see that the range of some metrics has changed (APD) compared to the reference table (\cref{tab:amass_intra}). The reason behind this range shift is that the metrics are computed by averaging over a smaller number of frames (60 instead of 120). Hence, to ensure a fair comparison with other methods and with other tables of conventional HMP evaluation, we decide to upsample the output to the original dataset FPS.

\section{Additional Experiments}
\label{app:experiments}

\begin{table}[ht!]\footnotesize	
\vspace{-0.3cm}
\centering
\caption{\small Model footprint for a single H36M inference (RTX 6000). Our model does not require multiple instances or more parameters to generalize to additional kinematics or datasets, while this does not hold for all other models (see \cref{fig:plots_params}).
}
\begin{tabular}{lrrr}
\toprule
& Memory$\downarrow$ & NumParams$\downarrow$ & Time$\downarrow$  \\
\midrule
 DLow~\cite{yuan2020dlow}  & 31 MB & 8.1 M & 111 ms   \\
 DivSamp~\cite{dang2022diverse}   & 88 MB & 23.1 M & 8 ms    \\
 BeLFusion~\cite{barquero2023belfusion}  & 53 MB & 17.8 M & 10\,341 ms    \\
 HumanMAC~\cite{chen2023humanmac}   & 114 MB & 28.7 M & 7\,438 ms  \\
 CoMusion~\cite{suncomusion}  & 87 MB & 19 M & 153 ms %
 \\
 SkelDiff~\cite{curreli2025nonisotropic}  & 106 MB & 26.5 M & 412 ms   \\
  \midrule
   \methodname  & 31MB & 7.9M* & 192 ms \\
\bottomrule
\end{tabular}

\label{tab:computational_efficiency}
\end{table}

\subsection{Computational Efficiency and Inference Time}
\label{app:exp:efficiency}
\paragraph{Footprint.} The state-of-the-art performance of our method is also accompanied by efficiency, both in terms of memory usage and computational complexity. In \cref{tab:computational_efficiency}, we compare our model's computational footprint to that of other competitors for a \underline{single} dataset, H36M. Our method is the smallest in terms of parameters, being from two to three times smaller compared to the closest competitor (SkelDiff). At the same time, it is twice as fast at inference. Methods with a similar time and memory footprint, such as DLow, are VAE-based and produce significantly worse results in all metrics (see, for example, \cref{tab:h36m_intra}). On a single dataset, training takes  around half of the time of the closest competitor, SkelDiff. Additionally, we do not need to retrain a new model for each dataset, unlike other approaches. This leads to the scalability advantages described in the next paragraph.

\paragraph{Scalabilty.} As proven mathematically in \cref{app:pe:proof_observation}, we do not scale with the number of kinematics or joints, while previous approaches do. We scale constantly, as shown in \cref{fig:plots_params}. Hence, a single model, trained once, in less time than others, is enough for all datasets. When considering one dataset, we are 75\% more compact than the most competitive baseline SkelDiff. When considering three datasets (AMASS, Nymeria, H36M), we are 90\% more compact than SkelDiff: we still require a single model, while others require three instances. Our efficiency becomes particularly advantageous for applications that require zero-shot kinematics reasoning. To train and natively process different skeleton kinematics, previous methods require retraining and storing a dedicated network for each. Hence, their memory usage grows linearly with the number of skeletal structures considered. Instead, our method naturally operates across kinematics and performs both training and inference with a single model. 

\subsection{Zero-Shot Kinematics on AMASS}
\label{app:exp:zero-shot-amass}
In the main paper, we reported the results for methods trained on AMASS and tested on H36M (\cref{tab:crossdataset_trainAMASSevalH36M}. For completeness, we also report the inverse, where models are trained on H36M and tested on AMASS. We highlight that such a case is particularly challenging for method generalization, as 1) H36M is a significantly smaller dataset, 2) the kinematics of H36M $\Kkin{H}$ has only 17 joints, while $\Kkin{A}$ of AMASS has 22 joints. In general, methods trained on H36M may lead to overfitting, also due to the low number of subjects. We report results in \cref{tab:crossdataset_trainH36MevalAMASS} with analogous results as when investigate the opposite direction (AMASS $\leftrightarrow$ H36M in (\cref{tab:crossdataset_trainAMASSevalH36M}). When trained only on H36M, \methodname~ performs in line with the state of the art. However, an advantage of our method is the possibility to experiment with other data priors without any modification. We observe that training on Nymeria yields improved results, suggesting that this dataset is a better fit to the target distribution. For other methods, this would require defining specific retargeting techniques for every pair of training and test distributions. The retargeting method used between AMASS and H36M \cite{holden2016deep} cannot be applied directly to Nymeria, since the spine and hips have distinct structures, and so we can't compare with other methods trained in same conditions.

\begin{table*}[!th]
\scriptsize 
\centering
\setlength{\tabcolsep}{0.75pt}
\caption{
\textbf{Evaluation of Zero-Shot Kinematics on  AMASS($\Kkin{A}$)~\cite{mahmood2019amass}}. Baselines are trained on H36M~\cite{Ionescu2014} (H, kinematics $\Kkin{H}$). This is a challenging retargeting case, the inverse direction of the case presented in the main body \cref{tab:crossdataset_trainAMASSevalH36M}: the inference kinematics $\Kkin{A}$ has more joints than the training one $\Kkin{H}$ (22 vs 17 joints).  
We are the only existing method supporting inference on novel kinematics out-of-the-box, while previous approaches require additional kinematics conversion \cite{chen2023humanmac, holden2016deep}.   We are the first method to support multiple kinematics natively and thus present a model trained additionally on Nymeria~\cite{ma2024nymeria} (N, $\Kkin{N}$).  The best results are highlighted in \textbf{bold}, second-best are \underline{underlined}. Conventional metrics rank identically, see \cref{tab:crossdataset_trainH36MevalAMASS_old_metrics}. 
}
\label{tab:crossdataset_trainH36MevalAMASS}
\begin{tabular}{L{76pt} rrr rrr r r  rrrr}
\toprule 
 & \multicolumn{3}{c}{Precision $\downarrow$} &  \multicolumn{3}{c}{Multimodal GT $\downarrow$} & \multicolumn{1}{c}{Div $\uparrow$} & \multicolumn{1}{c}{Real $\downarrow$} & \multicolumn{4}{c}{Body Real $\downarrow$}\\
&&&&&&&&  & \multicolumn{2}{c}{mean $\downarrow$} & \multicolumn{2}{c}{RMSE $\downarrow$} \\
 \cmidrule(lr){2-4} \cmidrule(lr){5-7} \cmidrule(lr){8-8}  \cmidrule(lr){9-9}
 \cmidrule(lr){10-13}
 Units &  \SI{}{\centi\metre} & \SI{}{\centi\metre} & deg\textdegree & \SI{}{\centi\metre} & \SI{}{\centi\metre} & – & \SI{}{\metre} & –  & \SI{}{\percent} & \SI{}{\percent} & \SI{}{\percent} & \SI{}{\percent} \\

Method & \textit{u}ADE  & \textit{u}FDE  & MAE & \textit{u}MMA & \textit{u}MMF & APDE & \textit{u}APD  & CMD  &   str  & jit & str  & jit\\ %
\midrule
\multirow{1}{*}{ZeroVel}&1.234 & 1.629 & 7.779 & 1.360 & 1.694 & 9.292 & 0.000 & 39.34 & 0.00 & 0.00 & 0.00 & 0.00 \\
ZeroVel+\cite{holden2016deep}&2.829 & 2.862 & 14.969 & 2.847 & 2.872 & 9.292 & 0.000 & 39.34 & 12.92 & 0.00 & 12.92 & 0.00 \\
\multirow{1}{*}{TPK~\cite{walker2017pose}+\cite{holden2016deep}}& 14.24 & 15.32 & 19.87 & 14.52 & 15.26 & 2.321 & 1.465 & 22.66 & 30.88 & 0.32 & 32.83 & 0.49 \\
\multirow{1}{*}{DLow~\cite{yuan2020dlow}+\cite{holden2016deep}}& 13.85 & 14.83 & 19.86 & 14.19 & 14.81 & 4.941 & 2.593 & 21.03 & 31.24 & 0.35 & 33.44 & 0.53 \\
\multirow{1}{*}{GSPS~\cite{mao2021generating}+\cite{holden2016deep}}& 11.17 & 12.73 & 13.33 & 11.88 & 12.94 & 7.578 & 2.915 & 23.60 & 22.37 & 0.24 & 23.89 & 0.35 \\
\multirow{1}{*}{DivSamp~\cite{dang2022diverse}+\cite{holden2016deep}}& 11.20 & 12.84 & 13.37 & 11.91 & 12.95 & 8.611 & \textbf{3.289} & 21.05 & 21.45 & 0.26 & 23.49 & 0.37 \\
\multirow{1}{*}{BeLFusion~\cite{barquero2023belfusion}+\cite{holden2016deep}}& 10.94 & 12.59 & 12.77 & 11.69 & 12.77 & 3.233 & 1.159 & 24.60 & 19.08 & 0.19 & 20.65 & 0.27 \\
\multirow{1}{*}{CoMusion~\cite{suncomusion}+\cite{holden2016deep}}& 12.37 & 13.59 & 18.94 & 13.12 & 13.76 & \underline{2.065 }& \underline{1.992} & 18.00 & 30.70 & 0.48 & 32.25 & 0.70 \\
SkelDiff~\cite{curreli2025nonisotropic}+\cite{holden2016deep}&11.88 & 13.75 & 19.72 & 12.71 & 13.97 & 2.080 & 1.643 & \underline{15.66} & 26.50 & 0.29 & 28.16 & 0.42 \\
\midrule
\midrule
\methodname~(H)& 9.71 & 11.39 & 7.88 & 10.81 & 11.86 & 3.700 & 1.070 & 23.39 & \textbf{0.00} & \textbf{0.00} & \textbf{0.00} & \textbf{0.00} \\
\methodname~(N)& \underline{9.27} & \underline{10.72} & \underline{7.52} & \underline{10.66} & \underline{11.39} & \bfseries{1.981} & 1.675 & \bfseries{13.19} & \textbf{0.00} & \textbf{0.00} & \textbf{0.00} & \textbf{0.00} \\
\methodname~(H+N)& \bfseries{8.76} & \bfseries{10.07} & \bfseries{7.26} & \bfseries{10.15} & \bfseries{10.76} & 2.841 & 1.333 & 17.63 & \textbf{0.00} & \textbf{0.00} & \textbf{0.00} & \textbf{0.00} \\
\bottomrule
\end{tabular}

\end{table*}

 \begin{table*}[!th]\scriptsize	
\centering
\caption{
\textbf{Version of \cref{tab:crossdataset_trainH36MevalAMASS} with conventional metrics instead of unified metrics. 
}
}
\begin{tabular}{H l HHH   rrr rrr r r  rrrr}
\toprule 
 & &\multicolumn{3}{c}{}  &  \multicolumn{3}{c}{Precision $\downarrow$} &  \multicolumn{3}{c}{Multimodal GT $\downarrow$} & \multicolumn{1}{c}{Div $\uparrow$} & \multicolumn{1}{c}{Real $\downarrow$} & \multicolumn{4}{c}{Body Realism $\downarrow$}\\
 \cmidrule(lr){6-8} \cmidrule(lr){9-11} \cmidrule(lr){12-12}  \cmidrule(lr){13-13}
 \cmidrule(lr){14-17}
\multirow{2}{*}{Type}  &\multirow{2}{*}{Method} & & & & &  &&&&&&& \multicolumn{2}{c}{mean $\downarrow$} & \multicolumn{2}{c}{RMSE $\downarrow$} \\

&& \multicolumn{1}{c}{}  & new & $\mathrm{RT}$ & ADE  & FDE  & MAE & MMA & MMF & APDE & APD  & CMD   &   str  & jit & str  & jit\\ %
\midrule

-&ZeroVel&-&\cmark&- & 0.755 & 0.992 & 7.779 & 0.814 & 1.015 & 9.299 & 0.000 & 39.338 & 0.00 & 0.00 & 0.00 & 0.00 \\
\multirow{12}{*}{VAE}&\multirow{1}{*}{TPK~\cite{walker2017pose}+\cite{holden2016deep}}&\multirow{1}{*}{H}&\multirow{1}{*}{\xmark}&$\mathrm{RT}_\mathrm{I/O}$ & 0.770 & 0.820 & 19.867 & 0.781 & 0.816 & 2.321 & 7.896 & 22.659 & 30.88 & 0.32 & 32.83 & 0.49 \\
-&\multirow{1}{*}{DLow~\cite{yuan2020dlow}+\cite{holden2016deep}}&\multirow{1}{*}{H}&\multirow{1}{*}{\xmark}&$\mathrm{RT}_\mathrm{I/O}$ & 0.749 & 0.792 & 19.859 & 0.764 & 0.790 & 4.941 & 13.713 & 21.031 & 31.24 & 0.35 & 33.44 & 0.53 \\
-&\multirow{1}{*}{GSPS~\cite{mao2021generating}+\cite{holden2016deep}}&\multirow{1}{*}{H}&\multirow{1}{*}{\xmark}&$\mathrm{RT}_\mathrm{I/O}$ & 0.660 & 0.738 & 13.334 & 0.688 & 0.744 & 7.578 & 16.198 & 23.596 & 22.37 & 0.24 & 23.89 & 0.35 \\
-&\multirow{1}{*}{DivSamp~\cite{dang2022diverse}+\cite{holden2016deep}}&\multirow{1}{*}{H}&\multirow{1}{*}{\xmark}&$\mathrm{RT}_\mathrm{I/O}$ & 0.663 & 0.742 & 13.371 & 0.692 & 0.744 & 8.611 & 17.745 & 21.046 & 21.45 & 0.26 & 23.49 & 0.37 \\
\multirow{9}{*}{DM}&\multirow{1}{*}{BeLFusion~\cite{barquero2023belfusion}+\cite{holden2016deep}}&\multirow{1}{*}{H}&\multirow{1}{*}{\xmark}&$\mathrm{RT}_\mathrm{I/O}$ & 0.649 & 0.734 & 12.770 & 0.680 & 0.739 & 3.233 & 6.512 & 24.604 & 19.08 & 0.19 & 20.65 & 0.27 \\
-&\multirow{1}{*}{CoMusion~\cite{suncomusion}+\cite{holden2016deep}}&\multirow{1}{*}{H}&\multirow{1}{*}{\xmark}&$\mathrm{RT}_\mathrm{I/O}$ & 0.683 & 0.739 & 18.938 & 0.717 & 0.744 & 2.065 & 10.856 & 17.999 & 30.70 & 0.48 & 32.25 & 0.70 \\
-&\multirow{1}{*}{SkelDiff~\cite{curreli2025nonisotropic}+\cite{holden2016deep}}&\multirow{1}{*}{H}&\multirow{1}{*}{\xmark}&$\mathrm{RT}_\mathrm{I/O}$ & 0.668 & 0.752 & 19.722 & 0.701 & 0.760 & 2.080 & 8.956 & 15.660 & 26.50 & 0.29 & 28.16 & 0.42 \\
\midrule
\midrule
-&\methodname~(H)&Hreal&\cmark&- & 0.588 & 0.688 & 7.883 & 0.644 & 0.708 & 3.700 & 6.370 & 23.393 & 0.00 & 0.00 & 0.00 & 0.00 \\
-&\methodname~(N)&Nreal&\cmark&- & 0.564 & 0.653 & 7.516 & 0.635 & 0.682 & 1.981 & 9.418 & 13.185 & 0.00 & 0.00 & 0.00 & 0.00 \\
-&\methodname~(H+N)&H+Nreal&\cmark&- & 0.536 & 0.617 & 7.259 & 0.607 & 0.646 & 2.841 & 7.776 & 17.630 & 0.00 & 0.00 & 0.00 & 0.00 \\
\bottomrule
\end{tabular}
\label{tab:crossdataset_trainH36MevalAMASS_old_metrics}
\end{table*}

\begin{table*}[t]
\tiny	
\centering
\caption{
\textbf{Evaluation of Zero-Shot Kinematics on  H36M($\Kkin{H}$)~\cite{Ionescu2014} with additional retargeting scenario}. Additionally to the scenarios $\mathrm{RT}_\mathrm{I/O}$ presented as default in \cref{tab:crossdataset_trainAMASSevalH36M}, we investigate two additional retargeting scenarios,  $\mathrm{RT}_\mathrm{I/GT}$ and $\mathrm{RT}_{\mathrm{GT}^2}$. See \cref{app:exp:retargeting_scenarios_explained} for their description. 
}
\setlength{\tabcolsep}{1.3pt}
\begin{tabular}{H l HHc   rrr rrr r rr  rrrr}
\toprule 
 & &\multicolumn{3}{c}{}  &  \multicolumn{3}{c}{Precision $\downarrow$} &  \multicolumn{3}{c}{Multimodal GT $\downarrow$} & \multicolumn{1}{c}{Div $\uparrow$} & \multicolumn{2}{c}{Realism $\downarrow$} & \multicolumn{4}{c}{Body Realism $\downarrow$}\\
 \cmidrule(lr){6-8} \cmidrule(lr){9-11} \cmidrule(lr){12-12}  \cmidrule(lr){13-14}
 \cmidrule(lr){15-18}
\multirow{2}{*}{Type}  &\multirow{2}{*}{Method} & & & & & & & &&&&&&\multicolumn{2}{c}{mean $\downarrow$} & \multicolumn{2}{c}{RMSE $\downarrow$} \\

& & \multicolumn{1}{c}{}  & new & $\mathrm{RT}$ & \textit{u}ADE  & \textit{u}FDE  & MAE & \textit{u}MMA & \textit{u}MMF & APDE & \textit{u}APD  & CMD  & FID &   str  & jit & str  & jit\\ %
\midrule

\multirow{4}{*}{Alg}&\multirow{4}{*}{ZeroVel} &\multirow{4}{*}{-}&\multirow{4}{*}{\cmark}&- & 11.77 & 17.88 & 6.753 & 13.74 & 18.56 & 8.085 & 0.000 & 22.822 & - & \bfseries{0.00} & \bfseries{0.00} & \bfseries{0.00} & \bfseries{0.00} \\
&&&&$\mathrm{RT}_\mathrm{I/O}$ & 11.77 & 17.88 & 6.362 & 19.88 & 23.72 & 8.085 & 0.000 & 22.822 & - & 0.05 & \bfseries{0.00} & 0.05 & \bfseries{0.00} \\
&&&&$\mathrm{RT}_{I/\mathrm{GT}}$ & 11.39 & 17.09 & 6.467 & 19.46 & 22.96 & 8.085 & 0.000 & 22.822 & - & 0.52 & \bfseries{0.00} & 0.52 & \bfseries{0.00} \\
&&&&$\mathrm{RT}_{I/\mathrm{GT}^2}$ & 11.77 & 17.88 & 6.362 & 19.88 & 23.72 & 8.085 & 0.000 & 22.822 &- & 0.05 & \bfseries{0.00} & 0.05 & \bfseries{0.00} \\
\cmidrule(lr){3-18}
\multirow{12}{*}{VAE}&\multirow{3}{*}{\shortstack[l]{TPK~\cite{walker2017pose}\\+\cite{holden2016deep}}}&\multirow{3}{*}{A}&\multirow{3}{*}{\xmark}&$\mathrm{RT}_\mathrm{I/O}$ & 13.81 & 16.13 & 22.276 & 14.60 & 16.22 & \underline{1.968} & 1.469 & 10.051 & 3.773 & 19.55 & 0.46 & 22.21 & 0.73 \\
-&&&&$\mathrm{RT}_{I/\mathrm{GT}}$ & 13.60 & 15.85 & 22.324 & 18.20 & 19.07 & \bfseries{1.914} & 1.410 & 10.362 & - & 23.11 & 0.58 & 26.85 & 0.91 \\
-&&&&$\mathrm{RT}_{I/\mathrm{GT}^2}$ & 13.80 & 16.12 & 22.259 & 18.34 & 19.29 & \underline{1.968} & 1.469 & 10.051 & 3.662 & 19.73 & 0.46 & 22.39 & 0.73 \\
\cmidrule(lr){3-18}
-&\multirow{3}{*}{\shortstack[l]{DLow~\cite{yuan2020dlow}\\+\cite{holden2016deep}}}&\multirow{3}{*}{A}&\multirow{3}{*}{\xmark}&$\mathrm{RT}_\mathrm{I/O}$ & 12.71 & 14.80 & 21.887 & 13.60 & 14.97 & 2.060 & 2.060 & 9.204 & 2.875 & 20.26 & 0.53 & 23.47 & 0.82 \\
-&&&&$\mathrm{RT}_{I/\mathrm{GT}}$ & 12.55 & 14.63 & 22.207 & 17.59 & 18.18 & 2.880 & 2.014 & 9.326 & - & 23.83 & 0.65 & 28.08 & 1.01 \\
-&&&&$\mathrm{RT}_{I/\mathrm{GT}^2}$ & 12.70 & 14.80 & 22.011 & 17.71 & 18.37 & 2.060 & 2.060 & 9.204 & 2.620 & 20.38 & 0.53 & 23.60 & 0.82 \\
\cmidrule(lr){3-18}
-&\multirow{3}{*}{\shortstack[l]{GSPS~\cite{mao2021generating}\\+\cite{holden2016deep}}}&\multirow{3}{*}{A}&\multirow{3}{*}{\xmark}&$\mathrm{RT}_\mathrm{I/O}$ & 9.29 & 11.91 & 8.107 & 10.85 & 12.35 & 2.373 & 2.069 & 7.409 & 1.735 & 11.51 & 0.38 & 14.00 & 0.49 \\
-&&&&$\mathrm{RT}_{I/\mathrm{GT}}$ & 9.06 & 11.66 & 7.889 & 16.23 & 16.84 & 3.034 & 1.966 & \underline{7.329} & - & 14.98 & 0.48 & 18.20 & 0.63 \\
-&&&&$\mathrm{RT}_{I/\mathrm{GT}^2}$ & 9.22 & 11.86 & 7.008 & 16.50 & 17.15 & 2.386 & 2.070 & 7.403 & 1.620 & 11.33 & 0.38 & 13.84 & 0.49 \\
\cmidrule(lr){3-18}
-&\multirow{3}{*}{\shortstack[l]{DivSamp~\cite{dang2022diverse}\\+\cite{holden2016deep}}}&\multirow{3}{*}{A}&\multirow{3}{*}{\xmark}&$\mathrm{RT}_\mathrm{I/O}$ & 9.27 & 12.61 & 8.374 & 11.42 & 13.27 & 10.510 & 4.210 & 47.783 & 5.629 & 18.47 & 1.01 & 24.51 & 1.39 \\
-&&&&$\mathrm{RT}_{I/\mathrm{GT}}$ & 9.08 & 12.40 & 8.322 & 17.07 & 18.04 & 13.525 & \bfseries{4.267} & 48.441 & - & 21.30 & 1.19 & 28.65 & 1.65 \\
-&&&&$\mathrm{RT}_{I/\mathrm{GT}^2}$ & 9.20 & 12.59 & 7.572 & 17.34 & 18.34 & 10.519 & \underline{4.213} & 47.852 & 5.082 & 18.33 & 1.02 & 24.42 & 1.40 \\
\cmidrule(lr){3-18}
\multirow{9}{*}{DM}&\multirow{3}{*}{\shortstack[l]{BeLFusion~\cite{barquero2023belfusion}\\+\cite{holden2016deep}}}&\multirow{3}{*}{A}&\multirow{3}{*}{\xmark}&$\mathrm{RT}_\mathrm{I/O}$ & 9.24 & 11.62 & 8.200 & 10.93 & 12.16 & 2.284 & 1.305 & 8.031 & 1.195 & 9.81 & 0.34 & 12.16 & 0.46 \\
-&&&&$\mathrm{RT}_{I/\mathrm{GT}}$ & 9.05 & 11.54 & 9.565 & 16.52 & 17.07 & 2.104 & 1.252 & 8.013 & - & 12.20 & 0.43 & 15.34 & 0.58 \\
-&&&&$\mathrm{RT}_{I/\mathrm{GT}^2}$ & 9.17 & 11.60 & 7.273 & 16.76 & 17.26 & 2.284 & 1.305 & 8.031 & 1.093 & 9.68 & 0.34 & 12.05 & 0.46 \\
\cmidrule(lr){3-18}
-&\multirow{3}{*}{\shortstack[l]{CoMusion~\cite{suncomusion}\\+\cite{holden2016deep}}}&\multirow{3}{*}{A}&\multirow{3}{*}{\xmark}&$\mathrm{RT}_\mathrm{I/O}$ & 10.07 & 12.15 & 21.066 & 12.49 & 12.96 & 2.370 & 2.070 & 8.587 & 1.426 & 15.98 & 0.51 & 17.57 & 0.68 \\
-&&&&$\mathrm{RT}_{I/\mathrm{GT}}$ & 9.95 & 12.00 & 21.730 & 17.20 & 17.11 & 3.354 & 2.029 & 8.664 & - & 21.07 & 0.68 & 23.31 & 0.90 \\
-&&&&$\mathrm{RT}_{I/\mathrm{GT}^2}$ & 10.08 & 12.16 & 21.186 & 17.39 & 17.29 & 2.370 & 2.070 & 8.587 & 1.175 & 15.85 & 0.51 & 17.45 & 0.68 \\
\cmidrule(lr){3-18}
-&\multirow{3}{*}{\shortstack[l]{SkelDiff~\cite{curreli2025nonisotropic}\\+\cite{holden2016deep}}}&\multirow{3}{*}{A}&\multirow{3}{*}{\xmark}&$\mathrm{RT}_\mathrm{I/O}$ & 10.81 & 14.99 & 14.947 & 12.75 & 15.42 & 2.995 & 0.992 & 7.616 & 5.252 & 11.25 & 0.28 & 12.86 & 0.39 \\
-&&&&$\mathrm{RT}_{I/\mathrm{GT}}$ & 10.45 & 14.48 & 14.146 & 17.78 & 19.58 & 2.805 & 0.923 & 8.139 & - & 15.14 & 0.32 & 16.89 & 0.46 \\
-&&&&$\mathrm{RT}_{I/\mathrm{GT}^2}$ & 10.73 & 14.93 & 14.722 & 18.14 & 20.12 & 2.995 & 0.992 & 7.616 & 4.906 & 11.08 & 0.28 & 12.69 & 0.39 \\
\midrule
\midrule
\multirow{2}{*}{DM}&\methodname~&A&\cmark&- & \underline{7.86} & \underline{10.47} & \underline{5.861} & \underline{10.66} & \underline{11.61} & 2.248 & 1.973 & \bfseries{7.061} & 0.691 & \bfseries{0.00} & \bfseries{0.00} & \bfseries{0.00} & \bfseries{0.00} \\
-&\methodname~&A+N&\cmark&- & \bfseries{7.71} & \bfseries{10.21} & \bfseries{5.683} & \bfseries{10.58} & \bfseries{11.36} & 2.597 & 1.797 & 7.349 & \bfseries{0.504} &\bfseries{0.00} & \bfseries{0.00} & \bfseries{0.00} & \bfseries{0.00} \\

\bottomrule
\end{tabular}

\label{tab:crossdataset_trainAMASSevalH36M_add_retargeting}
\vspace{-0.3cm}
\end{table*}

 \begin{table*}[!th]\tiny	
\centering
\caption{
\textbf{Evaluation of Zero-Shot Kinematics on  AMASS($\Kkin{A}$)~\cite{mahmood2019amass} with additional retargeting scenario}. Additionally to the scenarios $\mathrm{RT}_\mathrm{I/O}$ presented as default in \cref{tab:crossdataset_trainH36MevalAMASS}, we investigate two additional retargeting scenarios,  $\mathrm{RT}_\mathrm{I/GT}$ and $\mathrm{RT}_{\mathrm{GT}^2}$. See \cref{app:exp:retargeting_scenarios_explained} for their description. 
}
\label{tab:crossdataset_trainH36MevalAMASS_add_retargeting}
\begin{tabular}{H l HHc   rrr rrr r r  rrrr}
\toprule 
 & &\multicolumn{3}{c}{}  &  \multicolumn{3}{c}{Precision $\downarrow$} &  \multicolumn{3}{c}{Multimodal GT $\downarrow$} & \multicolumn{1}{c}{Div $\uparrow$} & \multicolumn{1}{c}{Real $\downarrow$} & \multicolumn{4}{c}{Body Real $\downarrow$}\\
 \cmidrule(lr){3-5} \cmidrule(lr){6-8} \cmidrule(lr){9-11} \cmidrule(lr){12-12}  \cmidrule(lr){13-13}
 \cmidrule(lr){14-17}
\multirow{2}{*}{Type}  &\multirow{2}{*}{Method} & & & & & &  &&&&&&\multicolumn{2}{c}{mean $\downarrow$} & \multicolumn{2}{c}{RMSE $\downarrow$} \\

& & \multicolumn{1}{c}{}  & new & $\mathrm{RT}$ & \textit{u}ADE  & \textit{u}FDE  & MAE & \textit{u}MMA & \textit{u}MMF & APDE & \textit{u}APD  & CMD  &   str  & jit & str  & jit\\ %
\midrule
\multirow{4}{*}{Alg}&\multirow{4}{*}{ZeroVel}&\multirow{4}{*}{-}&\multirow{4}{*}{\cmark}& -&1.234 & 1.629 & 7.779 & 1.360 & 1.694 & 9.292 & 0.000 & 39.338 & 0.00 & 0.00 & 0.00 & 0.00 \\
&&&&$\mathrm{RT}_\mathrm{I/O}$  & 2.829 & 2.862 & 14.969 & 2.847 & 2.872 & 9.292 & 0.000 & 39.338 & 12.92 & 0.00 & 12.92 & 0.00 \\
&&&& $\mathrm{RT}_{I/\mathrm{GT}}$& 1.288 & 1.713 & 8.634 & 1.951 &  2.262 & 9.292 & 0.000 & 39.338 & 0.38 & 0.00 & 0.38 & 0.00 \\
&&&&$\mathrm{RT}_{I/\mathrm{GT}^2}$ & 1.232 & 1.626 & \underline{7.548} & 1.859 & 2.143 & 9.292 & 0.000 & 39.338 & 0.98 & 0.00 & 0.98 & 0.00 \\
\cmidrule(lr){3-17}

\multirow{12}{*}{VAE}&\multirow{3}{*}{\shortstack[l]{TPK~\cite{walker2017pose}\\+\cite{holden2016deep}}}&\multirow{3}{*}{H}&\multirow{3}{*}{\xmark}&$\mathrm{RT}_\mathrm{I/O}$ & 14.24 & 15.32 & 19.87 & 14.52 & 15.26 & 2.321 & 1.465 & 22.66 & 30.88 & 0.32 & 32.83 & 0.49 \\
-&&&&$\mathrm{RT}_{I/\mathrm{GT}}$ & 14.13 & 15.42 & 18.25 & 15.58 & 15.85 & 2.581 & 1.531 & 22.68 & 23.99 & 0.35 & 26.51 & 0.53 \\
-&&&&$\mathrm{RT}_{I/\mathrm{GT}^2}$ & 13.65 & 14.82 & 18.84 & 14.98 & 15.35 & 2.321 & 1.465 & 22.66 & 22.76 & 0.33 & 25.08 & 0.50 \\
\cmidrule(lr){3-17}
-&\multirow{3}{*}{\shortstack[l]{DLow~\cite{yuan2020dlow}\\+\cite{holden2016deep}}}&\multirow{3}{*}{H}&\multirow{3}{*}{\xmark}&$\mathrm{RT}_\mathrm{I/O}$ & 13.85 & 14.83 & 19.86 & 14.19 & 14.81 & 4.941 & 2.593 & 21.03 & 31.24 & 0.35 & 33.44 & 0.53 \\
-&&&&$\mathrm{RT}_{I/\mathrm{GT}}$ & 13.67 & 14.82 & 18.27 & 15.35 & 15.35 & 4.069 & 2.707 & 21.05 & 24.49 & 0.38 & 27.36 & 0.58 \\
-&&&&$\mathrm{RT}_{I/\mathrm{GT}^2}$ & 13.22 & 14.27 & 18.76 & 14.79 & 14.95 & 4.941 & 2.593 & 21.03 & 23.17 & 0.35 & 25.78 & 0.54 \\
\cmidrule(lr){3-17}
-&\multirow{3}{*}{\shortstack[l]{GSPS~\cite{mao2021generating}\\+\cite{holden2016deep}}}&\multirow{3}{*}{H}&\multirow{3}{*}{\xmark}&$\mathrm{RT}_\mathrm{I/O}$ & 11.17 & 12.73 & 13.33 & 11.88 & 12.94 & 7.578 & 2.915 & 23.60 & 22.37 & 0.24 & 23.89 & 0.35 \\
-&&&&$\mathrm{RT}_{I/\mathrm{GT}}$ & 10.65 & 12.51 & 10.64 & 17.07 & 17.85 & 6.430 & 3.073 & 24.22 & 14.62 & 0.28 & 16.59 & 0.40 \\
-&&&&$\mathrm{RT}_{I/\mathrm{GT}^2}$ & 10.17 & 11.92 & 8.66 & 16.17 & 16.86 & 7.578 & 2.915 & 23.60 & 13.11 & 0.25 & 14.90 & 0.36 \\
\cmidrule(lr){3-17}
-&\multirow{3}{*}{\shortstack[l]{DivSamp~\cite{dang2022diverse}\\+\cite{holden2016deep}}}&\multirow{3}{*}{H}&\multirow{3}{*}{\xmark}&$\mathrm{RT}_\mathrm{I/O}$ & 11.20 & 12.84 & 13.37 & 11.91 & 12.95 & 8.611 & \underline{3.289} & 21.05 & 21.45 & 0.26 & 23.49 & 0.37 \\
-&&&&$\mathrm{RT}_{I/\mathrm{GT}}$ & 10.53 & 12.44 & 11.35 & 17.30 & 18.16 & 7.234 & \bfseries{3.457} & 21.56 & 11.82 & 0.28 & 14.37 & 0.40 \\
-&&&&$\mathrm{RT}_{I/\mathrm{GT}^2}$ & 10.15 & 11.99 & 9.03 & 16.57 & 17.44 & 8.611 & \underline{3.289} & 21.05 & 11.53 & 0.26 & 13.98 & 0.37 \\
\cmidrule(lr){3-17}
\multirow{9}{*}{DM}&\multirow{3}{*}{\shortstack[l]{BeLFusion~\cite{barquero2023belfusion}\\+\cite{holden2016deep}}}&\multirow{3}{*}{H}&\multirow{3}{*}{\xmark}&$\mathrm{RT}_\mathrm{I/O}$ & 10.94 & 12.59 & 12.77 & 11.69 & 12.77 & 3.233 & 1.159 & 24.60 & 19.08 & 0.19 & 20.65 & 0.27 \\
-&&&&$\mathrm{RT}_{I/\mathrm{GT}}$ & 10.35 & 12.31 & 9.84 & 16.37 & 17.15 & 3.615 & 1.229 & 24.49 & 10.07 & 0.22 & 12.10 & 0.31 \\
-&&&&$\mathrm{RT}_{I/\mathrm{GT}^2}$ & 9.90 & 11.72 & 8.26 & 15.53 & 16.21 & 3.233 & 1.159 & 24.60 & 9.91 & 0.20 & 11.88 & 0.29 \\
\cmidrule(lr){3-17}
-&\multirow{3}{*}{\shortstack[l]{CoMusion~\cite{suncomusion}\\+\cite{holden2016deep}}}&\multirow{3}{*}{H}&\multirow{3}{*}{\xmark}&$\mathrm{RT}_\mathrm{I/O}$ & 12.37 & 13.59 & 18.94 & 13.12 & 13.76 & 2.065 & 1.992 & 18.00 & 30.70 & 0.48 & 32.25 & 0.70 \\
-&&&&$\mathrm{RT}_{I/\mathrm{GT}}$ & 12.15 & 13.53 & 17.61 & 15.11 & 15.09 & \bfseries{1.840} & 2.089 & 18.02 & 22.46 & 0.54 & 24.57 & 0.79 \\
-&&&&$\mathrm{RT}_{I/\mathrm{GT}^2}$ & 11.67 & 13.00 & 17.51 & 14.70 & 14.77 & 2.065 & 1.992 & 18.00 & 23.04 & 0.50 & 24.82 & 0.73 \\
\cmidrule(lr){3-17}
-&\multirow{3}{*}{\shortstack[l]{SkelDiff~\cite{curreli2025nonisotropic}\\+\cite{holden2016deep}}}&\multirow{3}{*}{H}&\multirow{3}{*}{\xmark}&$\mathrm{RT}_\mathrm{I/O}$ & 11.88 & 13.75 & 19.72 & 12.71 & 13.97 & 2.080 & 1.643 & 15.66 & 26.50 & 0.29 & 28.16 & 0.42 \\
--&&&&$\mathrm{RT}_{I/\mathrm{GT}}$ & 11.40 & 13.52 & 18.06 & 15.13 & 15.43 & 2.190 & 1.745 & \underline{15.23} & 21.27 & 0.34 & 23.32 & 0.49 \\
--&&&&$\mathrm{RT}_{I/\mathrm{GT}^2}$ & 11.10 & 13.15 & 18.48 & 14.55 & 14.75 & 2.080 & 1.643 & 15.66 & 21.29 & 0.30 & 23.06 & 0.43 \\
\midrule
\midrule
-&\methodname~(H)&Hreal&\cmark&- & 9.71 & 11.39 & 7.88 & 10.81 & 11.86 & 3.700 & 1.070 & 23.39 & \underline{0.00} & 0.00 & 0.00 & 0.00 \\
-&\methodname~(N)&Nreal&\cmark&- & \underline{9.27} & \underline{10.72} & \underline{7.52} & \underline{10.66} & \underline{11.39} & \underline{1.981} & 1.675 & \bfseries{13.19} & 0.00 & \underline{0.00} & \underline{0.00} & \underline{0.00} \\
-&\methodname~(H+N)&H+Nreal&\cmark&- & \bfseries{8.76} & \bfseries{10.07} & \bfseries{7.26} & \bfseries{10.15} & \bfseries{10.76} & 2.841 & 1.333 & 17.63 & \bfseries{0.00} & 0.00 & 0.00 & 0.00 \\
\bottomrule
\end{tabular}

\vspace{-0.3cm}
\end{table*}

 \clearpage

\subsection{Analysis on Retargeting}
\label{app:exp:retargeting_scenarios_explained}

\subsubsection{An Upper Bound for Baseline's zero-shot Performance.}
Isolating the retargeting error from the baseline error completely is not possible, but via triangle inequality we estimate an upper bound: $\mathcal{E}(\text{H}\mid \text{A}) \;\leq\;\; \Delta_{\mathrm{RT}}(\Kkin{H}\rightarrow \Kkin{A}) + \mathcal{E}(\text{A}\mid \text{A})$. In other words, he total error must be lower than the GT retargeting error summed with the SHMP baseline error. We compute it for SkelDiff and obtain $\mathcal{E}(\text{H}\mid \text{A}) \;\leq\;\; \Delta_{\mathrm{RT}}(\Kkin{H}\rightarrow \Kkin{A}) + \mathcal{E}(\text{A}\mid \text{A}) =11.77 + 10$ where the first number comes from the ADE of the retargeted GT in \cref{tab:upper_bound} and the second from SkelDiff evaluated on the train kinematics $\Kkin{A}$ (\cref{tab:amass_intra}. Here  $\Delta_{\mathrm{RT}}(\Kkin{H}\rightarrow \Kkin{A}))$ is computed by applying a full retargeting cycle to all GT sequences of the test split as $\text{RT}_{A\rightarrow H}(\text{RT}_{H\rightarrow A}(\Kkin{H})$ and measuring their reconstruction error.

\begin{table}[h]\scriptsize	
\centering
    \setlength{\tabcolsep}{1.5pt}

\begin{tabular}{lcccccccc}
\toprule
Method & uADE & uFDE & MAE & uAPD & CMD & FID & str & jit \\
\midrule
Our variance over 3 seeds ($\sigma^2$) & $4 e^{-4}$ & $7.95 e^{-4}$ & 0.008 & 0.012 & 0.011 & 0.003 & 0.000 & 0.000\\
\midrule
\midrule
GT Error $\Delta_{\mathrm{RT}}(\Kkin{H}\rightarrow \Kkin{A})$ & 11.77 & 17.88 & 6.362 & 0.0 & 22.82 & 0.606 & 5.22 & 0.0 \\
\bottomrule
\end{tabular}
 \vspace{+0.4cm}
\caption{We report the variance of our main model of \cref{tab:crossdataset_trainAMASSevalH36M} for zero-shot kinematics on H36M and the reconstruction error of the GT for the retargeting procedure on the same scenario.}
\label{tab:upper_bound}
\vspace{-1.4cm}
\end{table}

\subsubsection{Additional Retargeting scenarios}
Additionally to the most straightforward retargeting scenario discussed in the main paper body, we investigate two additional ones.  
\paragraph{Additional Retargeting Scenarios.}
 To facilitate the following discussion, we refer to $\Kkin{\mathrm{GT}}$ as the skeleton kinematics of the experiment dataset and to $\Kkin{\mathrm{NN}}$ as the one actually adopted by the network during training (i.e. fixed for current approaches). When the two do not agree, the most correct approach to simulate a real-life scenario is to first convert the input to the $\Kkin{\mathrm{NN}}$ topology, pass it through the network, and then convert it back the output to $\Kkin{\mathrm{GT}}$, such that it can be used to compute our metrics.  This is the approach we follow in the main body, and we refer to this approach as $\mathrm{RT}_{\mathrm{I/O}}$. It simulates an actual applicative scenario, where the $\Kkin{\mathrm{GT}}$ specifies both the input and the target domain. However, the network's output topology $\Kkin{\mathrm{NN}}$ can be sufficient for some downstream applications, regardless of $\Kkin{\mathrm{GT}}$. Hence, we propose $\mathrm{RT}_{\mathrm{I/GT}}$, where the network's output is stored in $\Kkin{\mathrm{NN}}$, and the ground-truth future is instead retargeted to compute the metrics. Finally, we also consider that the retargeting function is not bijective and projects skeletons into a subspace. To isolate this effect from evaluation, we propose $\mathrm{RT}_{\mathrm{GT}^2}$: additionally to applying $\mathrm{RT}_{\mathrm{I/O}}$, we  retarget the ground truth twice (from $\Kkin{\mathrm{GT}}$ to $\Kkin{\mathrm{NN}}$ and back to $\Kkin{\mathrm{GT}}$). This way, both prediction and GT undergo the same retargeting procedure and belong to the same representation space. 

\paragraph{Evaluation on Zero-Shot Kinematics on H36M}. We present here in \cref{tab:crossdataset_trainAMASSevalH36M_add_retargeting} the same experiment of \cref{tab:crossdataset_trainAMASSevalH36M} but with additional retargeting scenarios. Here $\mathrm{RT}_{\mathrm{I/O}}$ is coincident with \cref{tab:crossdataset_trainAMASSevalH36M}. We see that the scenario $\Kkin{\mathrm{GT}}$ consistently delivers the lowest precision error, this is thus the most favourable setup for the model. We are evaluating in the output space of the model with a GT converted from  H36M to AMASS: the model projects the degraded input to a rather stable distribution - the one learned at train time - and the output is not further converted or degraded. The degradation resulting from the preprocessing retargeting (before feeding the input to the network) is not reflected in the output linearly, as shown by the dissimilarity of ca. 10cm to the degraded GT. Overall, in our experiments, it is not possible to  decouple the error of the SHMP model and the retargeting error, as a GT in the desired kinematics does not exist. The other case, $\mathrm{RT}_{\mathrm{GT}^2}$, performs similarly to $\mathrm{RT}_{\mathrm{I/O}}$, which is expected: the conversion error in this direction for a GT sequence amounts to 2.27mm (since feet for AMASS are inserted and then removed). 
\paragraph{Evaluation on Zero-Shot Kinematics on AMASS}. We present here in \cref{tab:crossdataset_trainH36MevalAMASS_add_retargeting} the same experiment of \cref{tab:crossdataset_trainH36MevalAMASS} but with additional retargeting scenarios. Here $\mathrm{RT}_{\mathrm{I/O}}$ is coincident with \cref{tab:crossdataset_trainH36MevalAMASS}. In this experiment setting, the retargeting error is easier on the networks: the input kinematics AMASS is strongly cropped to fit the number of joints in H36M, thus the input is more similar to the distribution seen by the network at train time. When retargeting both the model output and the GT to AMASS, newly added joints as the feet exhibit similar behavior in the two cases, thus $\mathrm{RT}_{\mathrm{GT}^2}$ delivers the lowest error.

\begin{table*}[t]\fontsize{6}{6}\selectfont
  \centering
  \caption{Comparison on Human3.6M \cite{Ionescu2014}. Bold and underlined results correspond to the best and second-best results among the diffusion based models (DM), respectively.}
\setlength{\tabcolsep}{1.2pt}
\begin{tabular}{l  l Hc  rrr rr  r rr   rrrr}
\toprule 
\multicolumn{2}{c}{} &  \multicolumn{2}{c}{} & \multicolumn{3}{c}{Precision $\downarrow$} & \multicolumn{2}{c}{MM GT $\downarrow$} & \multicolumn{1}{c}{Div $\uparrow$} & \multicolumn{2}{c}{Real $\downarrow$} & \multicolumn{4}{c}{Body Realism $\downarrow$} \\
 \cmidrule(lr){5-7} \cmidrule(lr){8-9} \cmidrule(lr){10-10}  \cmidrule(lr){11-12} \cmidrule(lr){13-16}
\multirow{2}{*}{}  & \multirow{2}{*}{Method} & \multirow{2}{*}{} & \multirow{2}{*}{$\Kkin{}$}
&  & & & &&&&  &\multicolumn{2}{c}{mean $\downarrow$} & \multicolumn{2}{c}{RMSE $\downarrow$} \\
&   
& & new  
 & ADE  & FDE  & MAE  &MMA  & MMF   & APD  & CMD & FID  & str  & jit & str  & jit\\ %
\midrule 
Alg & ZeroVelocity & - & \cmark  &  0.597 & 0.884 & 6.753 & 0.683 
 & 0.909 & 0.000 &  22.812 & 0.606 &  0.00 &  0.00 &  0.00 &  0.00\\
\midrule
\multirow{5}{*}{VAE}
& TPK \cite{walker2017pose} & H & \xmark  & 0.461 & 0.560 & 8.056 & 0.522 & 0.569 & 6.723 & 6.326 & 0.538 & 6.69 & 0.24 & 8.37 & 0.31 \\  %
& DLow \cite{yuan2020dlow}  & H & \xmark   & 0.425 & 0.518  & 6.856 &  0.495 & 0.531 & 11.741 & {4.927} & 1.255 & 7.67 & 0.28 & 9.71 & 0.36 \\
& GSPS \cite{mao2021generating} & H & \xmark   & 0.389 & 0.496 & 7.171 & 0.476 & 0.525  & {14.757} & 10.758 & 2.103 & 4.83 & 0.19 & 6.17 & 0.24 \\
& DivSamp \cite{dang2022diverse}  & H & \xmark  & {0.370} & 0.485 & 6.257  & {0.475} & {0.516}  & {15.310} & 11.692 & 2.083 & 6.16 & 0.23 & 7.85 & 0.29 \\
\midrule
\multirow{ 4}{*}{DM}
& HumanMAC \cite{chen2023humanmac}  & H & \xmark   & 0.369 & 0.480 & 6.167 & 0.509 & 0.545 & 6.301 & - & - & \underline{4.01} & 0.46 & 6.04 & 0.57\\ 
& BeLFusion \cite{barquero2023belfusion} & H & \xmark   & 0.372 & {0.474} & 6.107 & \textbf{{0.473}} & \underline{0.507}  & 7.602 & 5.988 & {0.209} & 5.39 & \underline{0.17} & 6.63 & \underline{0.22} \\
 & CoMusion \cite{suncomusion} & H & \xmark  & {0.350} & {0.458} & 5.904 & 0.494 & \textbf{0.506}  & \underline{7.632} & \textbf{3.202} & \textbf{0.102} &  4.61 & 0.41 & \underline{5.97} & 0.56 \\ 
&  SkelDiff~\cite{curreli2025nonisotropic} & H & \xmark  & \underline{0.344} & \textbf{0.450} & 5.556 & \underline{0.487} & 0.512 & 7.249 & \underline{4.178} & {0.123} &  \textbf{3.90} & \textbf{0.16}& \textbf{4.96} & \textbf{0.21}\\ 
 \cmidrule{1-16}\morecmidrules\cmidrule{1-16}

\multirow{3}{*}{DM}&\methodname(A+N)&A+N&\cmark & 0.395 & 0.522 & 5.683 &  0.533 & 0.574 &\textbf{8.492} & 7.349 & 0.504 & \textbf{0.00} & \textbf{0.00} & \textbf{0.00} & \textbf{0.00} \\

&\methodname(A+N+H)&A+N+H&\cmark & \textbf{0.347} & 0.456 & \underline{5.121} & 0.493 & 0.519 & 7.086 & 7.355 & \underline{0.105} & \textbf{0.00}& \textbf{0.00} & \textbf{0.00} & \textbf{0.00} \\
&\methodname(H)&H&\cmark & 0.351 & \underline{0.451} & \textbf{5.051} & 0.491 & 0.515 & 6.501 & 7.730 & 0.158 & \textbf{0.00}& \textbf{0.00} & \textbf{0.00} & \textbf{0.00} \\

\bottomrule
\end{tabular}

\label{tab:h36m_intra}
\end{table*}

\subsection{Single-Kinematics: AMASS, Nymeria, H36M}
\label{app:exp-same-topo}
Here we report results of methods trained and tested on the same kinematics (i.e. dataset), as in prior works\cite{barquero2023belfusion, yuan2020dlow, curreli2025nonisotropic, dang2022diverse, suncomusion, chen2023humanmac}. 

\paragraph{H36M}. In \cref{tab:h36m_intra}, we report evaluation results on the H36M dataset\cite{Ionescu2014}. It is remarkable that, while the main focus of our work is on enabling zero-shot kinematics processing, our method achieves very competitive results. We also observe that incorporating further datasets in this case is less beneficial in terms of precision, as network capacity is used to represent different distributions. Instead, it still provides improvements in the diversity of the generated movements. This demonstrates that our network is capable of exploiting the combination of different data priors to generate other realistic hypotheses. Particularly, it can leverage the very diverse prior of AMASS to novel kinematics distributions with high realism and precision. We believe this fact may be significant for further experiments investigating the effect of different training distributions. 

\paragraph{AMASS}. Following previous works, we also employ the AMASS cross-dataset evaluation protocol \cite{barquero2023belfusion, suncomusion, chen2023humanmac, curreli2025nonisotropic}. In \cref{tab:amass_intra}, We achieve competitive results across all metrics. Interestingly, this is the only dataset where leveraging more data or multiple data priors does not improve quantitative evaluation. We believe this is an indicator of the very high motion diversity present in the AMASS distribution compared to other datasets.

\begin{table*}[th!]\fontsize{6}{6}\selectfont
\centering
\caption{Quantitative results for AMASS dataset \cite{mahmood2019amass}. Not all metrics are available for HumanMAC(see \cref{app:exp_settings:baselines}). The best results are highlighted in \textbf{bold}, second-best are \underline{underlined}. The symbol `-' indicates that the results are not reported in the baseline work.} 
\setlength{\tabcolsep}{0.7pt}
\begin{tabular}{c  l Hc  rrr rrr  rr   rrrr}
\toprule 
\multicolumn{2}{c}{} &  \multicolumn{2}{c}{} & \multicolumn{3}{c}{Precision $\downarrow$} & \multicolumn{3}{c}{MM GT $\downarrow$} & \multicolumn{1}{c}{Div $\uparrow$} & \multicolumn{1}{c}{Real $\downarrow$} & \multicolumn{4}{c}{Body Realism $\downarrow$} \\
 \cmidrule(lr){5-7} \cmidrule(lr){8-10} \cmidrule(lr){11-11}  \cmidrule(lr){12-12} \cmidrule(lr){13-16}
\multirow{2}{*}{Type}  & \multirow{2}{*}{Method} & \multirow{2}{*}{train} & \multirow{2}{*}{$\Kkin{}$}
&  & & & &&&& & \multicolumn{2}{c}{mean $\downarrow$} & \multicolumn{2}{c}{RMSE $\downarrow$} \\
&   
& &  
 new & ADE  & FDE  & MAE  &MMA  & MMF  & APDE & APD  & CMD   & str  & jit & str  & jit\\ %
\midrule

Alg & ZeroVelocity  & - & \cmark &0.755 & 0.992 & 7.779 & 0.814 & 1.015 & - &  0.000 & 39.262 &  0.00 &  0.00 &  0.00 &  0.00\\
\midrule
\multirow{4}{*}{VAE}
& TPK \cite{walker2017pose} & A & \xmark & 0.656 & 0.675 & 10.191 & 0.658 & 0.674 & 2.265 & 9.283  & 17.127 & 7.34 & 0.34 &  9.69 & 0.48 \\ %
& DLow \cite{yuan2020dlow} & A & \xmark & 0.590 & 0.612 & 8.510 & 0.618 & 0.617& 4.243 & {13.170}  & {15.185} & 8.41 & 0.40 & 11.06 & 0.58\\ 
& GSPS \cite{mao2021generating} & A & \xmark & 0.563 & 0.613 & 9.045 & 0.609 & 0.633 & 4.678 & 12.465  & 18.404 & 6.65 & 0.29 & 8.98 & 0.37\\%
& DivSamp \cite{dang2022diverse} & A & \xmark & 0.564 & 0.647 & 8.027 & 0.623 & 0.667& 15.837 & \textbf{24.724}  & 50.239 & 11.17 & 0.82 & 16.71 & 1.0\\ %
\midrule
\multirow{3}{*}{DM}
& HumanMAC \cite{chen2023humanmac} & A & \xmark & 0.511 & 0.554 & - & 0.593 & 0.591& - & 9.321  & - & - & -& - & -\\
& BeLFusion \cite{barquero2023belfusion} & A & \xmark & {0.513} & {0.560} & 7.125 & {0.569} & {0.585} & \textbf{1.977} & 9.376 & 16.995 & 7.19 & 0.34 & 9.03 & 0.34\\ %
 & CoMusion \cite{suncomusion} & A & \xmark & \underline{0.494} & \underline{0.547} & 6.715 & \textbf{0.469} & \textbf{0.466}& 2.328 & \underline{10.848}  & \textbf{9.636} & 4.04 & 0.25& 5.63 & 0.52\\
 &  SkelDiff \cite{curreli2025nonisotropic} & A & \xmark & \bfseries 0.480 & \bfseries 0.545 & \bfseries 6.124 & \underline{0.561} & \underline{0.580} & \underline{2.067} & 9.456 & \underline{11.417} & 3.15 &  0.20 & 4.45 & 0.26 \\  %
\midrule
\midrule
\multirow{3}{*}{DM}&\methodname(A+N)~&A+N&\cmark & 0.504 & 0.573 & 6.314 & 0.582 & 0.608 & 2.568 & 8.055 & 15.450 & \bfseries{0.00}  & \bfseries{0.00} & \bfseries{0.00} & \bfseries{0.00} \\
&\methodname(A+N+H)&A+N+Hreal&  \cmark & 0.508 & 0.574 & 6.337 & 0.585 & 0.609 & 2.485 & 8.272 & 14.926 &  \bfseries{0.00}  & \bfseries{0.00} & \bfseries{0.00} & \bfseries{0.00} \\
&\methodname(A)&Areal&\cmark & 0.496 & 0.560 & 6.214 & 0.576 & 0.596 & 2.518 & 8.241 & 13.097 & \bfseries{0.00}  & \bfseries{0.00} & \bfseries{0.00} & \bfseries{0.00} \\

\bottomrule
\end{tabular}

\label{tab:amass_intra}
\end{table*}

\begin{table*}[th!]\scriptsize	
\centering
\caption{Quantitative results for Nymeria \cite{ma2024nymeria} dataset. We trained the latest diffusion baselines with code available following their configuration for AMASS, as the datasets are comparable in size. See \cref{app:exp_settings:baselines} for details. As our method has no limb stretching by definition of the motion parametrization, 0.12\% corresponds to the stretching present in the GT data due to minor sensor inaccuracies.  } 
\begin{tabular}{c  l Hc   r r r r r  r r r r}
\toprule 
\multicolumn{2}{c}{} & & & \multicolumn{3}{c}{Precision $\downarrow$}  & \multicolumn{1}{c}{Div  $\uparrow$} & \multicolumn{1}{c}{Real $\downarrow$} & \multicolumn{4}{c}{Body Realism $\downarrow$}\\
 \cmidrule(lr){5-7} \cmidrule(lr){9-9} \cmidrule(lr){8-8}  \cmidrule(lr){10-13}
\multirow{2}{*}{Type}  & \multirow{2}{*}{Method} & \multirow{2}{*}{trained on} & \multirow{2}{*}{$\Kkin{}$}
&  & & & & &  \multicolumn{2}{c}{mean $\downarrow$} & \multicolumn{2}{c}{RMSE $\downarrow$} \\
&   
& &  
 new & ADE  & FDE  & MAE  & APD & CMD  &  str  & jit & str  & jit\\ %
\midrule
Alg & ZeroVelocity &  - &  \cmark & 0.519 & 0.698 & 4.608 &     0.0 & 23.344 &  0.14  & 0.0 & 0.14  &  0.0 \\
\midrule
\multirow{3}{*}{DM} 

 & Belfusion~\cite{barquero2023belfusion} & N & \xmark & 0.343 & 0.419 & {4.318} & 5.280 & - & 4.76 & {0.13} & 5.69 & {0.16} \\
 & HumanMAC~\cite{chen2023humanmac} & N & \xmark & {0.318} & {0.403} & 4.547 & {5.689} & - &{2.37} & 0.50 & {4.59} & 0.62 \\
 & SkelDiff~\cite{curreli2025nonisotropic} & N & \xmark & \bfseries{0.278} & \bfseries{0.359} & \bfseries{3.227} & \bfseries{6.450} &  {4.267} &{1.78} & {0.09} & {2.34} & {0.12} \\

\midrule
\midrule
\multirow{3}{*}{DM}

&\methodname~(A+N)&A+N&\cmark & 0.299 & 0.375 & 3.373 & {6.016} & 4.475 & \textbf{0.12} & \textbf{0.00} & \textbf{0.12} & \textbf{0.00} \\
&\methodname~(A+N+H)&A+N+H&\cmark & 0.302 & 0.376 & 3.409 & \underline{6.399} & \underline{4.087} &\textbf{0.12} & \textbf{0.00} & \textbf{0.12} & \textbf{0.00} \\
&\methodname~(N)&N&\cmark & \underline{0.293} & \underline{0.369 }& \underline{3.305} & 6.358 & \textbf{ 3.818} & \textbf{0.12} & \textbf{0.00} & \textbf{0.12} & \textbf{0.00} \\
\bottomrule
\end{tabular}

\label{tab:nymeria_intra}
\end{table*}

\paragraph{Nymeria}. We train and evaluate latest diffusion baselines whose code was available on the Nymeria dataset in \cref{tab:nymeria_intra}. Our method performs on par with previous works and achieves significantly better realism. Comparing the range of ADE among AMASS and Nymeria, for example, on the ZeroVelocity algorithmic baseline, it is evident that the motion quality of Nymeria is overall more static.

 \begin{table*}[!th]\scriptsize	
\centering
\caption{\textbf{Zero-shot kinematics on H36M($\Kkin{H}$)~\cite{Ionescu2014} without unified metrics}. Version of \cref{tab:crossdataset_trainAMASSevalH36M} with conventional metrics instead of unified metrics. 
}
\setlength{\tabcolsep}{0.7pt}
\begin{tabular}{H l   rrr rrr r rr  rrrr}
\toprule 
 & &  \multicolumn{3}{c}{Precision $\downarrow$} &  \multicolumn{3}{c}{Multimodal GT $\downarrow$} & \multicolumn{1}{c}{Div $\uparrow$} & \multicolumn{2}{c}{Realism $\downarrow$} & \multicolumn{4}{c}{Body Realism $\downarrow$}\\
 \cmidrule(lr){3-5} \cmidrule(lr){6-8} \cmidrule(lr){9-9} \cmidrule(lr){10-11}  \cmidrule(lr){12-15}
\multirow{2}{*}{Type}  &\multirow{2}{*}{Method} & & & & & & & &&&\multicolumn{2}{c}{mean $\downarrow$} & \multicolumn{2}{c}{RMSE $\downarrow$} \\

&  & ADE  & FDE  & MAE & MMA & MMF & APDE & APD  & CMD  & FID &   str  & jit & str  & jit\\ %
\midrule
Alg & ZeroVel  &  0.597 & 0.884 & 6.753 & 0.683 
 & 0.909 & 8.085 & 0.000 &  22.812 &  0.606 &  0.00 &  0.00 &  0.00 &  0.00\\

\multirow{12}{*}{VAE}&TPK~\cite{walker2017pose}+\cite{holden2016deep} & 1.154 & 0.983 & 22.686 & 1.155 & 0.987 & 1.968 & 7.221 & 10.051 & 7.522 & 19.44 & 0.46 & 22.10 & 0.73 \\

&DLow~\cite{yuan2020dlow}+\cite{holden2016deep} & 1.094 & 0.948 & 22.272 & 1.096 & 0.951 & 2.060 & 9.683 & 9.204 & 6.192 & 20.12 & 0.53 & 23.34 & 0.82 \\

&GSPS~\cite{mao2021generating}+\cite{holden2016deep} & 1.193 & 1.051 & 11.439 & 1.194 & 1.053 & 2.373 & 9.985 & 7.409 & 5.335 & 11.71 & 0.38 & 14.23 & 0.49 \\

&DivSamp~\cite{dang2022diverse}+\cite{holden2016deep} & 1.285 & 1.120 & 11.854 & 1.282 & 1.123 & 10.510 & 18.576 & 47.783 & 7.749 & 18.63 & 1.01 & 24.68 & 1.39 \\

\multirow{9}{*}{DM}&BeLFusion~\cite{barquero2023belfusion}+\cite{holden2016deep} & 1.226 & 1.029 & 10.957 & 1.225 & 1.033 & 2.284 & 6.483 & 8.031 & 6.579 & 10.52 & 0.34 & 12.84 & 0.46 \\

&CoMusion~\cite{suncomusion}+\cite{holden2016deep} & 1.221 & 1.028 & 22.521 & 1.218 & 1.032 & 2.370 & 9.926 & 8.587 & 5.249 & 15.97 & 0.51 & 17.55 & 0.68 \\

&SkelDiff~\cite{curreli2025nonisotropic}+\cite{holden2016deep} & 1.372 & 1.193 & 17.232 & 1.371 & 1.194 & 2.995 & 5.420 & 7.616 & 7.146 & 11.25 & 0.28 & 12.86 & 0.39 \\
\midrule
\midrule
\multirow{2}{*}{DM}&\methodname(A)& 0.403 & 0.533 & 5.861 & 0.536 & 0.585 & 2.248 & 9.320 & 7.061 & 0.691 & 0.00 & 0.00 & 0.00 & 0.00 \\
&\methodname(A+N) & 0.395 & 0.522 & 5.683 & 0.533 & 0.574 & 2.597 & 8.492 & 7.349 & 0.504 & 0.00 & 0.00 & 0.00 & 0.00 \\

\bottomrule
\end{tabular}
\
\label{tab:crossdataset_trainAMASSevalH36M_old_metrics}
\vspace{-0.3cm}
\end{table*}

 \begin{table*}[h!t]\scriptsize	
\centering
\caption{Quantitative results for zero-shot on MoYo for models trained on AMASS. For completeness and future works, we include unified metrics (uADE, uFDE, uAPD). Full metric evaluation of 
\crefcrossfile{tab:moyo}
in main.
} 
\begin{tabular}{  l HH   r r r rr  r rr  r r r r}
\toprule 
 & \multicolumn{2}{c}{} & \multicolumn{5}{c}{Precision $\downarrow$}  & \multicolumn{2}{c}{Div  $\uparrow$} & \multicolumn{1}{c}{Real $\downarrow$} & \multicolumn{4}{c}{Body Realism $\downarrow$}\\
 \cmidrule(lr){2-3}\cmidrule(lr){4-8}  \cmidrule(lr){9-10}  \cmidrule(lr){11-11}\cmidrule(lr){12-15}
\multirow{2}{*}{Method} & \multirow{2}{*}{train} & \multirow{2}{*}{new}
&  & & & & &&&&  \multicolumn{2}{c}{mean $\downarrow$} & \multicolumn{2}{c}{RMSE $\downarrow$} \\
  
&  & 
 & ADE  & FDE  & MAE &uADE & uFDE & APD & uAPD & CMD  &  str  & jit & str  & jit\\ %
\midrule

ZeroVel&-&\cmark & 0.709 & 1.187 & 7.954 & 1.188 & 2.015 & 0.000 & 0.000 & 20.333 & 0.00 & {0.00} & 0.00 & {0.00} \\
SkelDiff&A&\xmark & 0.567 & 0.892 & 8.048 & {0.952} & {1.524} & \bfseries{13.304} & \bfseries{2.454} & 15.710 & 7.25 & 0.29 & 9.44 & 0.41 \\
\midrule
\midrule
\methodname(A+N)&A+N&\cmark & \bfseries{0.492} & \underline{0.786} & \bfseries{6.596} & \underline{0.813} & \underline{1.299} & 12.467 & 2.177 & \bfseries{7.095} & \bfseries{0.00} & \textbf{0.00} & \bfseries{0.00} & \textbf{0.00} \\
\methodname(A)&A&\cmark & \underline{0.501} & \bfseries{0.780} & \underline{6.982} & 0.834 & 1.302 & \underline{13.153} & \underline{2.333} & \underline{11.961} & \bfseries{0.00} & \textbf{0.00} & \bfseries{0.00} & \textbf{0.00} \\
\bottomrule
\end{tabular}

\label{tab:moyo_full}
\end{table*}

\begin{table*}[ht]\scriptsize	
\centering
\caption{{Ablations for the motion parametrization as bone directions on AMASS}. For both methods, our parametrization improves realism and body realism metrics by at least 10\%. Full metric evaluation of 
\crefcrossfile{tab:directions}
in main.} 
\begin{tabular}{  l cH   r r r rr  r rr  r r r r}
\toprule

 & Mot & \multicolumn{1}{c}{}  & \multicolumn{3}{c}{Precision $\downarrow$} & \multicolumn{3}{c}{Multimodal GT $\downarrow$} & \multicolumn{1}{c}{Div $\uparrow$} & \multicolumn{1}{c}{Real $\downarrow$} & \multicolumn{4}{c}{Body Realism $\downarrow$} \\
 \cmidrule(lr){2-2} \cmidrule(lr){3-3}\cmidrule(lr){4-6} \cmidrule(lr){7-9}  \cmidrule(lr){10-10}  \cmidrule(lr){11-11}\cmidrule(lr){12-15}
\multirow{2}{*}{Method} & \multirow{2}{*}{Mot} & \multirow{2}{*}{train}
&&  & & &&&& & \multicolumn{2}{c}{mean $\downarrow$} & \multicolumn{2}{c}{RMSE $\downarrow$} \\
&   
& 
 & ADE  & FDE  & MAE  &MMA  & MMF  & APDE & APD  & CMD   & str  & jit & str  & jit\\ %
 
\midrule

SkelDiff~\cite{curreli2025nonisotropic} & $\motion$&A & \bfseries{0.480} & \bfseries{0.545} & \bfseries{6.124} & \bfseries{0.562} & \bfseries{0.579} & 2.067 & 9.456 & 11.418 & 3.15 & 0.20 & 4.45 & 0.26 \\
SkelDiff~\cite{curreli2025nonisotropic} & $\dir{}{}{}$&A & 0.496 & 0.546 & 6.193 & 0.575 & 0.581 & \bfseries{1.900} & \bfseries{9.960} & \bfseries{9.143} & \bfseries{0.00} & \bfseries{0.00} & \bfseries{0.00} & \bfseries{0.00} \\
\midrule
\methodname~ & $\dir{}{}{}$ & A & 0.501 & 0.561 & 6.551 & 0.577 & \bfseries{0.595} & \bfseries{2.397} & 8.348 & 13.963 & 3.58 & 0.27 & 5.04 & 0.34 \\
\methodname~ & $\motion$ & A & \bfseries{0.498} & \bfseries{0.559} & \bfseries{6.173} & \bfseries{0.577} & 0.596 & 2.489 & \bfseries{8.413} & \bfseries{12.530} & \bfseries{0.00} & \bfseries{0.00} & \bfseries{0.00} & \bfseries{0.00} \\
\bottomrule
\end{tabular}

\label{tab:directions_full}
\end{table*}

\begin{table*}[t]\scriptsize	
\centering
\caption{{Occlusion of a random limb (leg or arm) at inference on AMASS.} CMD metric does not apply as it is related only to the motion distribution of the full joint skeleton. FID is not available for missing joints. When joints are missing, Multimodal GT becomes loosely related and is hence discarded. Extended version of main 
\crefcrossfile{tab:drop_limb}.
} 
\begin{tabular}{l HH   rrr rr r r rrrr}
\toprule 
\multicolumn{1}{c}{} & \multicolumn{2}{c}{}  & \multicolumn{5}{c}{Precision $\downarrow$}  & \multicolumn{2}{c}{Diversity $\uparrow$} & \multicolumn{4}{c}{Body Realism $\downarrow$} \\
 \cmidrule(lr){2-3} \cmidrule(lr){4-8} \cmidrule(lr){9-10} \cmidrule(lr){11-14} 
\multirow{2}{*}{Method} &\multirow{2}{*}{train/eval} & \multirow{2}{*}{new} &&&&&&&&\multicolumn{2}{c}{mean} & \multicolumn{2}{c}{RMSE}
\\
&&& ADE  & FDE  & uADE  & uFDE   & MAE  & APD   & uAPD &  str  & jit & str  & jit  \\ %
\midrule
SkelDiff~\cite{curreli2025nonisotropic} & A & \xmark & - & - & - & - & - & - & - & - & - & - & - \\
SkelDiff~\cite{curreli2025nonisotropic}+rp & A & \xmark & 0.574 & 0.727 & {9.050} & 1.105 & \bfseries{6.996} & 8.890 & 1.486 & 8.15 & 0.27 & 9.95 & 0.39 \\
SkelDiff~\cite{curreli2025nonisotropic}+sl & A & \xmark  & {0.567} & {0.683} & 0.926 & {1.103} & \underline{7.162} & \underline{9.274} & \underline{1.556} & {5.64} & {0.23} & {7.11} & {0.31} \\
\midrule

\midrule
\methodname(A+N)&A+N& \cmark & \bfseries{0.499} & \bfseries{0.553} & \bfseries{0.765} & \bfseries{0.853} & 8.777 & 9.099 & 1.413 & \bfseries{0.00} & \bfseries{0.00} & \bfseries{0.00} & \bfseries{0.00} \\
\methodname(A)&A& \cmark & \underline{0.553} & \underline{0.618} & \underline{0.868} & \underline{0.978} & 10.190 & \textbf{10.152} & \textbf{1.635} & \underline{0.00} & \underline{0.00} & \underline{0.00} & \underline{0.00} \\

\bottomrule
\end{tabular}

\label{tab:drop_limb_full}
\end{table*}

\subsection{Extended Tables from Main and not Unified Metrics}
In this section, we report for transparency and future works the same tables as in the main paper body, but with additional metrics. Since SHMP has a wide spectrum of metrics, many of which correlate, not all metrics were presented in the main paper body due to redundancy and space reasons.
\begin{enumerate}
    \item Conventional Metrics for \cref{tab:crossdataset_trainAMASSevalH36M}, without unified metrics. Here in \cref{tab:crossdataset_trainAMASSevalH36M_old_metrics} we see that ranking is maintained between conventional and unified metrics.
    \item Full metric evaluation for MoYoga of \cref{tab:moyo} can be found in \cref{tab:moyo_full}.
    \item Full metric evaluation for the ablation on the motion parametrization as bone direction in \cref{tab:directions} can be found in \cref{tab:directions_full}.
    \item Full metrics evaluation for occlusion of random limbs on AMASS in \cref{tab:drop_limb} can be found in \cref{tab:drop_limb_full}.
\end{enumerate}

\begin{table}[t]\footnotesize
\centering

\begin{subtable}{\linewidth}
\centering
\setlength{\tabcolsep}{3.8pt}
\vspace{2mm}
\begin{tabular}{l c r r r r r}
\toprule 
 & \multicolumn{1}{c}{Top} & \multicolumn{3}{c}{Precision $\downarrow$} & \multicolumn{1}{c}{Div $\uparrow$} & \multicolumn{1}{c}{Eff $\downarrow$}\\
\cmidrule(lr){2-2}\cmidrule(lr){3-5}\cmidrule(lr){6-6} \cmidrule(lr){7-7}
Component & eval & uADE & uFDE & MAE & uAPD & \#par\\
\midrule

drop3joint30 & \multirow{5}{*}{A} & {0.834} & 0.948 & 6.395 & 1.355 & 8M\\
$L=256$ & & 0.847 & 0.960 & 6.522 & \underline{1.431} & 17M\\
CondTop & & \textbf{0.825} & \bfseries{0.933} & \textbf{6.389} &\textbf{ 1.443} & 8M\\
CondTop+$L=256$ & & 0.846 & 0.959 & \underline{6.396} & {1.301} & 17M\\
\underline{Ours} & & \underline{0.832} & \underline{0.940} & 6.401 & {1.350} & \textbf{8M}\\
\cmidrule(lr){2-7}

drop3joint30 & \multirow{5}{*}{H}  & 0.787 & 1.044 & 5.859 & \underline{2.060} & 8M\\
$L=256$ & & \bfseries{0.769} & \textbf{1.032} & \bfseries{5.650} & 1.826 & 17M\\
CondTop &  & 0.841 & 1.100 & 6.319 & \textbf{2.704} & 8M\\
CondTop+$L=256$ &  & \underline{0.783} & \underline{1.040} & \underline{5.841} & 1.751 & 17M\\
\underline{Ours} &  & 0.801 & 1.055 & 5.844 & 1.888 & \textbf{8M}\\

\bottomrule
\end{tabular}
\caption{Ablations with k=50.}
\end{subtable}

\vspace{4mm}

\begin{subtable}{\linewidth}
\centering

\vspace{2mm}
\begin{tabular}{l c r r r r r}
\toprule 
 & \multicolumn{3}{c}{Precision $\downarrow$} & \multicolumn{1}{c}{Div $\uparrow$} & \multicolumn{1}{c}{Real $\downarrow$} \\
\cmidrule(lr){2-4}\cmidrule(lr){5-5}\cmidrule(lr){6-6}
Component & ADE & FDE & MAE & APD & CMD \\
\midrule

+CondBoneLength    & \underline{0.519} & \underline{0.610} & \bfseries{6.580} & \underline{5.591} & \bfseries{18.093} \\
+posEmbedAdd       & 0.559 & 0.661 & 7.257 & 4.642 & 20.636 \\
+posEmbedConcat    & 0.528 & 0.627 & 6.843 & 5.026 & 19.347 \\
\midrule
\midrule
Ours & \bfseries{0.519} & \bfseries{0.608} & \underline{6.639} & \bfseries{5.610} & \underline{18.188} \\
\bottomrule
\end{tabular}
\caption{Ablations with k=1.}
\end{subtable}
\caption{Ablations for early stages of our model trained on AMASS and Nymeria (A+N) . (A): with k=50. (B): with k=1.}
\label{tab:app:model_validation}
\end{table}

\section{Ablations and Validations on \methodname}
\label{app:ablations}

We validate our model through extensive experiments, investigating the training methodology and the cross-topology application. 
In \cref{tab:app:model_validation} (A), we present ablations for an early stage of our model, trained on AMASS and Nymeria (A+N) with a relaxation of the diffusion objective of k=50 (as our final model) and tested for cross-topology on H36M.

\paragraph{Random Topology Augmentation.}We first investigate in \cref{tab:app:model_validation} (drop3joints30) whether randomly removing up to 3 joints with a probability of 30\% during training increases the cross-topology performance. While the improvement is present, we consider it as minor and not worth the additional component.

\paragraph{Diffusion Conditioning on Topology.} As our model is designed to be topology-agnostic and topology information derives only from the input adjacency matrix, we investigate whether a stronger, explicit conditioning on learned graph topology features strengthens performance in both same- and cross-topology settings. We remark here that such feature computation must be permutation equivariant with respect to both the input motion and adjacency matrix, a not straightforward challenge. Interestingly, we see in \cref{tab:app:model_validation} (CondTop) that this improves the same-dataset performance, but strongly affects the cross-dataset performance negatively. Further attempts to let the learned conditioning generalize to unseen graphs via data augmentation have not led to significant improvements. 

\paragraph{Latent size 96 vs 256.}We increase the latent size from 96 to 256 ($L=256$), for a total of 17M parameters against the previous 8M. This additional capacity translates into worse precision on the seen dataset AMASS, but better precision on generation on H36M. It seems the additional capability results in an overfitting phenomenon with respect to seen "data",  but not seen "topologies". In combination with conditioning the diffusion model on permutation equivariance topology features (CondTop+$L=256$), the increased capacity strongly increases the cross-generalization precision compared to conditioning with less parameters. At equal number of parameters, the conditioning still performs worse. 

\paragraph{AE vs VAE.}In the early stages of our training, we also attempted a variational autoencoder instead of an autoencoder. The results for generation were rather poor, and we discarded the option. Training autoencoder and latent diffusion models together can be challenging, as a good latent space for reconstruction is not directly a good latent space for generation \cite{yao2025reconstruction}. Yao \etal mention indeed that VAE have among the worst generation quality when paired with a latent diffusion model.

\paragraph{Anisotropic vs isotropic diffusion.} We decided against the recent anisotropic diffusion paradigm~\cite {curreli2025nonisotropic}, in contrast to the conventional isotropic training we employed: exploiting correlation in the noise and aiming for permutation equivariance are at two opposite spectra. 

\paragraph{Permutation Equivariant Positional Embeddings.} In \cref{tab:app:model_validation} (b), we ablate against permutation equivariant formulations of positional embeddings \cite{ma2021graph} in encoder and decoder, in the variants of addition and concatenation (posEmbedAdd, posEmbedConcat). We implement an equivariant version of Graphormer's attention bias \cite{ying2021transformers} for the attention layers, a non-trivial procedure as the equivariance constraint must be fulfilled for a matrix and not a vector in this case. We implement a learned and not learned variant, but find that in both cases it does not lead to improved performance and hence do not include it in further experiments.  %

\section{Qualitative Examples}

\begin{figure*}[ht]
  \centering
    \vspace{0.0cm}
    \includegraphics[trim={0cm, 0.cm, 0cm, 0.cm},clip,width=\textwidth]{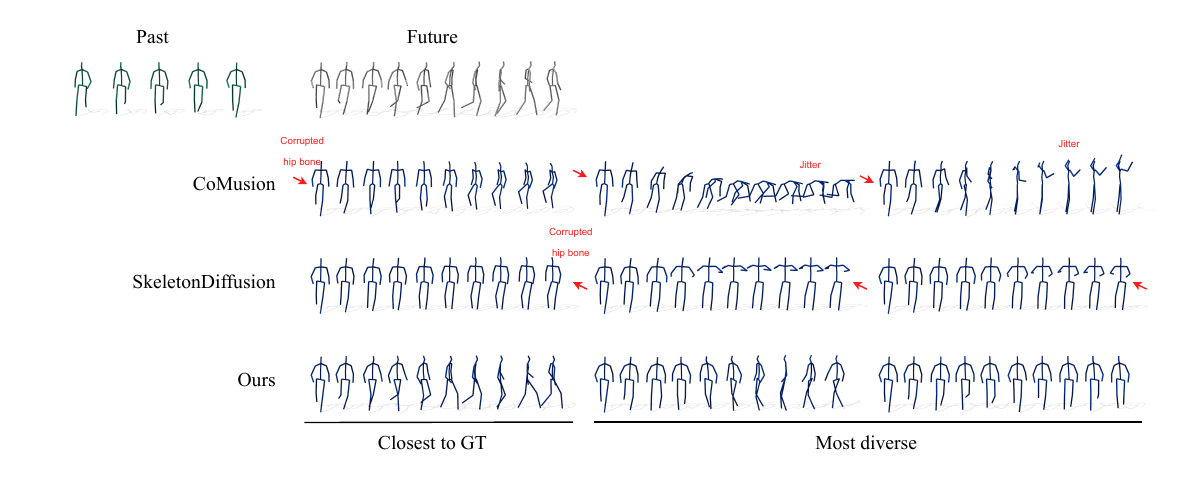}
  \vspace{-0.4cm}
  \caption{Qualitative Results for the cross-topology experiment on H36M of 
  \cref{tab:crossdataset_trainAMASSevalH36M}. 
  We report out of 50 predictions, the sample closest to the GT, and the two predictions that maximize diversity when paired with the closest to GT sample. Segment n. 605.}
  \label{fig:qualitative_retarget1}
  \vspace{-0.0cm}
\end{figure*}

\begin{figure*}[t]
  \centering
    \vspace{0.0cm}
    \includegraphics[trim={0cm, 0.cm, 0cm, 0.cm},clip,width=\textwidth]{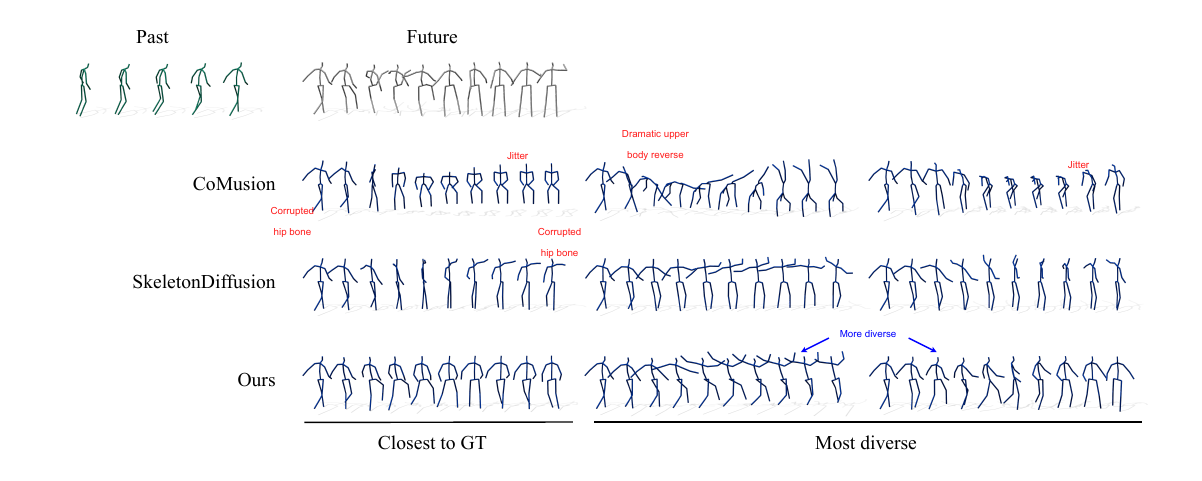}
  \vspace{-0.4cm}
  \caption{Qualitative Results for the cross-topology experiment on H36M of 
  \cref{tab:crossdataset_trainAMASSevalH36M}. 
  Segment n. 1774. }
  \label{fig:qualitative_retarget2}
  \vspace{-0.0cm}
\end{figure*}

\begin{figure*}[t]
  \centering
    \vspace{0.0cm}
    \includegraphics[trim={0cm, 0.cm, 0cm, 0.cm},clip,width=0.9\textwidth]{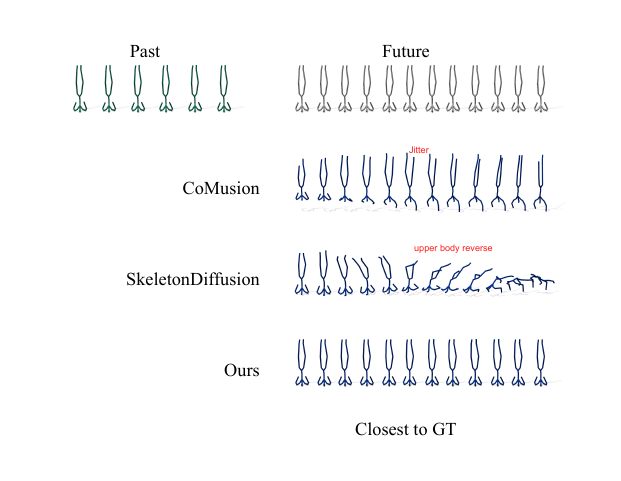}
  \vspace{-0.4cm}
  \caption{Qualitative Results for the out-of-distribution testing on the MoCap Yoga dataset 
  \cref{tab:moyo}. 
  Segment n. 4651. As the setting is quite challenging, we report only the example closest to GT.}
  \label{fig:qualitative_moyo1}
  \vspace{-0.0cm}
\end{figure*}

\begin{figure*}[t]
  \centering
    \vspace{0.0cm}
    \includegraphics[trim={0cm, 0.cm, 0cm, 0.cm},clip,width=0.9\textwidth]{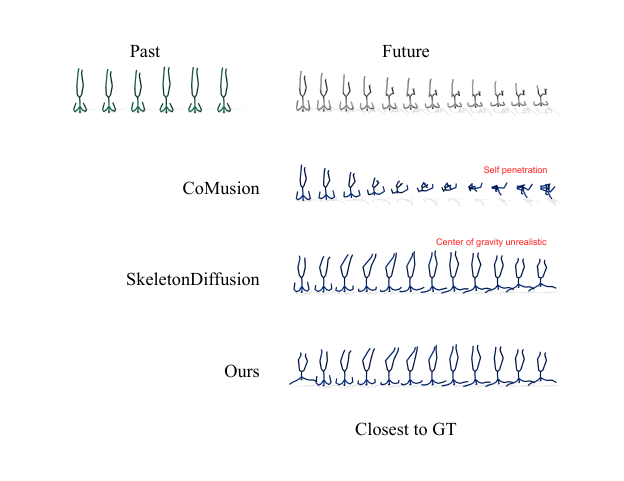}
  \vspace{-0.4cm}
  \caption{Qualitative Results for the out-of-distribution testing on the MoCap Yoga dataset 
  \cref{tab:moyo}. 
  Segment n. 4656.}
  \label{fig:qualitative_moyo2}
  \vspace{-0.0cm}
\end{figure*}

\begin{figure*}[t]
  \centering
    \vspace{0.0cm}
    \includegraphics[trim={0cm, 0.cm, 0cm, 0.cm},clip,width=\textwidth]{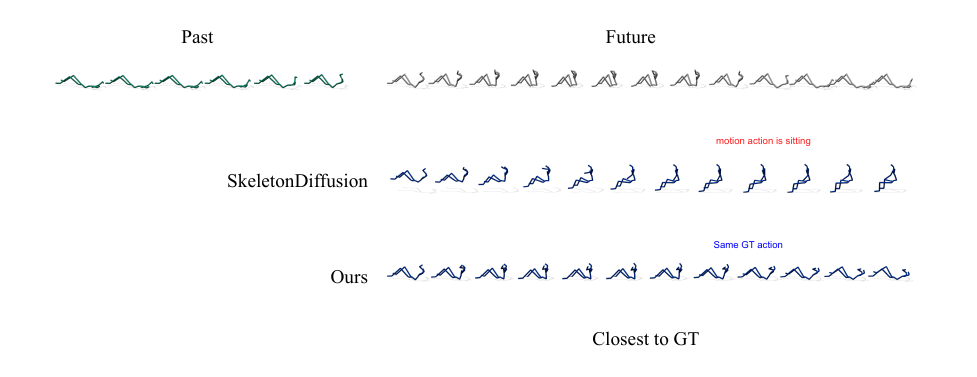}
  \vspace{-0.4cm}
  \caption{Qualitative Example of missing left arm in the observation. SkelDiff has been paired with the symmetric limb pipeline for input completion. Test on AMASS Segment n. 12324.}
  \label{fig:qualitative_missing_left_arm}
  \vspace{-0.0cm}
\end{figure*}

\begin{figure*}[t]
  \centering
    \vspace{0.0cm}
    \includegraphics[trim={0cm, 0.cm, 0cm, 0.cm},clip,width=0.9\textwidth]{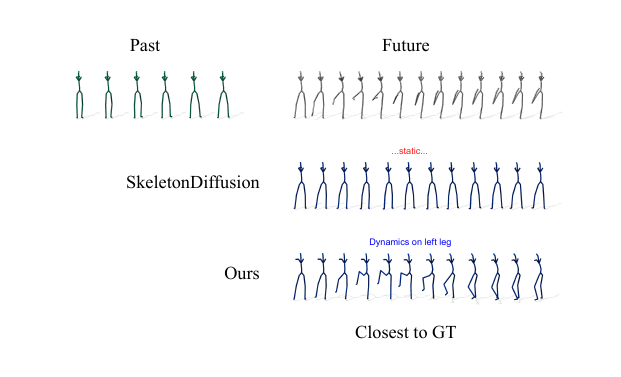}
  \vspace{-0.4cm}
  \caption{Qualitative Example of missing both arms in the observation. SkelDiff can only be paired with the restpose approach to complete the input before further processing. Test on AMASS Segment n. 11100.}
  \label{fig:qualitative_missing_both_arms}
  \vspace{-0.0cm}
\end{figure*}

\begin{figure*}[t]
  \centering
    \vspace{0.0cm}
    \includegraphics[trim={0cm, 0.cm, 0cm, 0.cm},clip,width=0.9\textwidth]{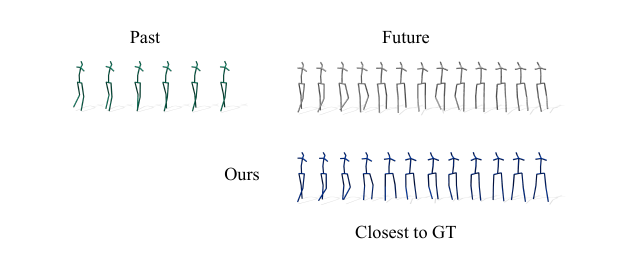}
  \vspace{-0.4cm}
  \caption{Qualitative Example of missing both arms in the observation in a cross-topology setting. We do not compare with other methods, as they would require being extended with both retargeting and completion and be exposed to too high degradation.  Test on H36M Segment n. 200.}
  \label{fig:qualitative_missing_both_arms_cross_topology}
  \vspace{-0.0cm}
\end{figure*}

In \cref{fig:qualitative_retarget1,fig:qualitative_retarget2,fig:qualitative_moyo1,fig:qualitative_moyo2,fig:qualitative_missing_left_arm,fig:qualitative_missing_both_arms,fig:qualitative_missing_both_arms_cross_topology} 
we report qualitative examples for our experiments. Following previous works \cite{yuan2020dlow, barquero2023belfusion, suncomusion, curreli2025nonisotropic},  we report out of 50 predictions, the sample closest to the GT, and the two predictions that maximize diversity when paired with the closest to GT sample. For the most challenging settings (missing limbs, out-of-distribution data), we are only interested in the sample closest to GT.

\clearpage

\end{document}